\documentclass[10pt]{article} 
\pdfoutput=1
\usepackage[preprint]{tmlr}

\usepackage[utf8]{inputenc}                         
\usepackage[T1]{fontenc}                            
\usepackage{microtype}                              
\usepackage{natbib}
\usepackage{color}
\usepackage{xcolor,colortbl}                        
\usepackage{longtable}
\usepackage{booktabs}                               
\usepackage{makecell}
\usepackage[export]{adjustbox}
\usepackage{multirow}
\usepackage{paralist}         
\usepackage{multirow}

\usepackage{comment}                                
\usepackage{soul}                                   
\soulregister\cite7
\soulregister\ref7
\soulregister\pageref7
\usepackage{enumerate}
\usepackage{url}
\usepackage{nth}                                    
\usepackage{courier}
\usepackage{threeparttable}
\usepackage{footnote}
\usepackage{listings}
\usepackage[inline]{enumitem}
\usepackage{verbatim}
\usepackage{etoolbox}                               
\usepackage{pifont}                                 
\usepackage[figuresright]{rotating}
\usepackage{moresize}
\usepackage{fontawesome}

\iftrue
\usepackage[bookmarks=false]{hyperref}
\hypersetup{
    colorlinks = true,
    citecolor  = blue,
    linkcolor  = blue,
    urlcolor   = blue,
}
\usepackage{balance}
\usepackage{setspace}                             
\usepackage{wrapfig}
\fi

\usepackage{amsmath}
\usepackage{amssymb}
\usepackage{amsfonts}                               
\usepackage{amsthm}
\usepackage{physics}                                
\usepackage[mathcal]{eucal}
\usepackage{mathrsfs}
\usepackage{bm}                                     
\usepackage{blkarray}                               
\usepackage{nicefrac}                               

\usepackage{graphicx}                               

\iftrue
\usepackage[subrefformat=parens,farskip=0pt,justification=centering]{subfig}

\else
\usepackage{subfigure}
\fi
\usepackage{caption}
\usepackage{cleveref}
\Crefformat{figure}{Fig.~#2#1#3}                    
\Crefname{subfigure}{Fig.}{Figs.}
\Crefname{figure}{Fig.}{Figs.}
\Crefformat{table}{TABLE~#2#1#3}                    

\usepackage{tikz}                                          
\usepackage{circuitikz}
\usetikzlibrary{patterns,snakes}
\usetikzlibrary{positioning,calc,fit,decorations.pathmorphing,shapes.geometric, shapes.gates.logic.US, calc}
\usetikzlibrary{arrows,arrows.meta,decorations.markings,shapes,shapes.arrows}
\usetikzlibrary{decorations,decorations.pathreplacing}
\usetikzlibrary{backgrounds}
\usepackage{filecontents}                           
\usepackage{pgfplots}
\usepackage{pgfplotstable}
\usepgfplotslibrary{groupplots}
\usepackage{scalefnt}
\pgfplotsset{compat=newest}

\usepackage{algorithm}
\iffalse
\usepackage{algpseudocode}                          
\algrenewcommand\textproc{\texttt}
\makeatletter
\let\OldStatex\Statex
\renewcommand{\Statex}[1][3]{%
  \setlength\@tempdima{\algorithmicindent}%
  \OldStatex\hskip\dimexpr#1\@tempdima\relax
}
\makeatother
\else
\usepackage{algorithmic}
\fi

\newtheorem{mytheorem}{\textbf{Theorem}}

\newtheorem{remark}{Remark}
\newtheorem{proposition}{Proposition}
\newtheorem{myfinding}{\textbf{Finding}}

\crefname{mytheorem}{Theorem}{Theorems}
\crefname{mylemma}{Lemma}{Lemmas}
\crefname{myclaim}{Claim}{Claims}
\crefname{myproperty}{Property}{Properties}
\crefname{mycorollary}{Corollary}{Corollaries}
\crefname{myfinding}{Finding}{Findings}

\newcommand{\cbit}{\begin{compactitem}}
	\newcommand{\ceit}{\end{compactitem}}
\newcommand{\cben}{\begin{compactenum}}
	\newcommand{\ceen}{\end{compactenum}}

\newcommand{\minisection}[1]{\vspace{.1in}\noindent{\textbf{#1}}.}

\usepackage{tcolorbox}
\tcbuselibrary{skins,breakable}
    {\endtcolorbox}
    {\endtcolorbox}

\definecolor{CUHKorange}{RGB}{244,106,18} 
\definecolor{CUHKblue}{RGB}{0,111,190}    
\definecolor{CUHKgreen}{RGB}{0,127,128}   
\definecolor{CUHKred}{RGB}{228,46,36}     
\definecolor{CUHKyellow}{RGB}{198,148,34} 
\definecolor{CUHKdark}{RGB}{114,44,114}   
\definecolor{CUHKmiddle}{RGB}{144,44,144} 
\definecolor{CUHKlight}{RGB}{167,44,167} 
\definecolor{CUHKpurple}{RGB}{117,15,109}
\definecolor{CUHKgold}{RGB}{221,163,0}
\definecolor{CUHKribbon}{RGB}{244,223,176}
\definecolor{CUHKblack}{RGB}{34,24,21}

\usepackage{amsmath,amsfonts,bm}

\def\eqref#1{equation~\ref{#1}}

\def\1{\bm{1}}

\DeclareMathAlphabet{\mathsfit}{\encodingdefault}{\sfdefault}{m}{sl}
\SetMathAlphabet{\mathsfit}{bold}{\encodingdefault}{\sfdefault}{bx}{n}

\graphicspath{{./figs/}}

\title{Consistent Distributed Ranking of Generative Models via Kernel Distances}

\author{\name Zixiao Wang\thanks{Equal Contribution}, Farzan Farnia$^*$, Zhenghao Lin, Yunheng Shen, Bei Yu\\
      \addr \texttt{\{zxwang22, farnia, byu\}@cse.cuhk.edu.hk}\\
      \addr \texttt{elvis@stu.scau.edu.cn \quad shenyh19@mails.tsinghua.edu.cn}}

\def\month{MM}  
\def\year{YYYY} 
\def\openreview{\url{https://openreview.net/forum?id=XXXX}} 

\begin{document}

\maketitle

\begin{abstract}

Ranking generative models based on the fidelity and diversity of their outputs is required to identify the best generator in a group of candidate generative AI models. To rank a group of models in a conventional centralized setting, a standard score is commonly evaluated for each involved model. The selection and design of reference-based evaluation scores have been extensively studied in centralized settings, where the reference samples are drawn from a single probability distribution. However, in practical scenarios including distributed learning contexts, reference samples are distributed across multiple clients, each potentially with a heterogeneous data distribution. In this work, we investigate the ranking of generative models in such distributed settings with heterogeneous data distributions across clients. We focus on the widely used family of kernel distance (KD) evaluation metrics. We prove that, for every kernel function, ranking models by the averaged KD scores of individual clients yields the same ordering as a centralized KD evaluation using the combined reference data from all the clients. We further extend our analysis to other popular metrics, including the Fréchet Distance (FD), for which the individual client scores could be insufficient for accurate model ranking.  We present the numerical results of several experiments on standard image datasets and generative models to validate our theoretical findings regarding distributed ranking using various evaluation scores.

\end{abstract}
\section{Introduction}

Deep generative models, including variational autoencoders~\citep{kingma2013auto}, generative adversarial networks (GANs)~\citep{goodfellow2014generative}, and diffusion models~\citep{ho2020denoising} have attained remarkable results over a wide array of machine learning tasks. These frameworks' success can be attributed to the large capacity of multi-layer neural networks in modeling complex distributions of data, as well as the intricate design of the training mechanisms in these algorithms. The promising results of these generative modeling schemes have inspired the development of several assessment methods to properly compare and rank a group of generative models using their produced data. 

A standard approach to evaluating generative models is to compare the distribution of their generated data with the distribution of samples in a reference set, typically chosen to be the test sample set in a learning problem. This comparison is often made using a distance or similarity measure between the generative and reference distributions. Several standard metrics have been designed, such as the Fréchet Distance~\citep{heusel2017gans}, Kernel Distance~\citep{binkowski2018demystifying}, Precision/Recall~\citep{sajjadi2018assessing,kynkaanniemi2019improved}, Density/Coverage~\citep{naeem2020reliable}, and Feature Likelihood Divergence (FLD)~\citep{jiralerspong2023feature}. In a standard centralized comparison of generative models, only a single reference distribution is considered for the evaluation. Therefore, to rank a group of generative models in a centralized setting, one can sort them according to scores based on the reference distribution.     

\begin{figure}[th]
    \centering
    \includegraphics[width=.44\linewidth]{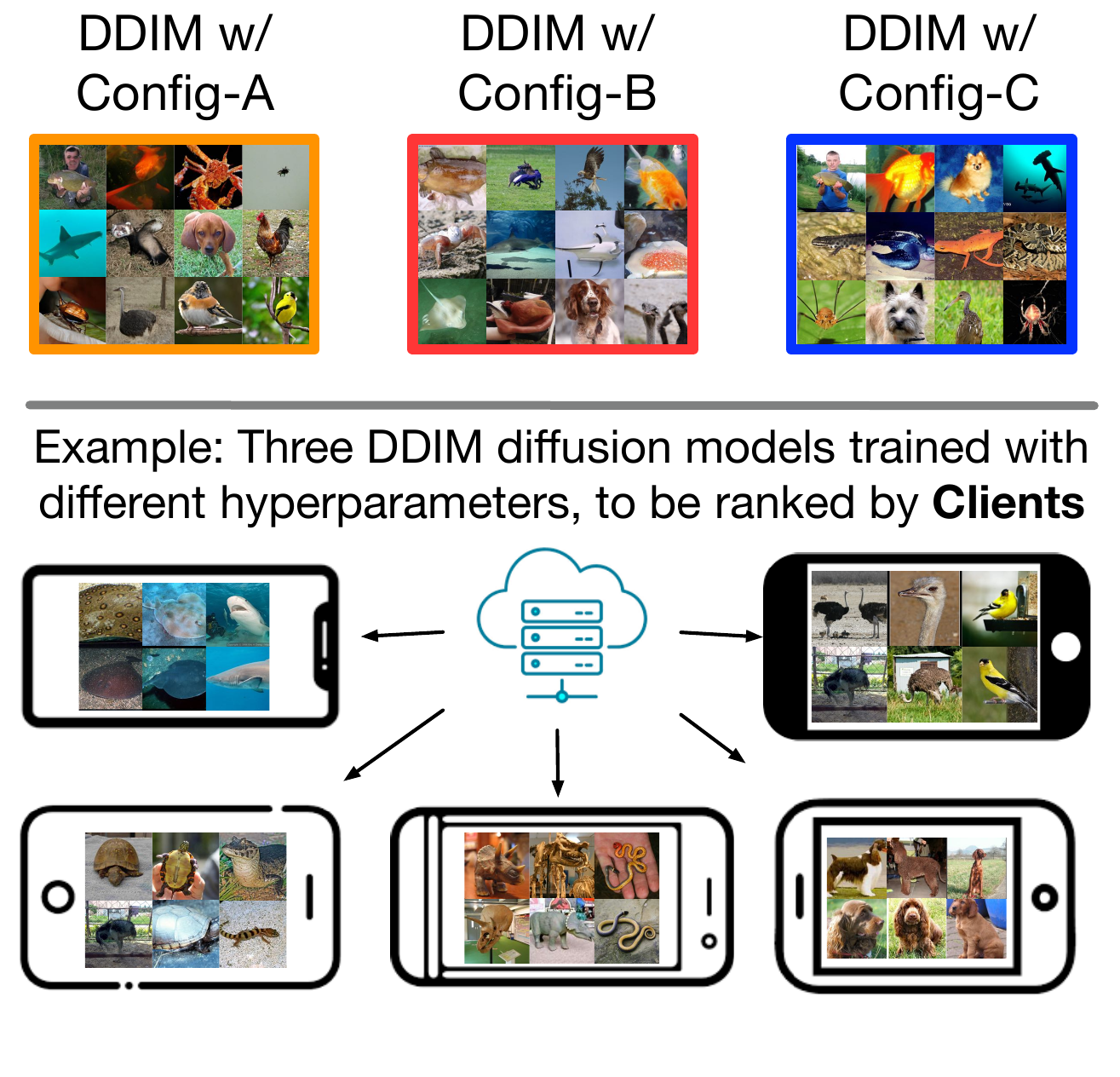}
    \caption{The distributed model ranking setting: the goal is a private ranking of generative models using the reference data of all clients in the network, without them sharing their data.}
    \label{fig:main}
\end{figure}

On the other hand, several modern applications of deep generative models concern \textit{distributed settings} where the reference data are provided by multiple clients in a network. A well-known distributed scenario is the \textit{federated learning setting}~\citep{mcmahan2017communication}, where several clients are connected to a server and aim to train and assess a decentralized model through their communications with the server node while preserving the privacy of their collected data. Considering various architectures, training schedules, and hyperparameters, how to rank these models in such distributed settings can be a fundamental question, which has been less explored in current literature. An important consideration for conducting evaluation in distributed settings is the potential heterogeneous data distributions across clients, since the background features of every client could lead to a different data distribution. For example, as displayed in \Cref{fig:main}, the distribution of photos of clients in a mobile network could be highly diverse as they could possess different interests and characteristics.

In this work, we focus on ranking generative models in distributed settings and attempt to study the assessment task's complexity under diverse client data distributions.
Following the existing literature on training generative models in distributed settings~\citep{rasouli2020fedgan, zhao2018federated,hardy2019md,yonetani2019decentralized,wuf2022edcg,su2023dual}, the ultimate goal is to perform a centralized evaluation where the reference set is the collective dataset of all clients' data. In our analysis, we call the evaluation score following such a centralized approach the score-all metric, which means the score is computed using a reference set with all the clients' data. 

However, privacy restrictions in distributed settings would limit the use of such an evaluation method because the assessment process requires clients to share their data with the server. Another sensible approach to private evaluation in distributed settings is to query every client about her own evaluation score for the generative model and then aggregate the clients' individual evaluation scores. While this aggregation-based approach needs every client to share only her score, the resulting ranking of generative models could be different from the centralized score-all evaluation. Therefore, a key question is how consistent a score aggregation-based evaluation could be with the score-all assessment.

We answer the mentioned question for the class of KD evaluation metrics. For any kernel function, we prove that accessing the clients' individual KD scores will provide sufficient information for ranking a group of generative models using the centralized KD-all metric. Specifically, we prove that for every kernel function used in the KD definition, the weighted average of clients' KD scores, which we call KD-avg, will result in the same ranking of generative models as the potentially inaccessible KD-all score. This result reveals the feasibility of a private ranking of the models according to the centralized KD-all score, which is identical to the ranking of the generative models using the averaged KD-avg score. Apart from KD distance and the Recall score~\citep{sajjadi2018assessing} for which we observe that Recall-all and Recall-avg take the same value, we show that such a conclusion does not universally hold for other evaluation metrics, including the FD metric, Precision and Density/Coverage scores.

Finally, we present several numerical results on standard image datasets and generative model architectures that align with our analytical findings on the aggregation of distance-based evaluation scores. In the distributed ranking scenario, we show that the average Kernel Distance (KD-avg) consistently maintains a fixed gap with the centralized Kernel Distance (KD-all), resulting in identical model rankings, regardless of whether the models are trained in a centralized or distributed manner. We also highlight cases where two models yield identical client-based Fréchet Distance scores, yet their FD scores with respect to the collective reference data differ significantly. Experimental results using other evaluation metrics such as Precision, Density, and Coverage further demonstrate the challenges of ranking generative models in distributed settings. We summarize the main contributions of our study as follows:

\begin{itemize}
\item We highlight and study the task of distributed ranking of generative models.
\item We show the identical model rankings of the average KD score (KD-avg) and the centralized KD score (KD-all) for the entire family of KD scores induced by different kernel functions.
\item We demonstrate the insufficiency of the individual client evaluated Frechet distance (FD) scores in obtaining the global ranking of generative models. 
\item We numerically validate our theoretical results on the consistent distributed ranking of generative models using the KD metric.
\end{itemize}

\section{Related Work}
A large body of related works has focused on the evaluation of generative models in standard centralized learning settings. The existing evaluation scores can be categorized into two general groups: 1) Distance-based metrics, which define a distance between the distribution of training data and the learned generative model. The distance between the real and fake distributions is usually computed after passing the samples through a pre-trained neural net, which offers a proper embedding of image data. The well-known evaluation scores in this category are the FD \citep{heusel2017gans} and KD \citep{binkowski2018demystifying} scores.
2) Quality and diversity-based scores, which output a score based on the sharpness and variety of the generated samples. The widely-used evaluation metrics in this category are the Inception score \citep{salimans2016improved}, precision and recall metrics \citep{sajjadi2018assessing,kynkaanniemi2019improved}, density and coverage scores \citep{naeem2020reliable}, authenticity score~\citep{alaa_how_2022}, feature likelihood divergence~\citep{jiralerspong2023feature}, and the rarity score~\citep{han2022rarity}. We note that our goal is to analyze the consistency of the rankings in the distributed case, and specifically show the KD metric family provides consistent score-averaged and score-all rankings. For the other scores, we observe that Recall-all and Recall-avg are always equal, while we discuss that the local FD, precision, density, and coverage scores could be insufficient for obtaining the global model ranking.

We note that the evaluation scores are usually measured in an embedding space, and our study of the distributed evaluation applies the standard assumption of a shared evaluation embedding model across clients. The related works by \citet{stein2023exposing} and \citet{kynkaanniemi_role_2022} recommend using the DINOv2~\citep{oquab2023dinov2} and CLIP~\citep{radford2021learning} models for image generative models, respectively. Also, we note that we focus on the reference-based scores where a set of reference data is needed for the evaluation, which does not apply to reference-free diversity metrics such as Vendi score~\citep{friedman_vendi_2023} and RKE score~\citep{jalali_information-theoretic_2023}. We note that the fidelity and novelty cannot be directly measured via a reference-free score in unconditional sample generation.

In another set of related works, extensions of generative model training methods, including GANs and diffusion models to distributed federated learning, have been studied.
\citet{rasouli2020fedgan} propose Fed-GAN to train GANs in a federated learning setting.
In \citet{hardy2019md}, the gradient from sample generation for the generator is exchanged on the server, while each client possesses a personalized discriminator.
According to \citet{yonetani2019decentralized}, different weights are assigned to local discriminators in the non-i.i.d.~setting.
Conversely, \citet{wuf2022edcg} ensures client privacy by sharing the discriminator across clients, while keeping the generator private.
Additionally, \citet{su2023dual} explores a dual diffusion paradigm to extend diffusion-based models into the federated learning setting, addressing concerns related to data leakage. However, in these works, the evaluation and ranking of generative models are still based on the centralized evaluation scores, which are difficult to obtain in real distributed training settings.

\section{Preliminaries}

To rank a set of generative models based on their performance, a common approach is to assign each model a score that reflects the distance between the distributions of real and generated data. The models are then ranked by comparing these scores. Given the high dimensionality of image data, evaluations of image-based generative models are typically conducted in a feature space extracted by a pre-trained network, such as Inception-V3~\citep{szegedy2016rethinking}, DINOv2~\citep{oquab2023dinov2}, or CLIP~\citep{radford2021learning}.

A widely used class of distance-based scores for the evaluation of generative models is the kernel distance, which measures the squared maximum mean discrepancy (MMD$^2$) between two distributions using a kernel similarity function $k:\mathbb{R}^d\times \mathbb{R}^d\rightarrow \mathbb{R}$. Here, the definition of the MMD$^2$ distance between $P_X$ and $P_G$ based on kernel $k$ follows from
\begin{equation*}
\begin{aligned}  \;\mathrm{KD}\bigl(P_X,P_G\bigr):=\;
     \sup_{\substack{f\in \mathcal{H}_k:\, \Vert f\Vert_{\mathcal{H}_k}\le 1}} \Bigl(\mathbb{E}_{X\sim P_X}\bigl[f(X) \bigr] - \mathbb{E}_{X'\sim P_G}\bigl[f(X') \bigr] \Bigr)^2.
\end{aligned}
\end{equation*}
In the above, $\mathcal{H}_k$ represents the reproducing kernel Hilbert space (RKHS) corresponding to kernel $k$, i.e., the function set that can be expressed by linear combination of kernel-based elements $k_{x}(\cdot)=k({x},\cdot)$ over $x\in\mathcal{X}$, and $\Vert\cdot\Vert_{\mathcal{H}_k}$ denotes the norm of this Hilbert space of functions.

Additionally, another standard distance-based metric is the Fréchet distance defined as the $2$-Wasserstein distance between two Gaussian distributions with the mean and covariance parameters of the data distribution $P_X$, denoted by $\boldsymbol{\mu}_X ,C_X$, and with the mean and covariance of the generative model $P_G$, denoted by $\boldsymbol{\mu}_G ,C_G$:
\begin{align*}
    \mathrm{FD}(P_X, P_{G})\: :=\: \bigl\Vert \boldsymbol{\mu}_X- \boldsymbol{\mu}_G \bigr\Vert^2_2 
    \: + \mathrm{Tr}\Bigl(C_X+C_G - 2\bigl(C_XC_G\bigr)^{1/2}\Bigr).
\end{align*}
Note that we can interpret the FD score as an approximation of the 2-Wasserstein distance given the first and second-order moments of the distributions.

\section{Evaluation of Generative Models in Distributed Contexts}
\label{sec:4}

In this section, we discuss two extensions of distance-based evaluation scores from a centralized case to heterogeneous distributed learning settings. In our analysis, we use $\mathcal{D}(P_X,P_{G})$ to denote a general distance between data distribution $P_X$ and the generative model $P_G$. Considering the standard distance-based metrics, $\mathcal{D}$ can be chosen to be the FD or KD score, which we will analyze in this section.

In a standard centralized setting, we have only a single distribution $P_X$ for real data. However, the main characteristic of a heterogeneous distributed learning problem is the multiplicity of the involved clients' distribution. Here, we suppose a distributed setting with $k$ clients and use $P_{X_1},\ldots, P_{X_k}$ to denote their underlying distributions, i.e. $P_{X_i}$ stands for the data distribution at client $i$. In addition, we assume that every client $i$ has a fraction $0\le \lambda_i \le 1$ of the data in the network, i.e, $\lambda_i = \frac{n_i}{n}$ where $n=\sum_{j=1}^k n_j$ is the total number of samples in the network and $n_i$ is the number of samples at client $i$.

As a result of multiple input distributions, we need to define an aggregate evaluation score that is based on distance measure $\mathcal{D}$. The aggregate distance is supposed to summarize the performance of the generative model $P_G$ in only one score. To do this, we consider and analyze two reasonable ways of defining the aggregate score:
\begin{enumerate}[leftmargin=15pt]
\item \textbf{Collective-data-based Score $\mathcal{D}_{\mathrm{all}}$}: The score-all with respect to the collective data of the clients is the distance between $P_G$ to the averaged distribution $\widehat{P}_X := \sum_{i=1}^k \lambda_i P_{X_i}$: 
    \begin{equation}
        \mathcal{D}_{\mathrm{all}}\Bigl( P_{X_1},\ldots,P_{X_k}\, ; \, P_G \Bigr)  \, :=\,  \mathcal{D}\Bigl(\widehat{P}_X \, ,\, P_G\Bigr).
    \label{eq:all}
    \end{equation}
    
    \item \textbf{Average Score $\mathcal{D}_{\mathrm{avg}}$}: The score-avg is the mean of the client's individual distance measures, i.e.
    \begin{equation}
        \mathcal{D}_{\mathrm{avg}}\Bigl( P_{X_1},\ldots,P_{X_k}\, ; \, P_G \Bigr)  \, :=\, \sum_{i=1}^k \lambda_i\mathcal{D}\bigl(P_{X_i} , P_G\bigr).
    \label{eq:avg}
    \end{equation}
    
\end{enumerate}
In the above, note that $\sum_{i=1}^k \lambda_i P_{X_i}$ is indeed a mixture distribution with $k$ components $P_{X_1},\ldots , P_{X_k}$ with frequency parameters $\lambda_1,\ldots,\lambda_k$. As discussed in the introduction, the score-all metric is the desired evaluation score following the literature on distributed generative modeling. However, due to the privacy and communication costs, the score-avg metric is cheaper to compute over the network. Therefore, understanding the connections between the rankings of the models following the two aggregate scores is the key to apply the score-avg as a surrogate for the score-all metric in the assessment process.


\subsection{KD-based Evaluation in Distributed Settings}
\label{sec:4.2}

Here, we focus on the KD score and analyze the consistency of model rankings following KD-all and KD-avg scores. The following theorem proves that KD-all and KD-avg are guaranteed to provide a consistent ordering of a group of generative models, and there is a monotonic relationship between the two scores independent of the choice of the generative model. We defer the proof to the Appendix.

\begin{mytheorem}\label{Thm: KD}
    Consider a kernel function $k:\mathbb{R}^d\times \mathbb{R}^d\rightarrow \mathbb{R}$ and the resulting KD score. Then for the clients' distributions $P_{X_1},\ldots , P_{X_k}$ with frequency parameters $\lambda_1,\ldots , \lambda_k$, we will have the following for the average distribution $\widehat{P}_X=\sum_{j=1}^k \lambda_jP_{X_j}$:
    \begin{equation*}
    \begin{aligned}
        \mathrm{KD}_{\mathrm{avg}}\Bigl( P_{X_1},\ldots,P_{X_k}\, ; \, P_G \Bigr) \, 
        = \, \mathrm{KD}_{\mathrm{all}}\Bigl( P_{X_1},\ldots,P_{X_k}\, ; \, P_G \Bigr) + \sum_{i=1}^k \lambda_i\, \mathrm{KD}\bigl(\widehat{P}_X,P_{X_i}\bigr), 
    \end{aligned}
    \end{equation*}
    which implies a monotonic  relationship between KD-all and KD-avg as a function of $P_G$. 
\end{mytheorem}
The above theorem shows the consistent rankings implied by the aggregate KD-avg and KD-all scores, as the difference between the scores remains constant while changing the model $P_G$. Therefore, to rank a set of models based on the desired KD-all score, the clients only need to share their own evaluated KD scores, and the server can average the reported scores to recover the models' ranking. It is worth noting that \Cref{Thm: KD} holds for any kernel function satisfying the standard conditions for RKHS.\vspace{2mm} 

\noindent \textbf{FD-based Evaluation in Distributed Settings}. 
However, not all evaluation metrics support the recovery of centralized rankings from distributed evaluation results. In the case of FD score, we utilize the formulation of FD as the 2-Wasserstein distance between fitted Gaussian distributions, which leads to a Riemannian geometry. This observation results in the following theorem on FD-all and FD-avg aggregations. 

\begin{mytheorem}\label{Thm: FD}
Suppose that $P_{X_1},\ldots , P_{X_k}$ are the clients' distributions with the mean parameters $\boldsymbol{\mu}_1,\ldots ,\boldsymbol{\mu}_k$ and covariance matrices $C_1,\ldots ,C_k$, respectively, in the network embedding. Then, the following holds for a generative model $P_G$ with mean $\boldsymbol{\mu}_G$ and covariance $C_G$.
\begin{enumerate}[leftmargin=*]
    \item For FD-all, if we define random $\widehat{X}$ with mean $\widehat{\boldsymbol{\mu}} = \sum_{i=1}^k \lambda_i \boldsymbol{\mu}_i$ and  covariance matrix $\widehat{C} = \sum_{i=1}^k \lambda_i\bigl( C_i + \boldsymbol{\mu}_i\boldsymbol{\mu}_i^\top - \widehat{\boldsymbol{\mu}}\widehat{\boldsymbol{\mu}}^\top \bigr)$, then
    \begin{equation}
    \begin{aligned}
        \mathrm{FD}_{\mathrm{all}}\Bigl( P_{X_1},\ldots,P_{X_k}\, ; \, P_G \Bigr) \, = \, \mathrm{FD}\bigl( P_{\widehat{X}} , P_G \bigr).
    \end{aligned}
    \end{equation}
    \item For FD-avg, if we define 
 $\widetilde{X}$ with  mean $\widehat{\boldsymbol{\mu}} = \sum_{i=1}^k \lambda_i \boldsymbol{\mu}_i$ and  covariance matrix $\widetilde{C}$ as the unique solution to $\widetilde{C} = \sum_{i=1}^k \lambda_i \bigl(\widetilde{C}^{1/2}C_i \widetilde{C}^{1/2}\bigr)^{1/2}$, then
    \begin{equation*}
    \begin{aligned}
        \mathrm{FD}_{\mathrm{avg}}\Bigl( P_{X_1},\ldots,P_{X_k}\, ; \, P_G \Bigr) \, = \, \mathrm{FD}\bigl( P_{\widetilde{X}}, P_G \bigr) + \sum_{i=1}^k \lambda_i \mathrm{FD}\bigl(P_{\widetilde{X}},P_{X_i}\bigr). 
    \end{aligned}
    \end{equation*}
\end{enumerate}
\end{mytheorem}
\begin{remark}
    In Theorem~\ref{Thm: FD}, $\widetilde{C}$ represents the covariance of the Wasserstein barycenter of the Gaussian distributions. If $C_1,\ldots ,C_k$ commute, i.e. $\forall i,j:\: C_iC_j=C_jC_i$, it simplifies to $\widetilde{C}  = \bigl(\sum_{i=1}^k \lambda_i C_i^{{1}/{2}} \bigr)^2 $.
\end{remark}

    In the above theorem, we show the optimal mean vectors for FD-all and FD-avg aggregations are the same. In contrast, the optimal covariance matrix of FD-avg denoted by $\widetilde{C}$ has no dependence on the choice of $\boldsymbol{\mu}_i$'s, while the optimal covariance matrix of FD-all denoted by $\widehat{C}$ will be affected by the difference between $\boldsymbol{\mu}_i$'s due to the term $\sum_{i=1}^k\lambda_i (\boldsymbol{\mu}_i\boldsymbol{\mu}_i^\top- \widehat{\boldsymbol{\mu}}\widehat{\boldsymbol{\mu}}^\top)$. 
Therefore, Theorem \ref{Thm: FD} shows the distance-minimizing covariance matrices for FD-all and FD-avg are different in heterogeneous settings where $\boldsymbol{\mu}_i$'s are not identical, which shows the existence of models inconsistently ranked by FD-all and FD-avg. 


As the FD-avg aggregation may not  provide the FD-all's ranking, a natural question is whether there exists another aggregation of the individual $\mathrm{FD}(P_{X_i},P_G)$ scores which maintains the ranking of centralized FD-all evaluation. The following proposition shows that the answer to this question is no and it is possible that a group of clients assign identical individual FD scores to two generative models $P_G$ and $P_{G'}$, while $P_G$ and $P_{G'}$ have different FD-all scores. 

\begin{proposition}\label{prop: FD-equal}
In Theorem \ref{Thm: FD}'s setting, suppose there exists $i,j$ that $\boldsymbol{\mu}_i \neq \boldsymbol{\mu}_j$. Also, assume the number of clients is less than the dimension of embedding.  Then, if we consider the FD-all minimizing $\widehat{G}$ with $\widehat{\boldsymbol{\mu}}$ and $\widehat{C}$ parameters, there exists a different generator $G'$ that shares the same client-assigned FD scores $\mathrm{FD}(P_{X_i}, P_{G'}) = \mathrm{FD}(P_{X_i}, P_{\widehat{G}})$ for all $i$, while 
\begin{align*}
    0<2\,\mathrm{Tr}\Bigl(\sum_{i=1}^k \lambda_i \bigl(\boldsymbol{\mu}_i\boldsymbol{\mu}_i^\top - \widehat{\boldsymbol{\mu}}\widehat{\boldsymbol{\mu}}^\top\bigr) \Bigr) \, \le\,  \mathrm{FD}_{\mathrm{all}}\Bigl( P_{X_1},\ldots,P_{X_k}\, ; \, P_{G'} \Bigr)  -  \mathrm{FD}_{\mathrm{all}}\Bigl( P_{X_1},\ldots,P_{X_k}\, ; \, P_{\widehat{G}} \Bigr).
\end{align*}

\end{proposition}

As a result, two generative models may receive identical FD scores from a group of clients, yet their centralized FD-all scores which are computed from the same client data, can differ. This implies that neither averaging nor any other aggregation of clients’ FD scores can consistently preserve the ranking induced by the FD-all metric. Although this conclusion may seem counter-intuitive, we provide an example in \Cref{tab:same_client_fid,fig:same_client_fid} to illustrate and support this finding.

\section{Numerical Results}

In this section, we first present numerical results demonstrating that KD-avg provides a ranking consistent with KD-all for models trained in both centralized and distributed settings. Next, we provide an example supporting our claim that it is possible for two models to yield nearly identical FD scores across all clients, while their overall quality and FD-all scores differ significantly. We then extend the discussion by examining additional evaluation metrics, including precision, recall, density, and coverage. Finally, we illustrate how the consistency property of KD can be leveraged to support the distributed training of generative models.

\subsection{Distributed Evaluation with KD-avg}
\label{subsec:kid}

In our theoretical analysis, KD-avg has a fixed gap from KD-all, which guarantees the consistency of a given ranking. Here we conduct numerical experiments on standard models and datasets to support this claim.

\minisection{Models} To further elaborate on the discussion in \Cref{fig:main}, we evaluate several prominent generative models that are trained in standard centralized settings, including BigGAN~\citep{brock2018large}, GigaGAN~\citep{kang2023scaling}, StyleGAN-XL~\citep{sauer2022stylegan}, ADM~\citep{dhariwal2021diffusion}, LDM~\citep{rombach2022high}, and DiT-XL-2~\citep{peebles2023scalable}, on the ImageNet generation task at a resolution of $256\times256$. Image samples are sourced from a recent research project \citep{stein2023exposing}. Furthermore, in distributed training scenarios, it is often necessary for trainers to decide among various configurations, including model architecture, hyperparameters, data augmentation methods, and more. To simulate such a scenario, we trained StyleGAN-XL in a distributed way with the FedAvg \citep{mcmahan2017communication} algorithm with three variants. In all variants, we employ one hundred clients, each of which holds random samples drawn from the ImageNet training set according to a Dirichlet distribution parameterized by $\beta=0.5$, as suggested by \citet{li2022federated}. The total training length in each variant is aligned with the official implementation. In \textbf{Config-A}, we adapt all the hyperparameters from the official documents and utilize a relatively small interval (one global average after ten local updates). In \textbf{Config-B}, we enlarge the communication interval by ten times to save communication cost. In \textbf{Config-C}, we reduce the stages of progressive growing and enlarge the resolution by four times instead of doubling. 

\minisection{Evaluation Metrics} In the community of generative models, several evaluation metrics have been proposed to rank the model performance. We include the well-known Fréchet distance, which computes the FD between sets of 50k real and generated samples in the representation space (for example, Inception-V3 or DINOv2). Also, we include FD$_\infty$, KD, and feature likelihood score (FLS) for a full comparison. For KD, we choose the popular polynomial kernel with an order of 3. For these evaluation metrics, we report the centralized evaluation score. We also report the distributed KD evaluation score for the comparison as we discussed in the main text. For both centralized and distributed settings, we utilized a pre-trained DINOv2-VIT-L/14~\citep{oquab2023dinov2} as the feature extractor as suggested in \citet{stein2023exposing}.

\begin{table*}[t]
    \centering
    \footnotesize
    \caption{Evaluation results of generative models on the $256\times256$ ImageNet generation task. For each metric, the \textbf{best} value and the \underline{second best} are marked. Centralized evaluation metrics are sourced from~\citep{stein2023exposing}.}
    \label{tab:KD}
    \renewcommand{\arraystretch}{1.0}
    \begin{tabular}{c|c|ccccc|cc}
    \toprule
    \multirow{2}{*}{\textbf{Train Setup}} & \multirow{2}{*}{\textbf{Model} }
        & \multicolumn{5}{c|}{\textbf{Centralized Evaluation}} 
        & \multicolumn{2}{c}{\textbf{Distributed Evaluation}} \\
    \cmidrule(lr){3-7} \cmidrule(lr){8-9}
    & & \textbf{KD-all} & \textbf{FD-all} & \textbf{FD$_\infty$-all} & \textbf{FLS-all} &\textbf{Ranking}
      & \textbf{KD-avg} &\textbf{Ranking}\\
    \midrule
    \multirow{6}{*}{\rotatebox[origin=c]{0}{\shortstack{\textbf{Centralized}}}}
        & BigGAN       & 0.52 & 401.22 & 393.13 & 66.48 &6&  1.33 &6 \\
        & GigaGAN      & 0.24 & 228.37 & 219.27 & 74.61 &5&  1.04 &5 \\
        & StyleGAN-XL  & 0.22 & 214.88 & 207.09& 75.16 &4&  1.02 &4\\
        & ADM          & 0.19 & 203.45 & 195.01& 77.41 &3&  0.99 &3\\
        & LDM          & \underline{0.09} & \underline{112.40} & \underline{103.73} & \underline{87.92} &  2&\underline{0.91}&2 \\
        & DiT-XL-2     & \textbf{0.06} & \textbf{79.36} & \textbf{70.32}  & \textbf{92.20} &1& \textbf{0.87}&1 \\
    \midrule
    \multirow{3}{*}{\rotatebox[origin=c]{0}{\shortstack{\textbf{Distributed}}}} & Config-C     & 0.36 & 276.19 & 267.78 & 68.81 &3&  1.16 &3\\
        & Config-B     & \underline{0.30} & \underline{246.19} & \underline{237.73} & \underline{70.81} &  2&\underline{1.10}&2 \\
        & Config-A     & \textbf{0.25} & \textbf{226.81} & \textbf{218.41} & \textbf{73.11} & 1& \textbf{1.05} &1\\
       
    \bottomrule
    \end{tabular}
\end{table*}
\begin{figure}[t]
\centering 
\includegraphics[width=0.5\linewidth]{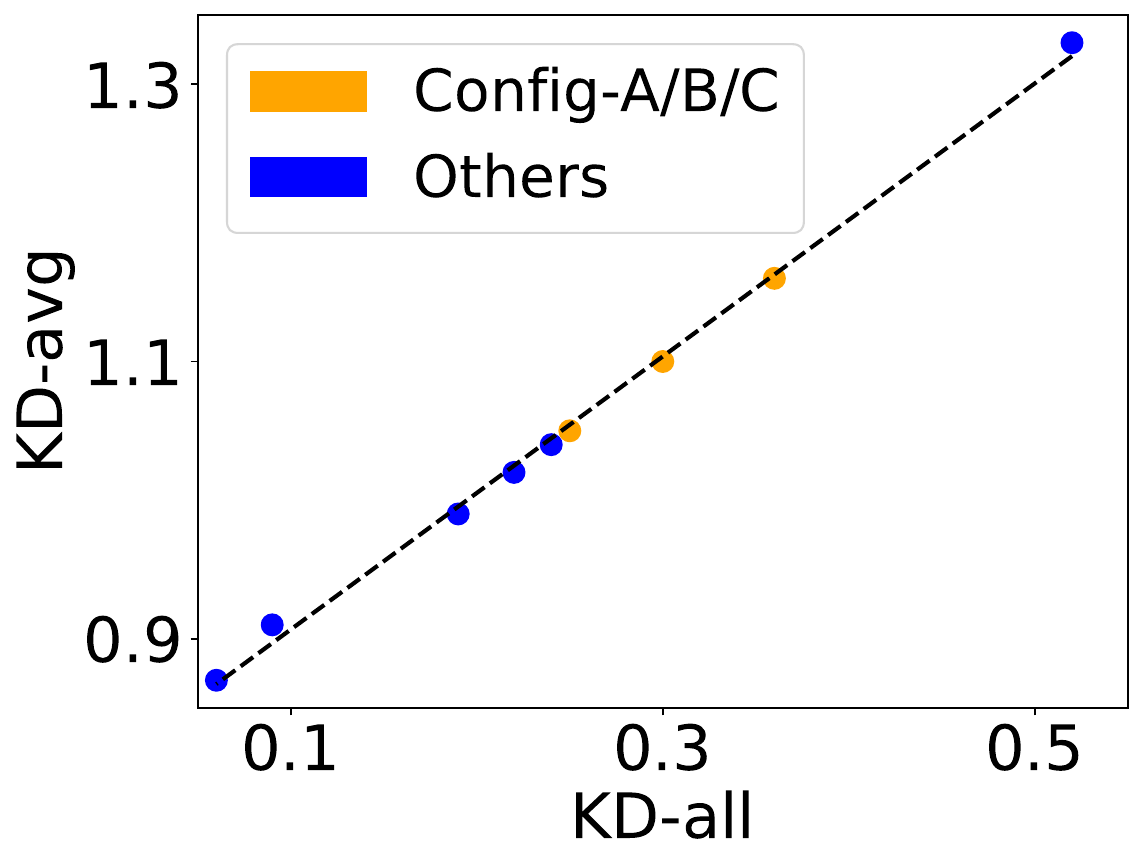} 
\caption{The relationship between KD-avg and KD-all in \Cref{tab:KD}. Models are marked by colors according to the training setup (centralized or distributed).} 
\label{fig:KD} 
\end{figure}

\minisection{Results} We report the evaluation results in \Cref{tab:KD}. For models that are trained in both centralized settings and distributed settings, the numerical results align closely with our theoretical predictions. In \Cref{fig:KD}, we visualize the gap between KD-avg and KD-all, which remains consistent ($\sim 0.8$ in our experiments) and depends solely on the client data distribution, confirming the stability of rankings derived from these metrics. Furthermore, we observe a strong correlation between rankings based on KD-avg scores and those based on standard centralized evaluation scores, corroborating findings from recent empirical evaluations~\citep{stein2023exposing}. These empirical experiments show that
\begin{myfinding}
KD-avg can serve as a practical metric in distributed settings, given its minimal privacy cost and the almost equal ranking with other evaluation metrics like FD, FD$_\infty$, and FLS.
\end{myfinding}

Additionally, we tested our hypothesis on several datasets, including Gaussian mixture clients, AFHQ \citep{Karras2020ada}, CIFAR \citep{krizhevsky2009learning}, and ImageNet \citep{deng2009imagenet, chrabaszcz2017downsampled}. Due to limited space, these additional results are provided in the Appendices.

An extended consideration is that, in scenarios where clients prefer not to share their local KD scores with the server, a zero-mean noise (e.g., Gaussian noise) can be added to the local scores. The server would still be able to accurately recover the ranking of KD-avg.

\subsection{Distributed Evaluation Metrics Beyond KD}
\label{sec:PRDC}

In the previous section, we focused on distributed evaluation using KD-avg. Here, we extend our discussion to other evaluation metrics to examine whether they also exhibit the property of ranking consistency in the distributed setting.

\minisection{Fréchet distance} FD is the most widely used evaluation metric to measure the quality of generated samples. A straightforward question is that can we obtain the model ranking given by FD-all by aggregating the local FD score with some aggregation strategies? Unfortunately, our theoretical analysis shows that \textbf{no aggregation strategy of local FD scores can guarantee a consistent ranking with FD-all}. We provide an example here to show that even when two models receive identical FD evaluation scores from individual clients, their centralized FD scores may still differ significantly.

\begin{table}[t!]
    \centering
    \caption{Individual FD-scores for each client and aggregated FD-scores. ($G_i$: Generator $i$, $C_j$: Client $j$.)}
    \label{tab:same_client_fid}
    \begin{tabular}{c c c c c c c c c c}
    \toprule
    \textbf{Generator} & \textbf{$C_1$} & \textbf{$C_2$} & \textbf{$C_3$} & \textbf{$C_4$} & \textbf{$C_5$} & \textbf{$C_6$} & \textbf{FD-avg} & \textbf{FD-all} \\ 
    \midrule
    $G_1$ & 281.76 & 198.54 & 212.00 & 225.21 & 129.14 & 123.59 & 195.04 & 100.49 \\[2pt]
    $G_2$ & 282.44 & 199.24 & 212.36 & 226.21 & 129.34 & 122.61 & 195.37 & 190.93 \\
    \midrule
    \textbf{Difference} & 0.68 & 0.70 & 0.36 & 1.00 & 0.20 & 0.98 & \textcolor{blue}{\textbf{0.33}} & \textcolor{red}{\textbf{90.44}} \\
    \bottomrule
    \end{tabular}
\end{table}

\begin{figure}[t!]
    \centering
    \includegraphics[width=0.68\textwidth]{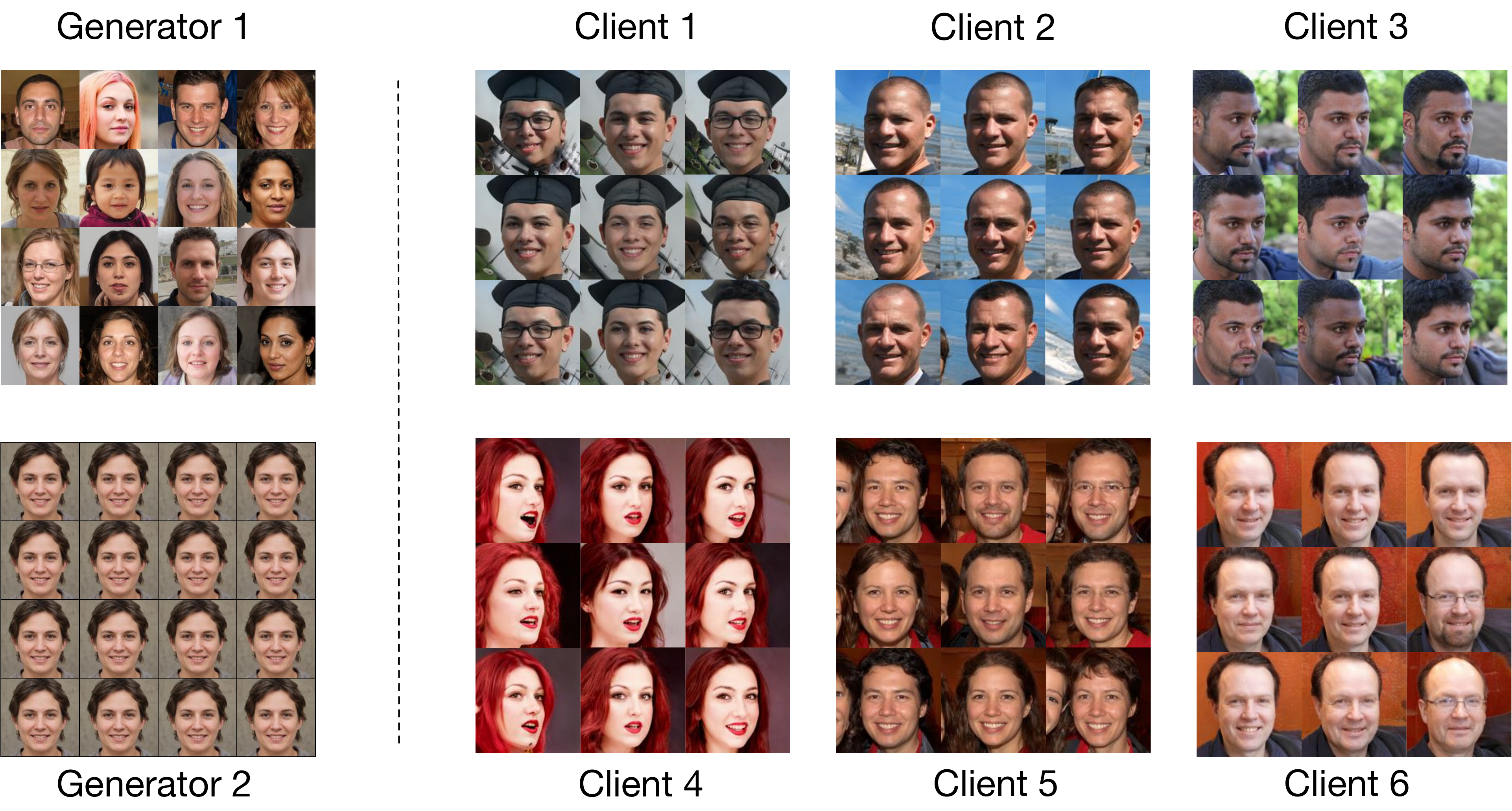}
    \caption{Visual examples of generators and clients in \Cref{tab:same_client_fid}.}
    \label{fig:same_client_fid}
\end{figure}

We employed the pre-trained StyleGAN2-ADA generator, as provided by the official repository, to generate images of resolution $1024 \times 1024$ pixels from the FFHQ dataset \citep{Karras2020ada}. To produce two generators with distinct image diversity, we applied the truncation technique \citep{kynkaanniemi2019improved} and varied the truncation factor ($\tau$) within the set \{0.01, 0.6\}; a higher truncation factor yields higher diversity. To simulate heterogeneous client distributions, we again used truncation to generate 50k image samples for each client, each centered around randomly selected mean latent vectors. Consequently, images generated for a given simulated client appeared similar internally but markedly different from those generated for other clients. Representative samples from these generators and simulated clients are illustrated in \Cref{fig:same_client_fid}. Such heterogeneity realistically mirrors scenarios in actual mobile networks, where image diversity within each user's collection (intra-client) is significantly lower than the diversity across different users (inter-client).

For evaluation, we calculated individual FD scores for each generator-client pair and computed their average (FD-avg) as defined in \Cref{eq:avg}. Subsequently, we calculated centralized FD-all scores by aggregating data across all clients according to \Cref{eq:all}. We use a pre-trained Inception-V3~\citep{szegedy2016rethinking} as the feature extractor here. 

As detailed in \Cref{tab:same_client_fid}, our experiments identified six simulated clients with negligible differences in their client-based FD scores (differences constrained within 1.0), signifying nearly identical performance of the two generators from each client's perspective. Correspondingly, the FD-avg scores computed from these clients also closely matched (195.04 vs. 195.37). However, a substantial discrepancy emerged when evaluating the centralized FD-all scores across all clients (100.49 vs. 190.93). This pronounced gap empirically confirms our theoretical assertion: Similar client-level FD scores do not necessarily imply similar centralized FD-all scores, irrespective of the aggregation strategy used.

We have extended our experiments on validating the inconsistency of FD across multiple settings to strengthen our findings. Specifically, we conducted more complex experiments on multiple datasets to demonstrate the non-monotonicity between centralized and distributed ranking of FD evaluation. A detailed experimental setting and results can be found in \Cref{sec:afhq}.

\begin{figure}
    \centering
    \subfloat[Precision]{\includegraphics[height=2.5cm]{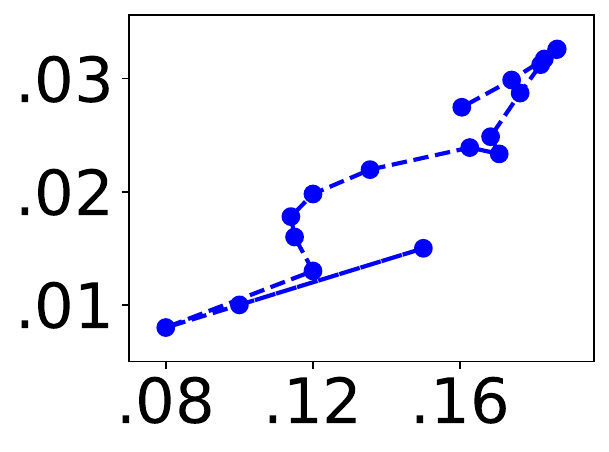}}
    \subfloat[Recall]    {\includegraphics[height=2.5cm]{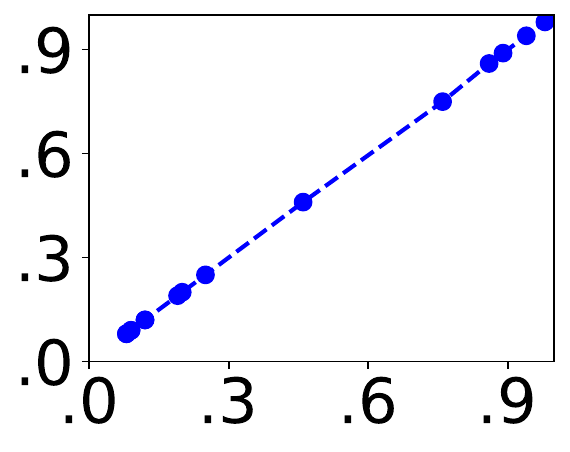}}
    \subfloat[Density]    {\includegraphics[height=2.5cm]{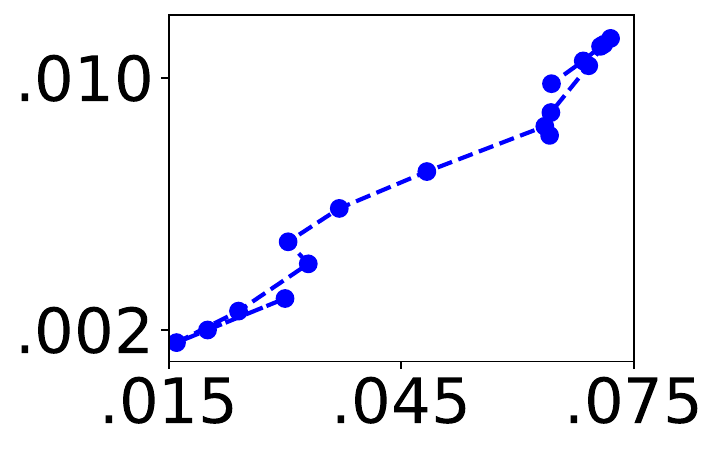}}
    \subfloat[Coverage]    {\includegraphics[height=2.5cm]{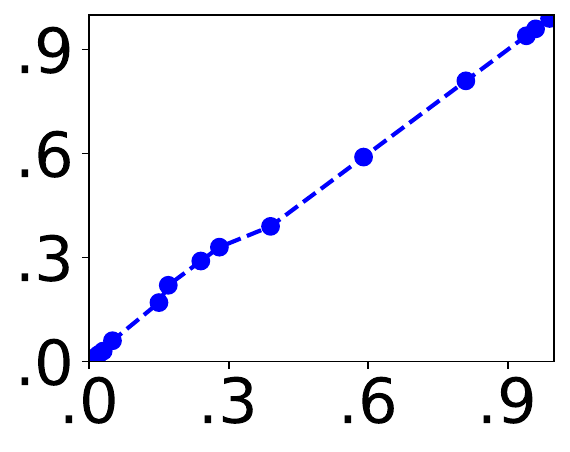}}
    \caption{Standard evaluation metrics in the distributed setting: X and Y axes show all-score and avg-score.}
    \label{fig:prdc}
\end{figure}

\minisection{Metrics for diagnosing fidelity and diversity} We extend our discussion to include precision \citep{sajjadi2018assessing,kynkaanniemi2019improved} and density \citep{naeem2020reliable}, which are proxies for sample fidelity. We also consider recall \citep{sajjadi2018assessing,kynkaanniemi2019improved} and coverage \citep{naeem2020reliable} to quantify sample diversity. For Precision/Recall evaluation, we utilized the official implementation with five clusters and assessed these metrics on a distributed CIFAR-10 dataset. For Density/Coverage evaluation, we adopted publicly available implementations. 

Our experiments involved ten clients, each assigned all samples from one distinct CIFAR-10 class. To simulate a progression of generators exhibiting varying degrees of diversity, we began with randomly selected samples and incrementally incorporated additional nearest-neighbor samples within the feature space. This approach produced datasets of increasing diversity levels. Feature extraction was performed using a pre-trained Inception-V3 network. As shown in \Cref{fig:prdc}, Precision and Recall metrics exhibited differing degrees of reliability when used to rank generative models in heterogeneous data scenarios. While recall measures (Recall-all and Recall-avg) consistently provided identical outcomes, precision results displayed notable variability. This variation underscores the challenges associated with relying solely on precision to rank models in distributed data contexts. Similarly, density-based metrics (Density-all and Density-avg) occasionally resulted in inconsistent rankings, whereas coverage-based metrics mirrored recall in providing stable and consistent model rankings across evaluations.

\minisection{Other metrics} Exploring distributed evaluation methods is similarly valuable for other metrics, though comprehensively addressing all existing metrics exceeds the scope of a single study and necessitates multiple dedicated investigations. Nonetheless, to enhance the breadth of our analysis, evaluations involving additional prominent metrics, including General 1-Wasserstein-Distance \citep{gulrajani2017improved,arjovsky2017wasserstein} and Log-likelihood, are included in \Cref{subsec:1Wasserstein} and \Cref{subsec:lls} in the Appendix. A key insight from our distributed evaluation of generative models is recognizing the heightened difficulty in accurately estimating the underlying real-data distribution, a task that is significantly more challenging than in centralized settings.

\subsection{Distributed Fine-tuning via MMD Loss}
\label{sec:mmdfinetune}

The monotonic nature of the KD-avg and KD-all scores motivates their application in distributed optimization for federated learning, where preserving client data privacy is essential \citep{kairouz2021advances}. Here, we explore a distributed fine-tuning setting with data heterogeneity, involving multiple clients collaboratively optimizing a StyleGAN2 generator that was pre-trained on the FFHQ dataset. Importantly, clients avoid sharing their data, and the objective of this distributed optimization is to adjust the generator’s distribution to better align with the client data.

\begin{figure}[ht]
    \centering
    \subfloat[Client 1 w/ glasses]{\includegraphics[width=.3\linewidth]{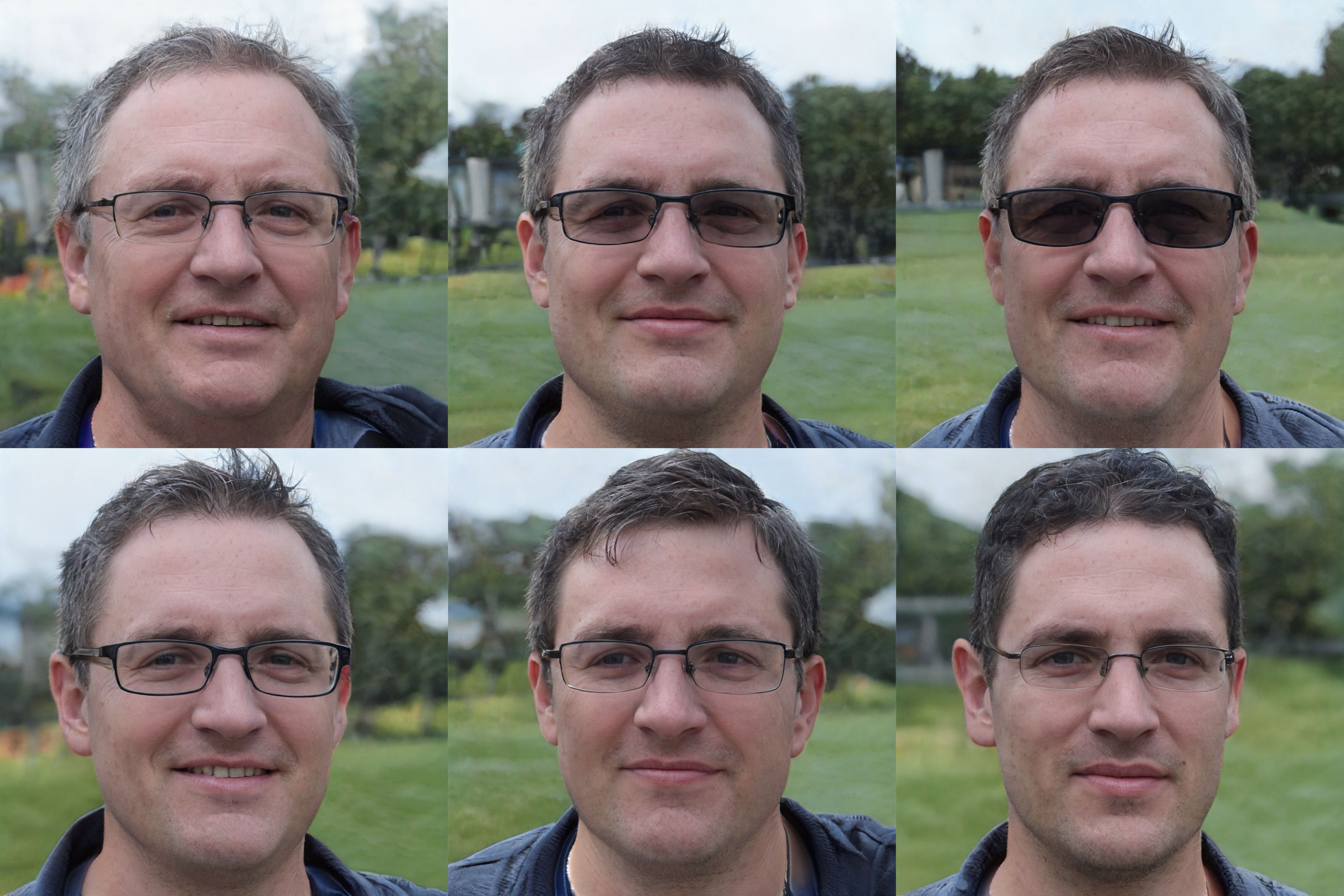}}
    \hfill
    \subfloat[Client 2 w/ glasses]{\includegraphics[width=.3\linewidth]{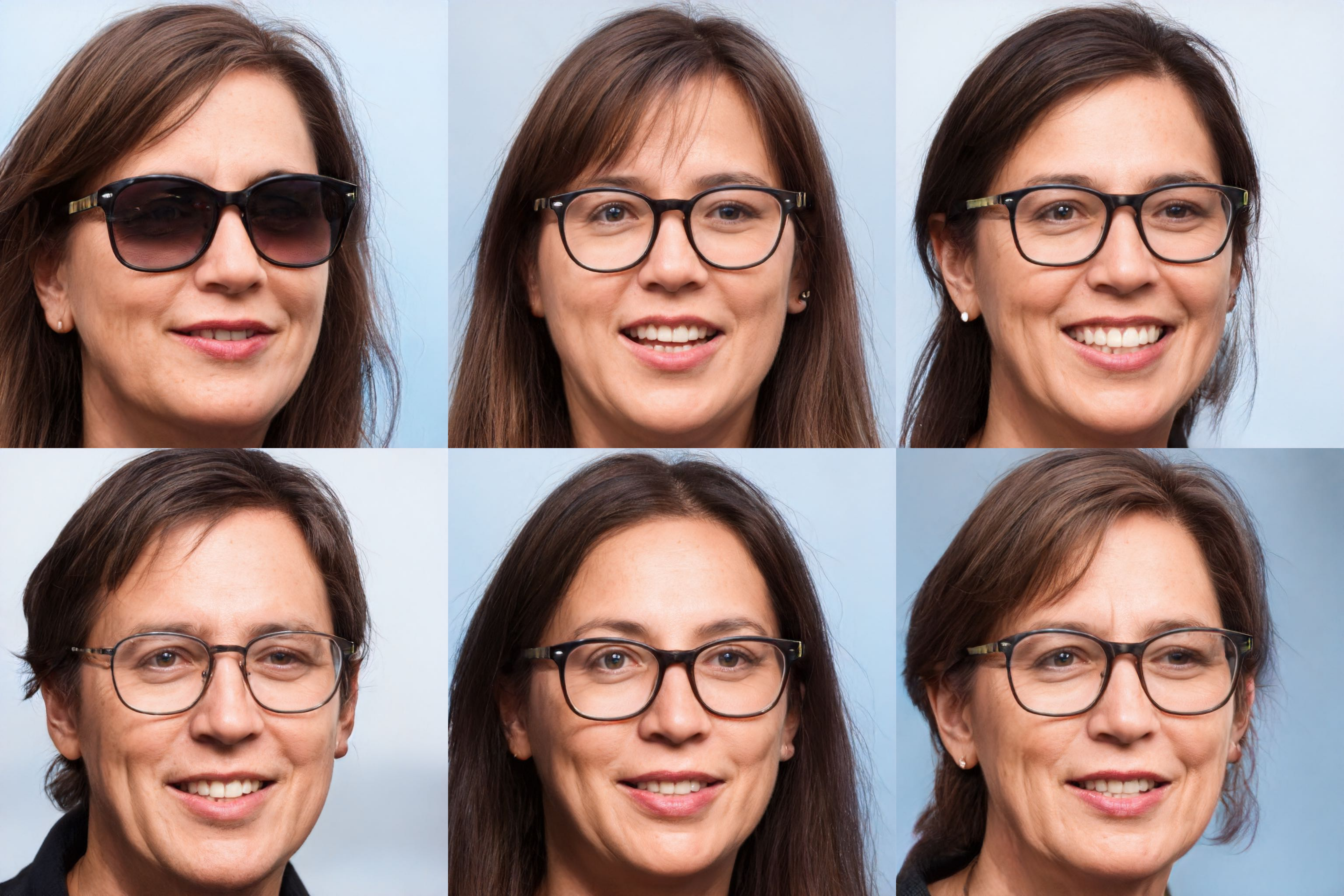}}
    \hfill
    \subfloat[Client 3 w/ glasses]{\includegraphics[width=.3\linewidth]{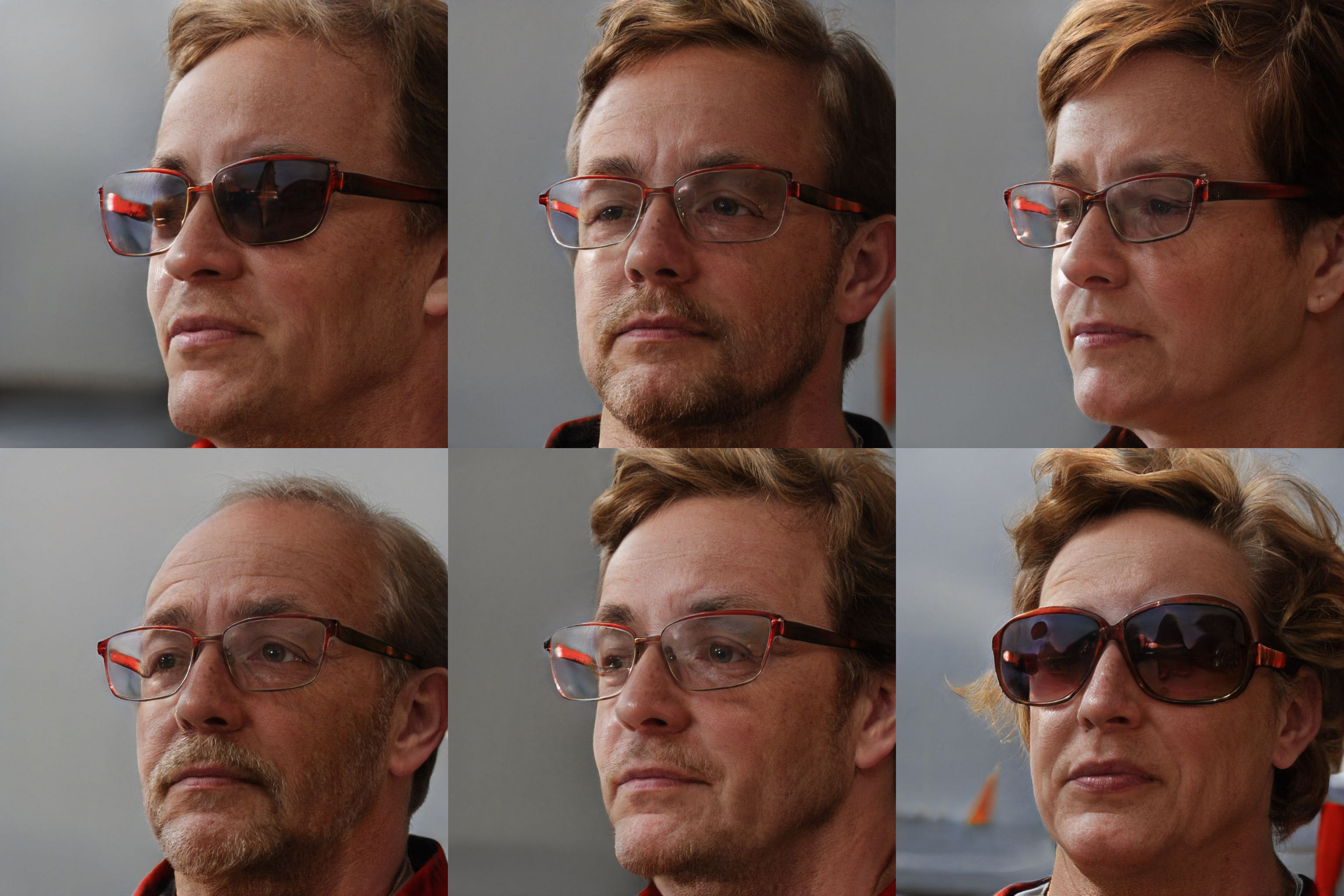}}\\
    \subfloat[Client 4 w/ head accessories]{\includegraphics[width=.3\linewidth]{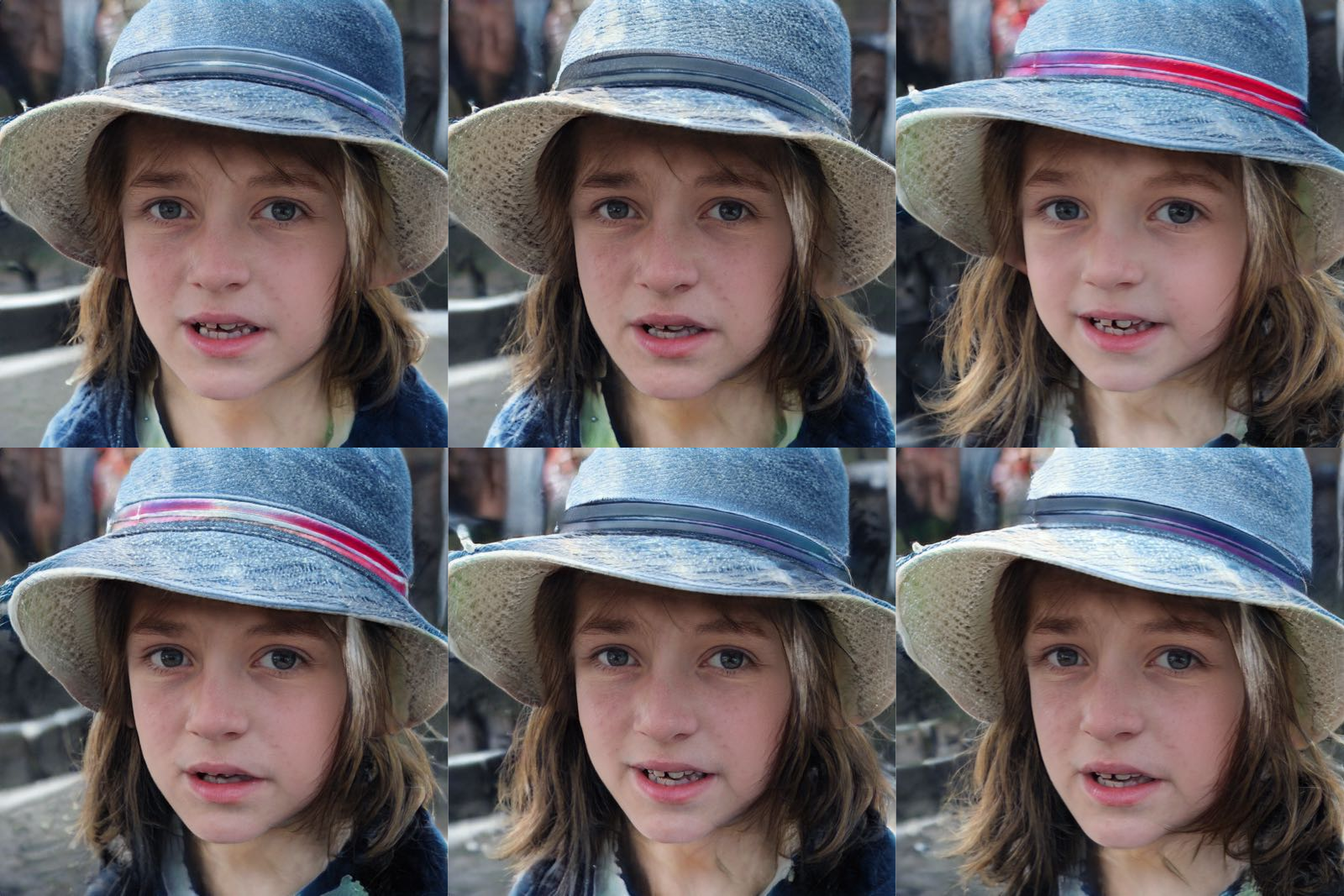}}
    \hfill
    \subfloat[Client 5 w/ head accessories]{\includegraphics[width=.3\linewidth]{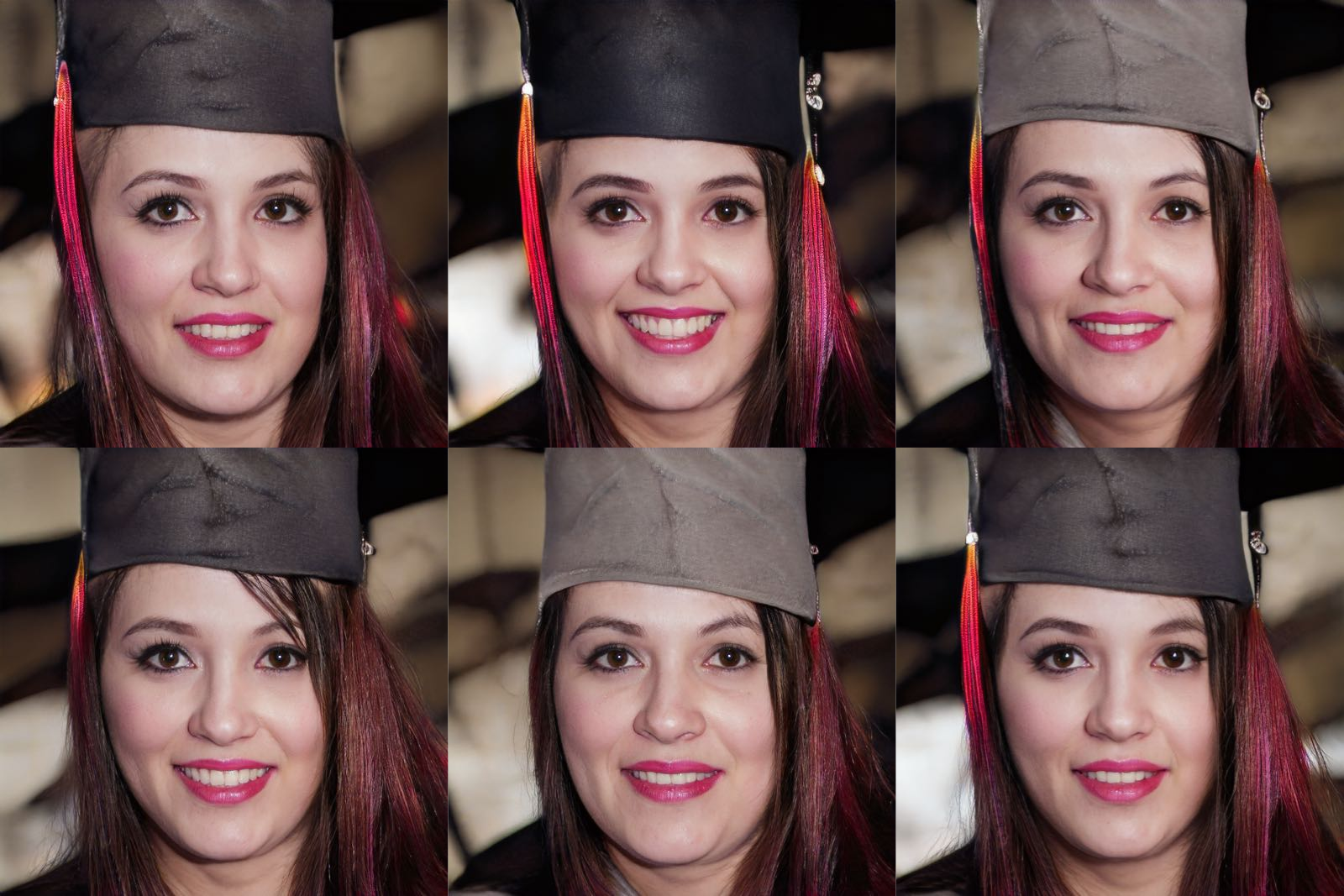}}
    \hfill
    \subfloat[Client 6 w/ head accessories]{\includegraphics[width=.3\linewidth]{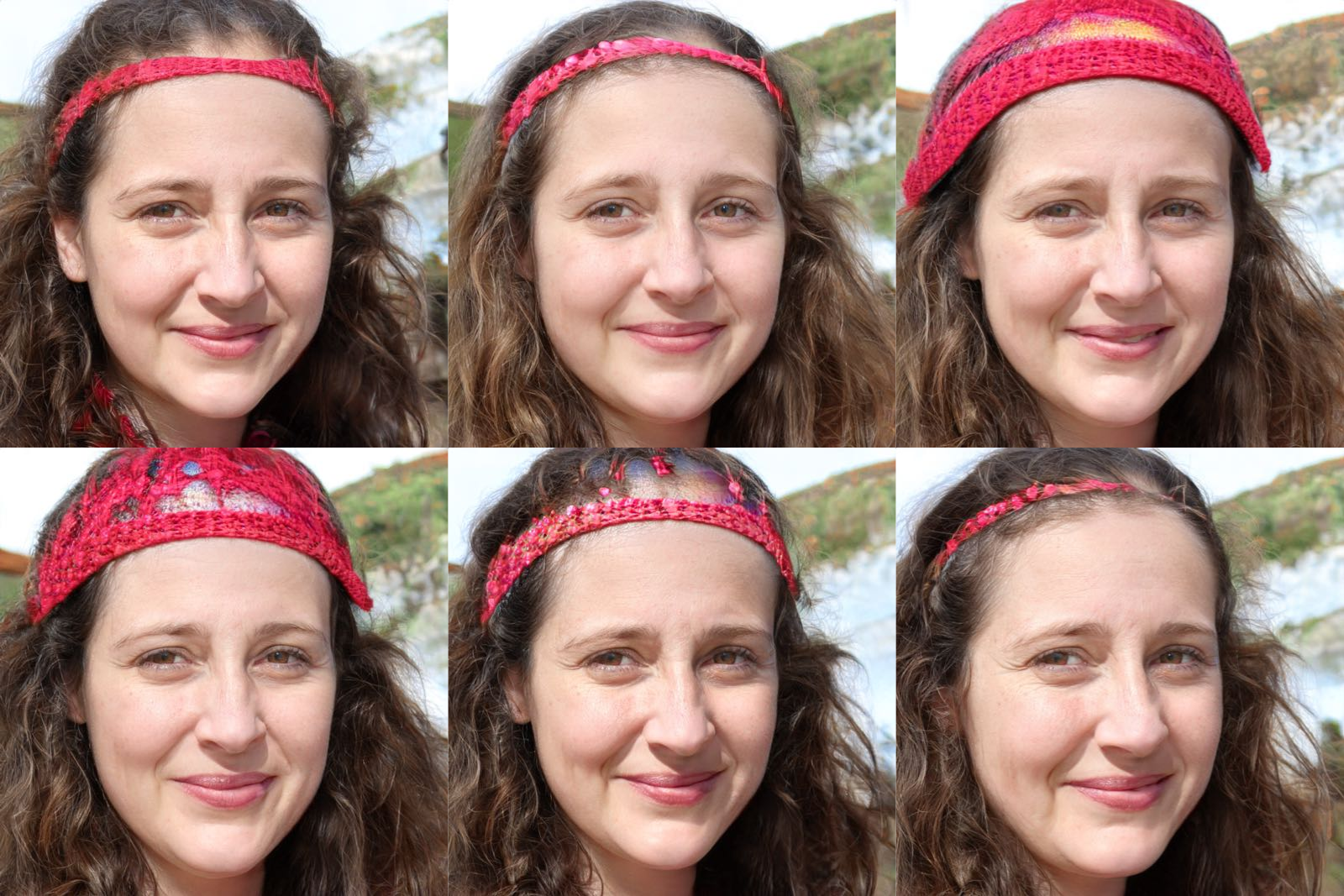}}
    \caption{Illustration of random samples from clients used in distributed optimization.}
    \label{fig:glassesclient}
\end{figure}

In our scenario, we employ twelve clients, each containing images of individuals wearing glasses generated via the previously introduced truncation method. A subset of client images is illustrated in \Cref{fig:glassesclient}. Given that FFHQ contains only a limited subset of individuals with glasses, the fine-tuned generator is expected to produce people with glasses more frequently after training. We follow the standard protocols of StyleGAN2-ADA, incorporating two primary loss functions. The first is the standard GAN training loss, applied to both the generated samples and the original FFHQ dataset. The second is the MMD$^2$ distance, computed using a polynomial kernel of order $3$, applied between the generated samples and client samples. To balance these losses, we set the weight factor for the MMD$^2$ loss to 5. Training is conducted for 1M images. Additional hyperparameters follow the "paper256" configuration from the official implementation.

\begin{figure}[t]
        \centering
        \subfloat[Fine-tuned on clients with people wearing glasses.]{
            \includegraphics[width=0.85\linewidth]{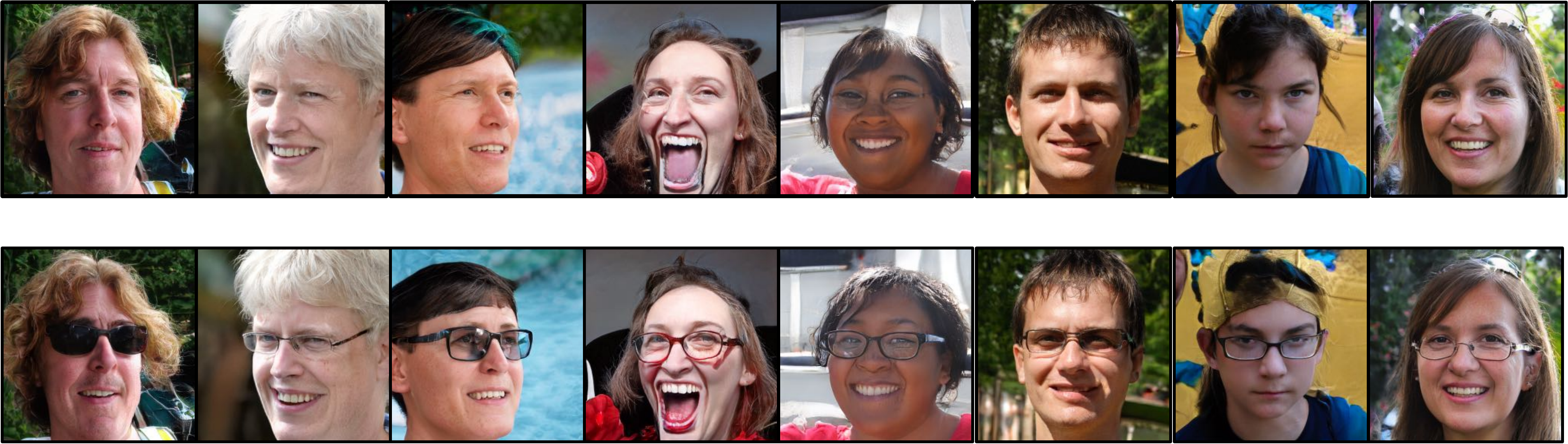}
            \label{fig:glasses}
        }\\[1ex]
        \subfloat[Fine-tuned on clients with people wearing head accessories.]{
            \includegraphics[width=0.85\linewidth]{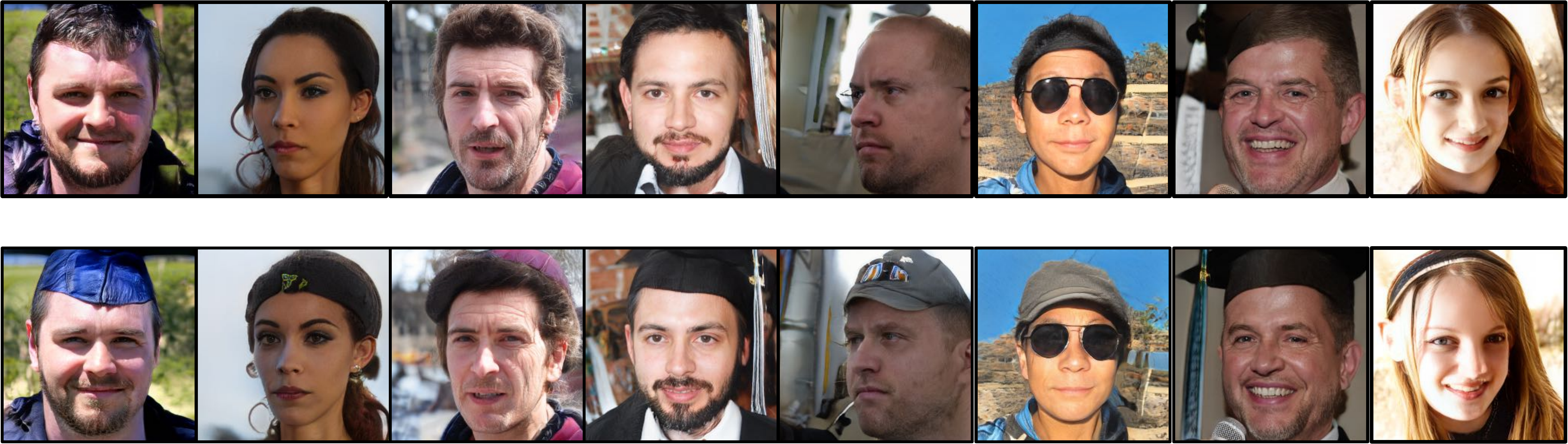}
            \label{fig:hat}
        }
        \caption{Effect of fine-tuning with MMD loss. Top: images from the pretrained model. Bottom: same seed after fine-tuning.}

\end{figure}

In \Cref{fig:glasses}, we present a comparison of images generated by the fine-tuned generator versus those produced by the original model. After fine-tuning, there is a notable increase in the frequency of generated individuals wearing glasses, rising from 20\% to 27\% based on statistics from a set of randomly sampled images. We also provide results on similar settings in \Cref{fig:hat}, where clients hold images of people wearing head accessories such as hats or hairbands. In this case, we also observed that the frequency of people with head accessories increases. This result underscores the feasibility of optimizing the MMD loss function in a privacy-preserving distributed learning setting, leveraging the monotonic nature of the KD distance metric. 

\section{Conclusion}
In this paper, we studied the problem of ranking a group of candidate generative models in distributed settings, where clients’ reference data may follow heterogeneous distributions. We showed that for the kernel distance (KD) evaluation metric, averaging the KD scores across clients yields a ranking that is equivalent to centralized evaluation, and this result holds for every kernel function. We also presented numerical results for several other evaluation metrics, where, except for the Recall score, the ranking consistency does not universally hold. Furthermore, we extended our analysis to show that distributed fine-tuning via a KD-based loss can serve as a viable approach under privacy restrictions.
A relevant direction for future work is to incorporate differential privacy techniques to analyze and mitigate potential data leakage in distributed ranking.
Another direction is to study the ranking of generative models under privacy and communication constraints in sharing candidate models.



\bibliography{main}
\bibliographystyle{tmlr}

\appendix
\clearpage
\section{Extended Evaluation Metrics}
\label{sec:extended_metrics}
\subsection{General 1-Wasserstein-Distance}
\label{subsec:1Wasserstein}

\begin{figure}[ht!]
    \centering
    \includegraphics[width=.48\textwidth]{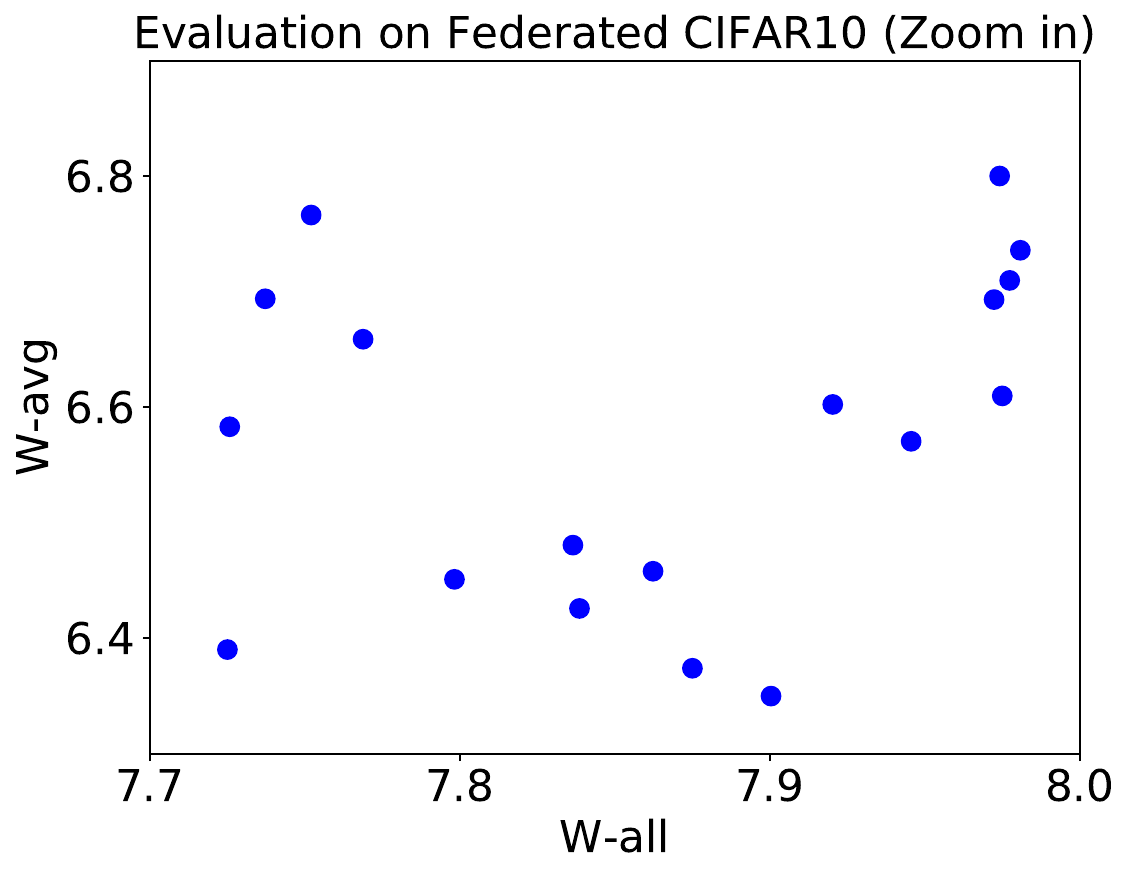}
    \caption{Evaluation with 1-Wasserstein-Distance on Federated CIFAR10.}
    \label{fig:1w}
\end{figure}

Let $\mathbb{P}_g$ and $\mathbb{P}_t$ represent the distributions of the generated set and training set. The Wasserstein-1 distance between $\mathbb{P}_g$ and $\mathbb{P}_t$ is

\begin{equation}
    W(\mathbb{P}_g,\mathbb{P}_t) = \inf_{\lambda\in\Pi(\mathbb{P}_g,\mathbb{P}_t)}\mathbb{E}_{(x,y)\sim\lambda}[\|x-y\|],
\end{equation}
where $\Pi(\mathbb{P}_g,\mathbb{P}_t)$ denotes the set of all joint distributions $\lambda(x,y)$ whose marginal distributions are respectively $\mathbb{P}_g$ and $\mathbb{P}_t$. However, the direct estimation of $W(\mathbb{P}_g,\mathbb{P}_t)$ is highly intractable. On the other hand, the Kantorovich-Rubinstein duality~\citep{villani2009optimal} gives

\begin{equation}
    W(\mathbb{P}_g,\mathbb{P}_t) = \sup_{\|f\|_L\leq 1}\mathbb{E}_{x\sim \mathbb{P}_g}[f(x)]-\mathbb{E}_{x\sim \mathbb{P}_t}[f(x)],
\label{eq:KRD}
\end{equation}
where the supremum is over all the 1-Lipschitz functions $f: \mathcal{R}^n\rightarrow \mathcal{R}$. Therefore, if we have a parameterized family of functions $\{f_\theta\}_{\theta\in\Theta}$ that are 1-Lipschitz, we could consider solving
\begin{equation}
    \mathop{max}\limits_{\theta\in\Theta} \mathbb{E}_{x\sim \mathbb{P}_g}[f_\theta(x)]-\mathbb{E}_{x\sim \mathbb{P}_t}[f_\theta(x)].
\end{equation}
To estimate the supremum of \Cref{eq:KRD}, we employ a family of non-linear neural networks $f_\theta$ that are built by repeatedly stacking fully connected layers, spectral normalization, and ReLU activation layers. There are three repeated blocks in the network $f_\theta$, and the last block does not have ReLU. The features are extracted by a pre-trained Inception-V3 network. By optimizing the parameters in $f_\theta$ to maximize $\mathbb{E}_{x\sim \mathbb{P}_g}[f_\theta(x)]-\mathbb{E}_{x\sim \mathbb{P}_t}[f_\theta(x)]$ over $\mathbb{P}_g$ and $\mathbb{P}_t$, we obtain an estimate of $W(\mathbb{P}_g,\mathbb{P}_t)$.
Similarly, we can also define the average score W-avg and collective-data-based score W-all under the distributed learning setting.
Similar to the CIFAR100 experiment in the main body of the paper, we extracted samples from each single class of CIFAR100 and evaluated these samples on the federated CIFAR10 dataset. We illustrate a subset of W-avg~/~W-all pairs in \Cref{fig:1w}. According to the experimental results, the general 1-Wasserstein-Distance evaluation metric also shows inconsistent behavior in distributed evaluation settings.

\subsection{Log-Likelihood Score}
\label{subsec:lls}

\begin{figure*}[th!]
    \centering
    \subfloat[]{ \includegraphics[height=3.4cm]{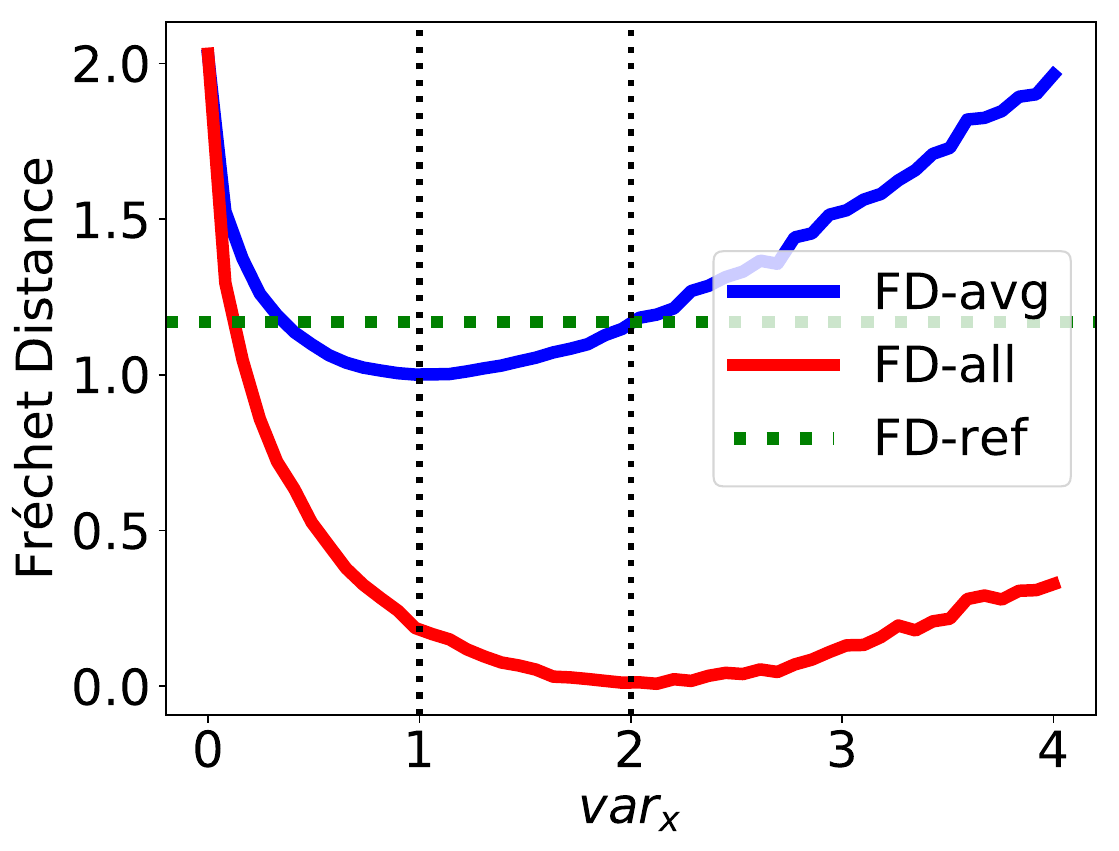} }
    \subfloat[]{ \includegraphics[height=3.4cm]{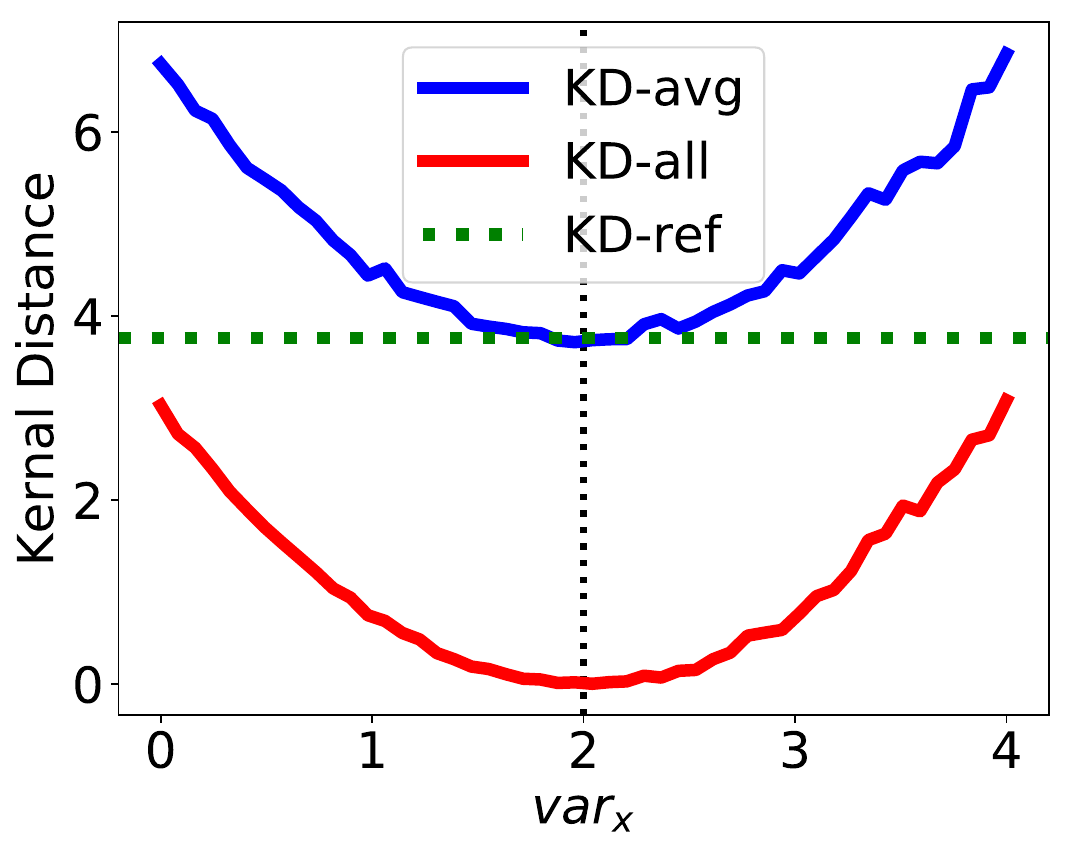} }
    \subfloat[]{ \includegraphics[height=3.4cm]{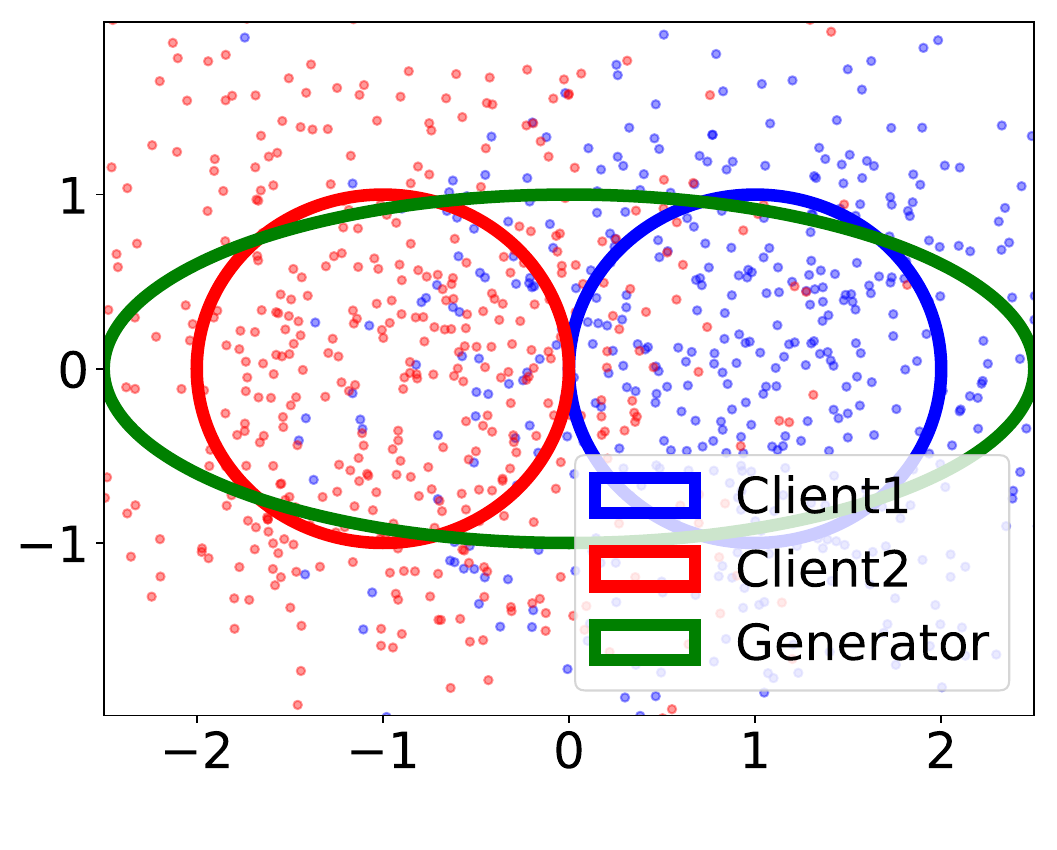} }
    \caption{Experimental results of Gaussian mixture dataset. (a): The optimal $\mathrm{var}_x$ parameters are different under FD-avg and FD-all evaluations. (b): Distance between KD-avg and KD-all remains the same. (c): the clients' and generator's samples. }
    \label{fig:toy}
\end{figure*}

\minisection{Setup} Our experimental setup involves two clients, denoted by $C_1$ and $C_2$. $C_1$ possesses a dataset consisting of 50,000 samples drawn from the Gaussian distribution $\mathcal{N}([1,0]^\top, \bm{\Sigma})$, while $C_2$ holds a dataset with 50,000 samples drawn from $\mathcal{N}([-1,0]^\top, \bm{\Sigma})$, where $\bm{\Sigma} = \text{diag}([1,1]^T)$.
We introduce a generator, denoted as $G_{\mathrm{var}_x}$, which is parameterized by $\mathrm{var}_x$. $\mathrm{var}_x$ regulates the variance of the generator along the X-axis. Specifically, $G_{\mathrm{var}_x}$ generates 50,000 data points following a Gaussian distribution $\mathcal{N}([0,0]^T, \bm{\Sigma_G})$, where $\bm{\Sigma_G} = \text{diag}([\mathrm{var}_x, 1]^T)$. The relationship between the two clients and the generator is visually depicted in \Cref{fig:toy}.
Additionally, we introduce an "ideal estimator" denoted as $\hat{E} = C_1 \cup C_2$. This ideal estimator possesses the unique ability to replicate the distribution of the training dataset perfectly. We employ the ideal estimator as a reference for our analysis.

\begin{figure}[th]
    \centering
    \subfloat[$G_{var_x}$]{\includegraphics[width=.5\linewidth]{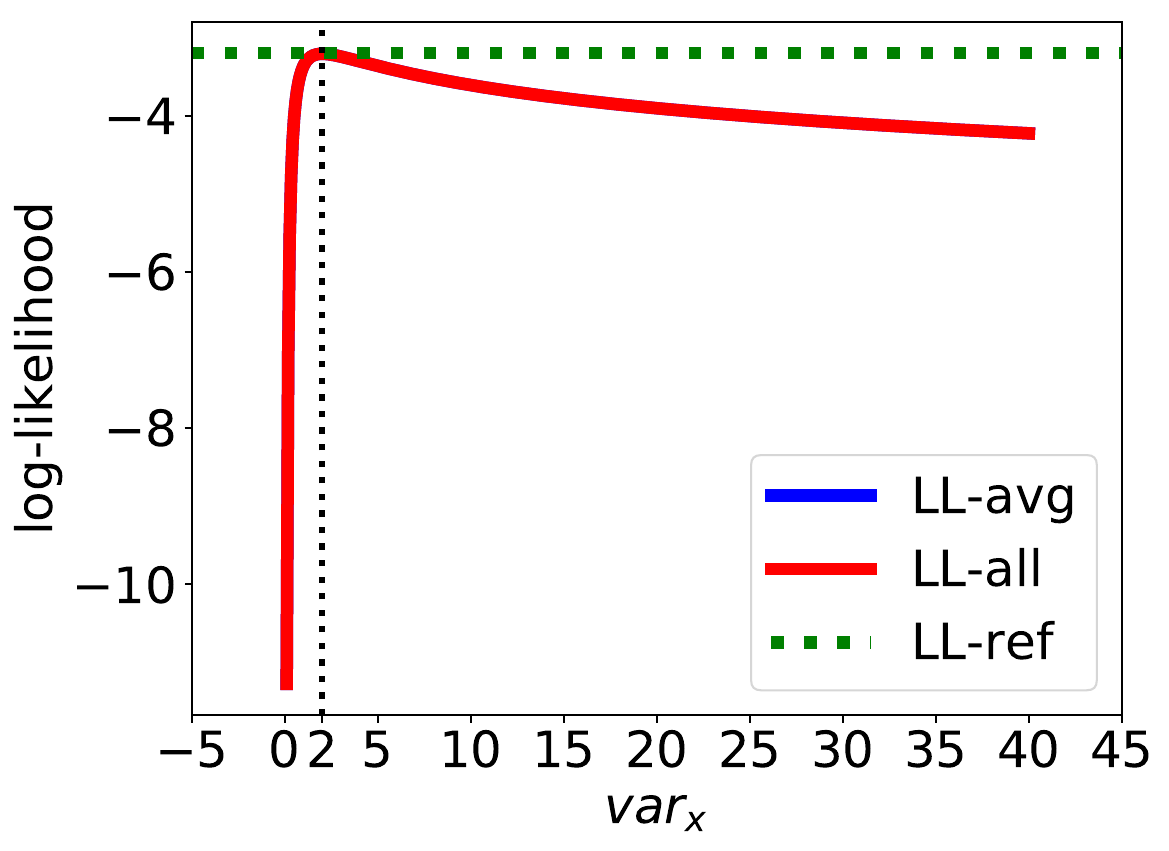}
    \label{fig:gvar}}
    \subfloat[$G_{mean_x}$]{\includegraphics[width=.5\linewidth]{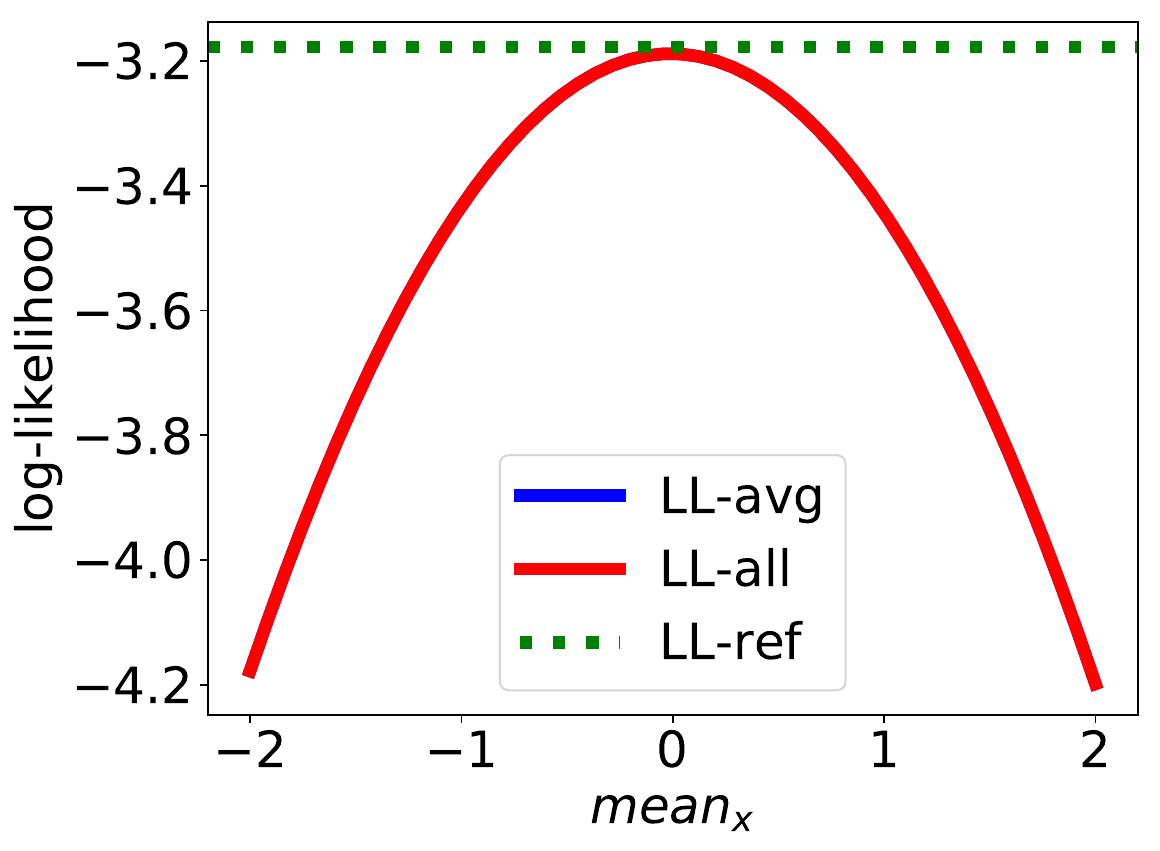}
    \label{fig:gmean}}
    \caption{Evaluation of synthetic Gaussian data with the aggregate log-likelihood scores.}
\end{figure}

\minisection{The Log-Likelihood Score} We evaluated the synthetic Gaussian mixture dataset with the standard log-likelihood (LL) score. In this experiment, we note that we have access to the probability density functions (PDF) of the simulated generator. We utilized the generator $G_{var_x}$ described in the main text and performed the evaluation over the parameter $\mathrm{var}_x$ in the range $[0,40]$. As can be shown in the general case, LL-avg and LL-all led to the same value for every evaluated model. As shown in \Cref{fig:gvar}, they reached their maximum value at $\mathrm{var}_x=2$. On the other hand, we set a new generator $G_{mean_x}$ generating samples according to $\mathcal{N}([\mathrm{mean}_x,0]^\top,\bm{\Sigma})$, where $\bm{\Sigma} = \text{diag}([2, 1]^T)$. We gradually increased $\mathrm{mean}_x$ from -2 to 2 and plotted LL-avg, LL-all, and LL-ref in \Cref{fig:gmean}.

\minisection{FD and KD} We measure the similarity between samples generated by clients and generators using the Fréchet distance (\text{\rm FD}), which follows from the Wasserstein-based definition of FD-all and FD-avg without the application of the pre-trained Inception network. We consider the aggregate scores $\text{\rm FD-avg}$ and $\text{\rm FD-all}$ as defined in \Cref{eq:avg} and \Cref{eq:all}. Note that the $\text{\rm FD-all}$ for the ideal estimator is zero and we use $\text{\rm FD-ref} = \frac{1}{2}\sum_{i=1}^2 \text{\rm FD}(\hat{E}, C_i)$ as a reference for $\text{\rm FD-avg}$. We also measure the Kernel distance (\text{\rm KD}), which follows the definition of KD-all and KD-avg without Inception network. $\text{\rm KD-ref}$ is defined for the kernel distance in a similar fashion to $\text{\rm FD-ref}$.

By increasing $\mathrm{var}_x$ from 0 to 4, we get a sequence of FD-avg~/~FD-all pairs and we plot them with the $\mathrm{var}_x$ in \Cref{fig:toy}. Our experimental results highlight the following conclusions. First, we observed that the minimum of $\text{\rm FD-all}$ occurs at $\mathrm{var}_x=2$, while that of FD-avg occurs at $\mathrm{var}_x = 1$, which indicates that the optimal solutions of $\mathrm{var}_x$ to minimize $\text{\rm FD-all}$ and $\text{\rm FD-avg}$ are inconsistent. In this case, $\text{\rm FD-all}$ and $\text{\rm FD-avg}$ lead to different rankings of the models with $\mathrm{var}_x=1$ and $\mathrm{var}_x=2$.
Additionally, we observed that, counterintuitively, the `ideal estimator' did not reach the minimum average of the Fréchet distances. The distance between KD-avg and KD-all remains the same with the change of $\mathrm{var}_x$ and both of which reach minimum at $\mathrm{var}_x=2$.
The toy experiment highlights how a covariance mismatch between clients and the collective dataset leads to inconsistent rankings according to aggregate Fréchet distances.

\section{Extended Results on KD and FD}
\label{sec:afhq}

\subsection{Results on Complex FFHQ Setting}

\begin{figure*}[tb]
    \centering
    \subfloat[FD-scores]{ \includegraphics[width=.28\textwidth]{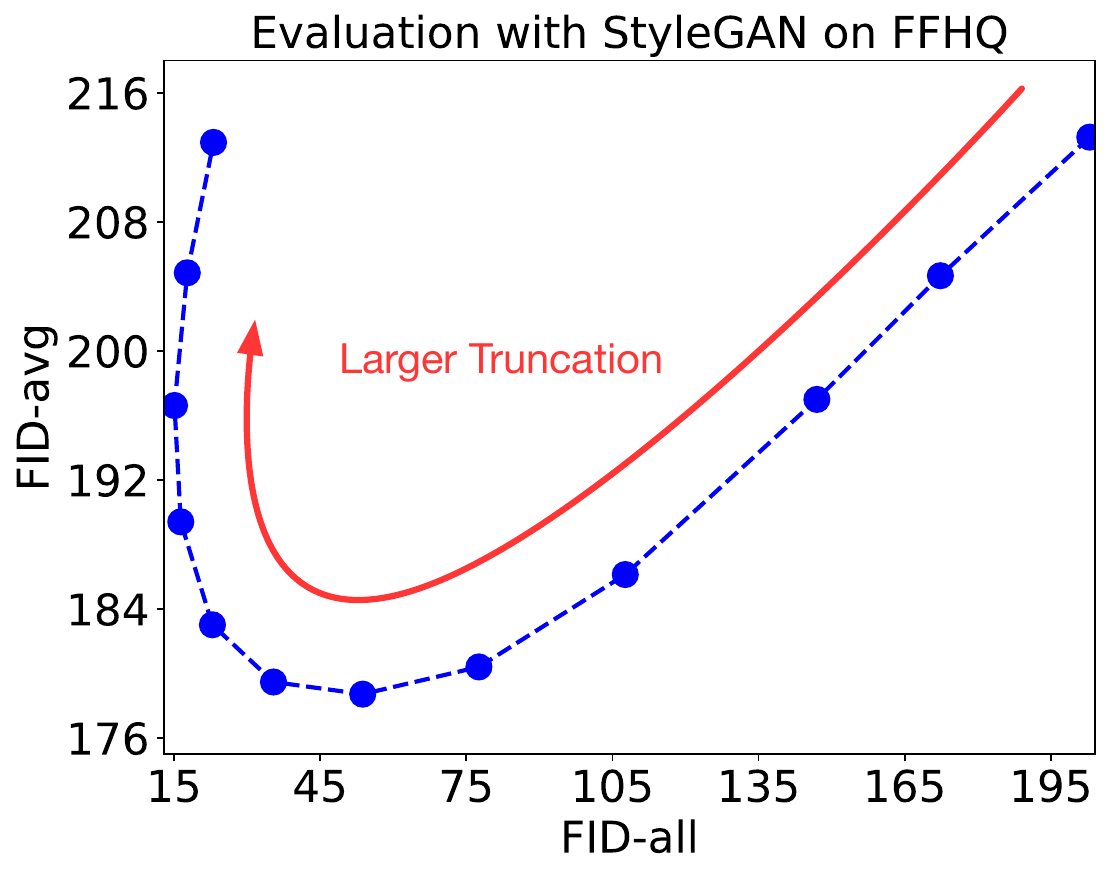} \label{fig:stylefid}}
    \hspace{.02\textwidth}
    \subfloat[KD-scores]{ \includegraphics[width=.28\textwidth]{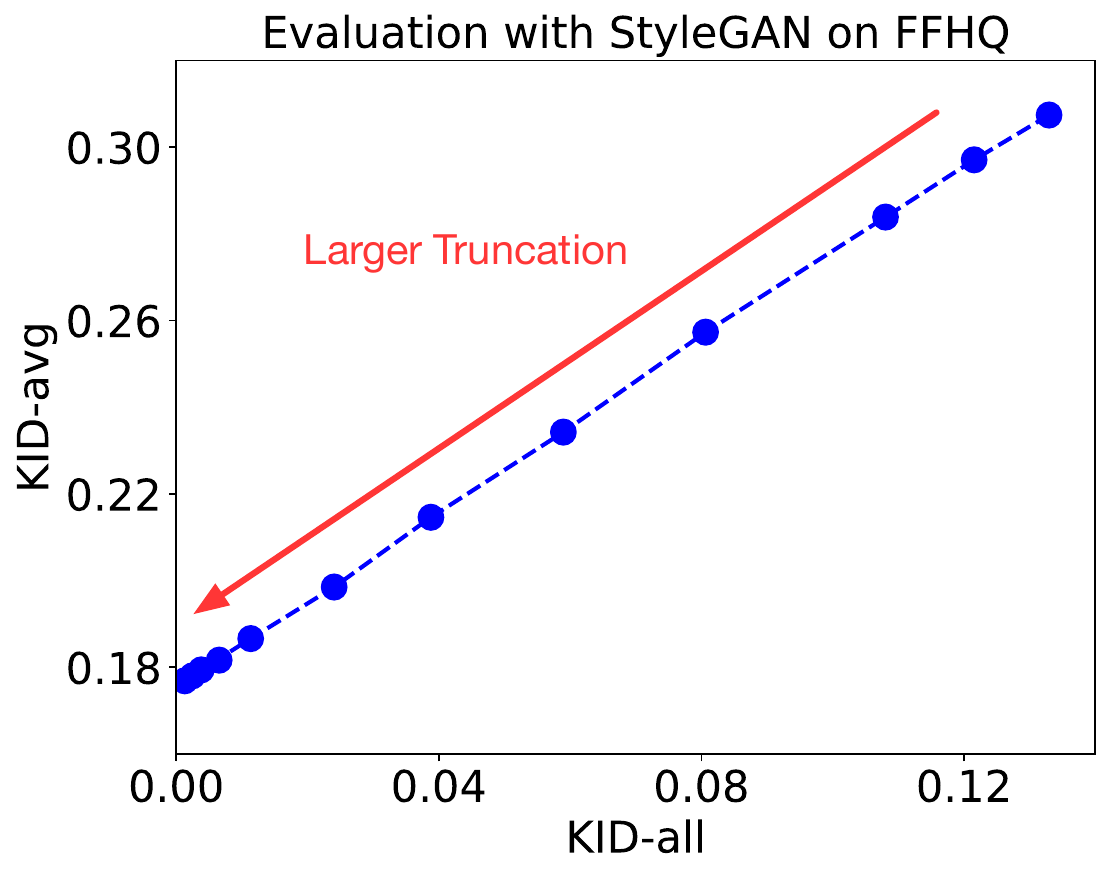} \label{fig:stylekid}} 
    \hspace{.02\textwidth}
    \subfloat[Rankings given by FD-scores]{ \includegraphics[width=.28\textwidth]{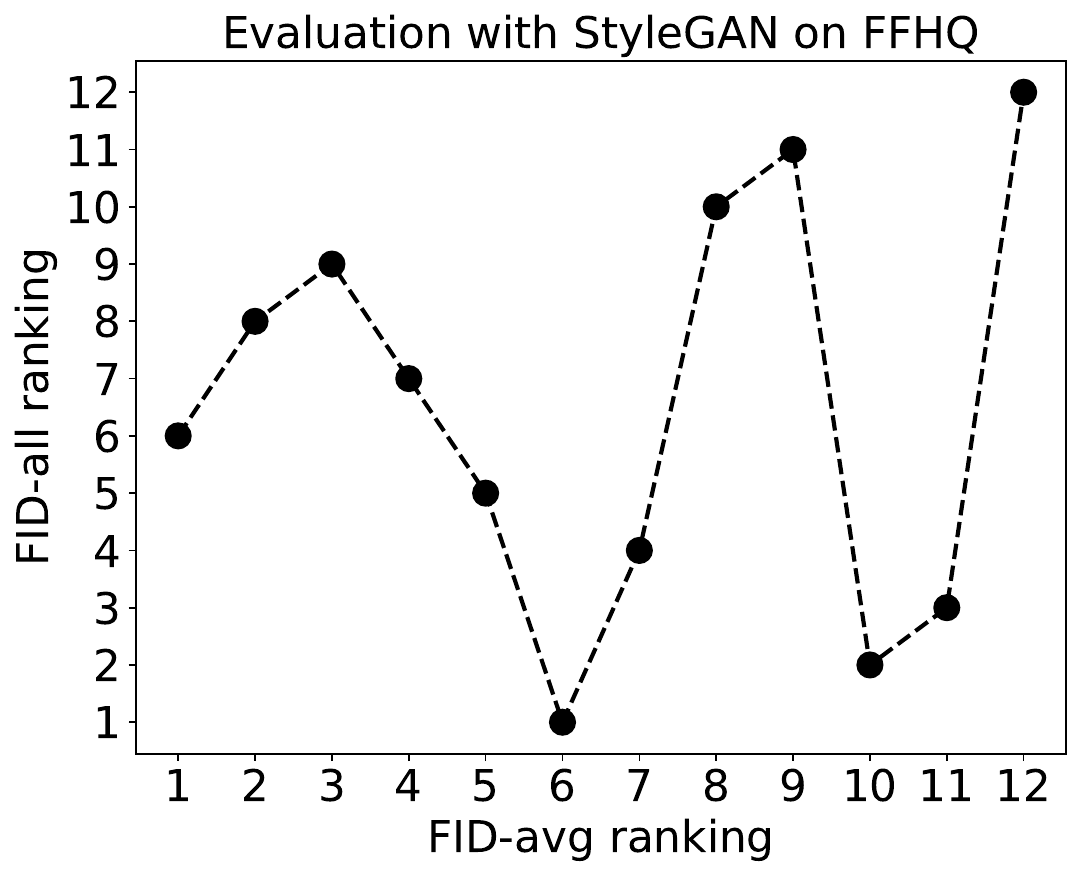}\label{fig:stylerank-2}}
    
    \caption{
        The results of evaluating generators in \Cref{fig:styleG} over clients in \Cref{fig:styleC}.
    }
   
\end{figure*}

\begin{figure}[th!]
    \centering
    \subfloat[Generator 1]{\includegraphics[width=.14\textwidth]{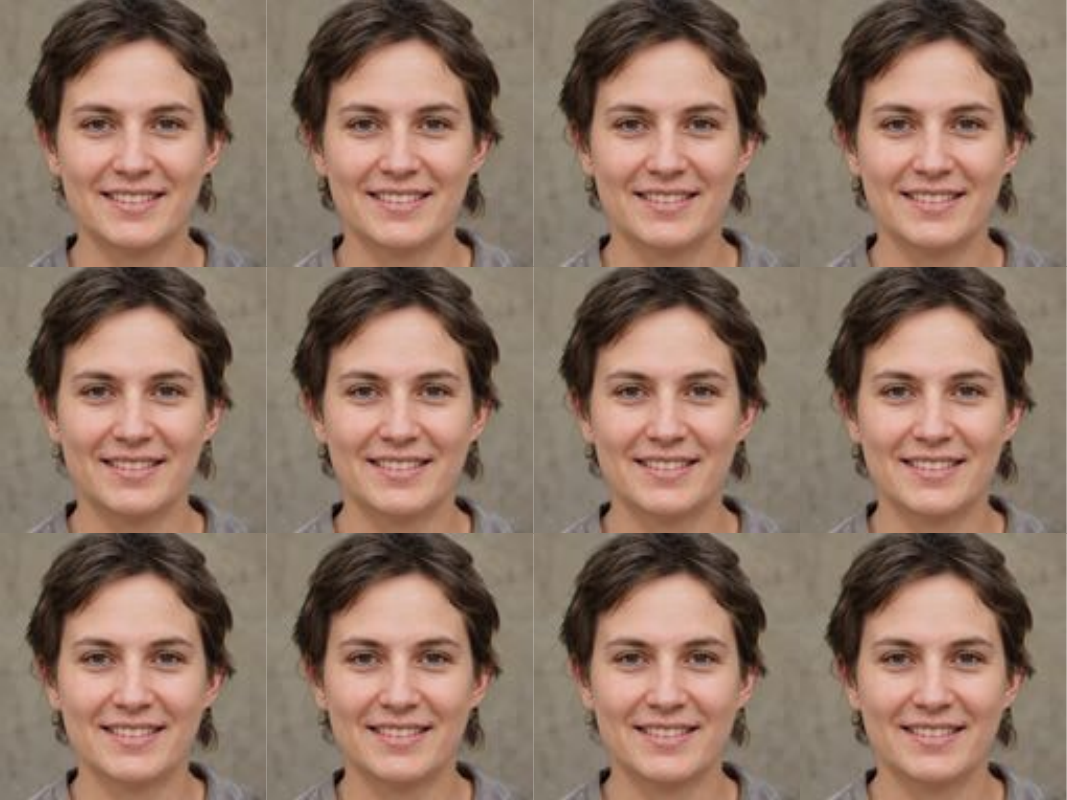}}
    \hspace{.01\textwidth}
    \subfloat[Generator 2]{\includegraphics[width=.14\textwidth]{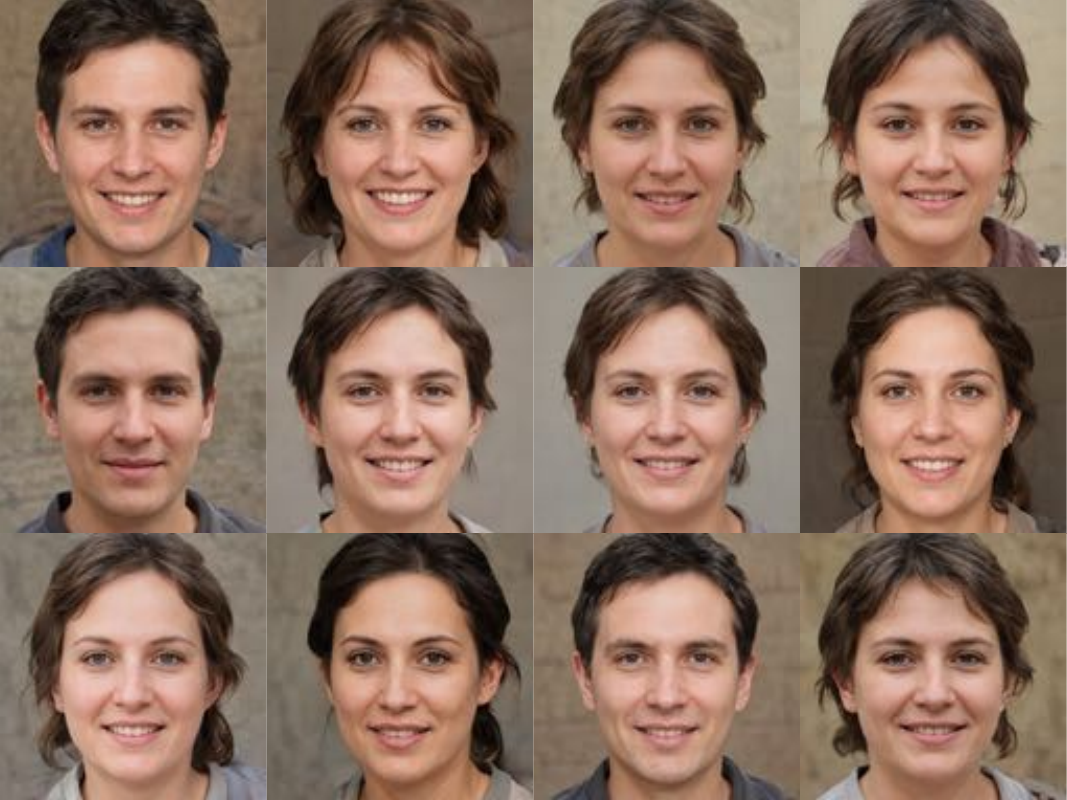}} 
    \hspace{.01\textwidth}
    \subfloat[Generator 3]{\includegraphics[width=.14\textwidth]{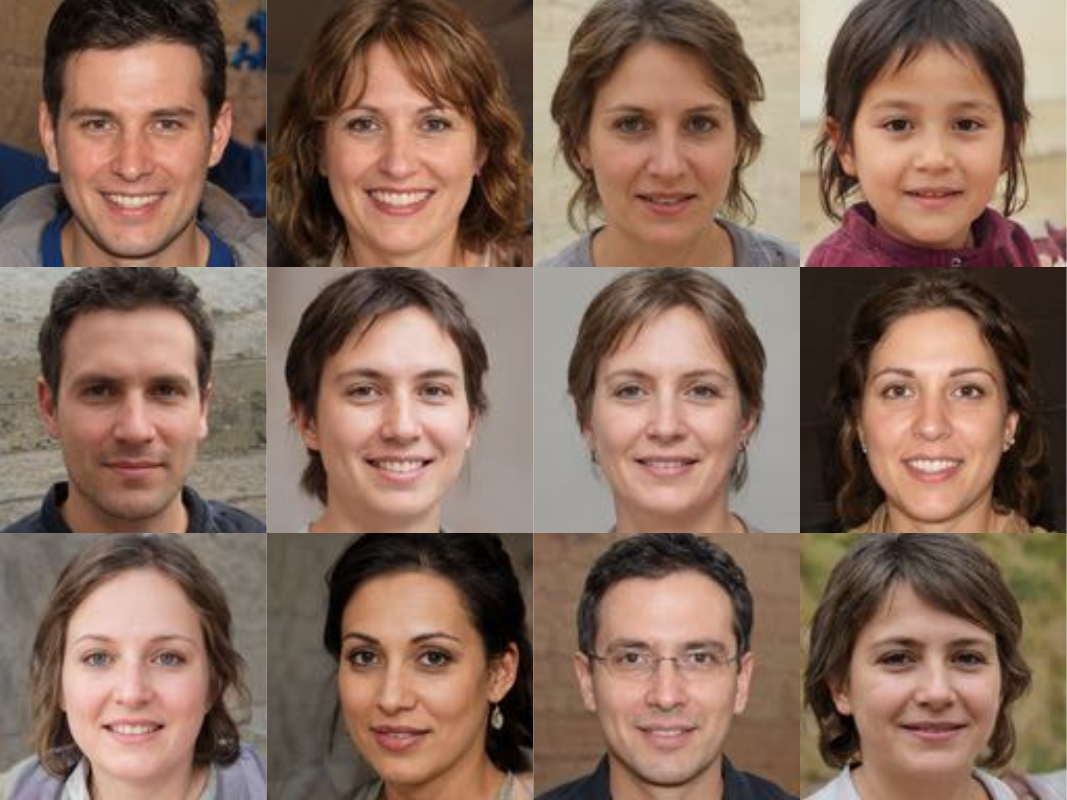}} 
    \hspace{.01\textwidth}
    \subfloat[Generator 4]{\includegraphics[width=.14\textwidth]{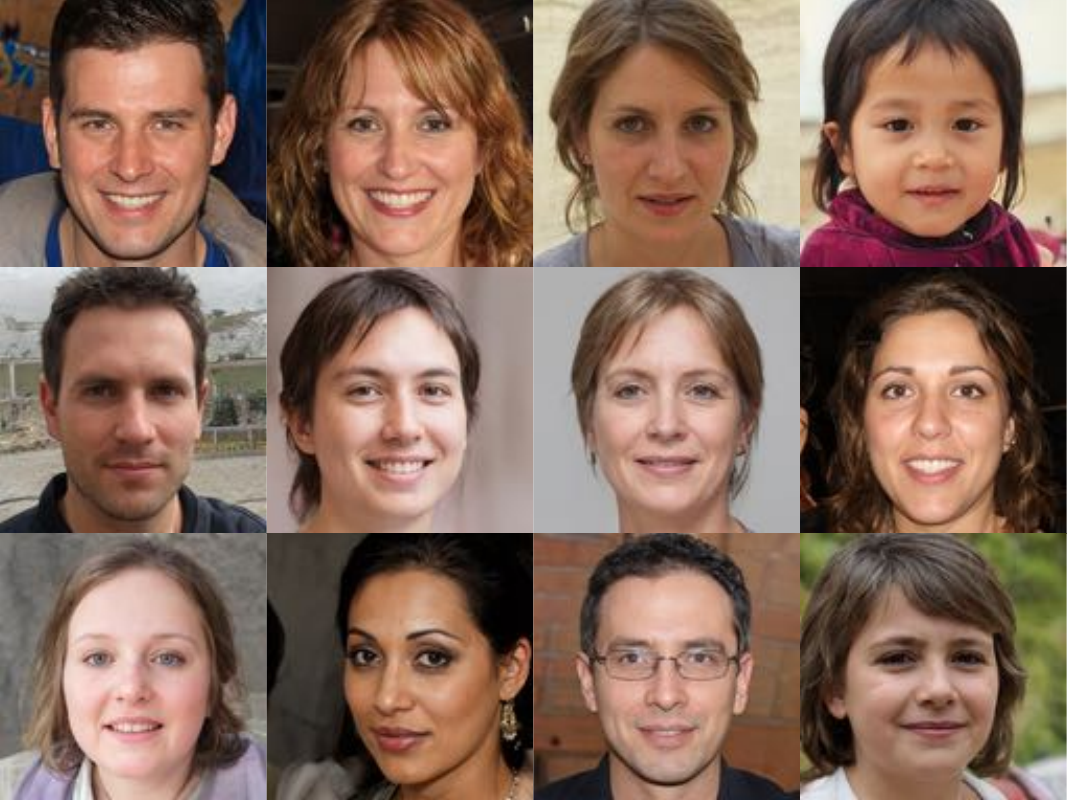}}
    \hspace{.01\textwidth}
    \subfloat[Generator 5]{\includegraphics[width=.14\textwidth]{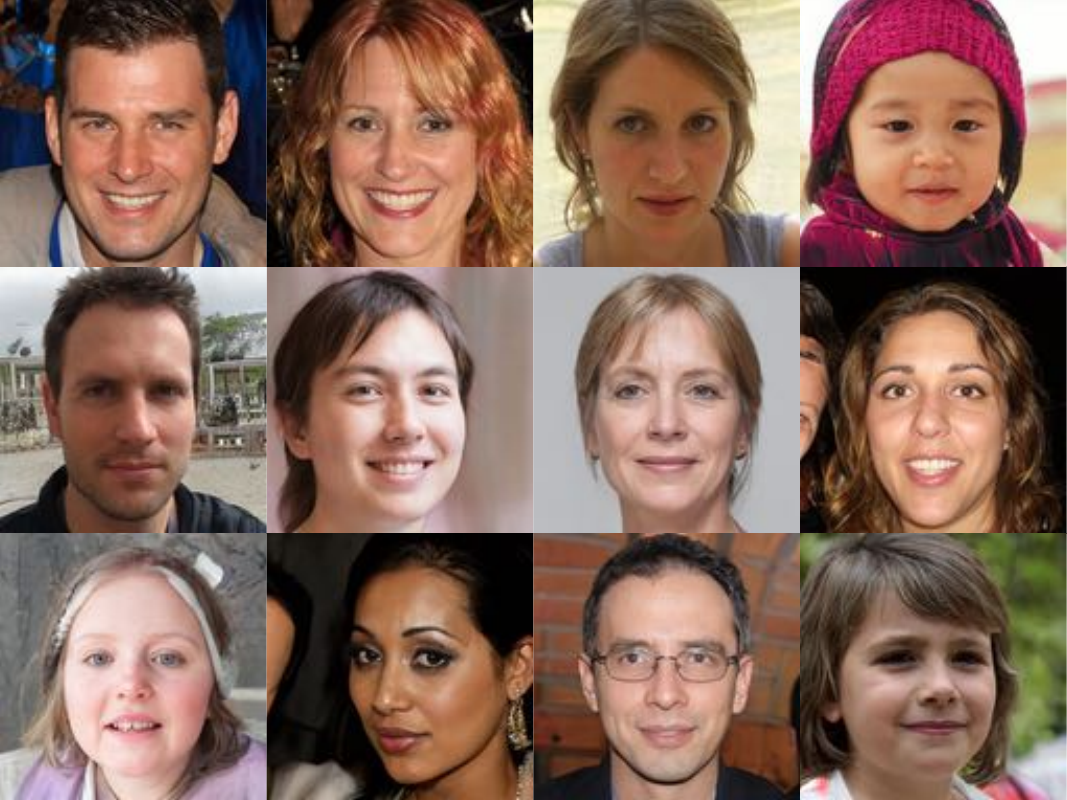}}
    \hspace{.01\textwidth}
    \subfloat[Generator 6]{\includegraphics[width=.14\textwidth]{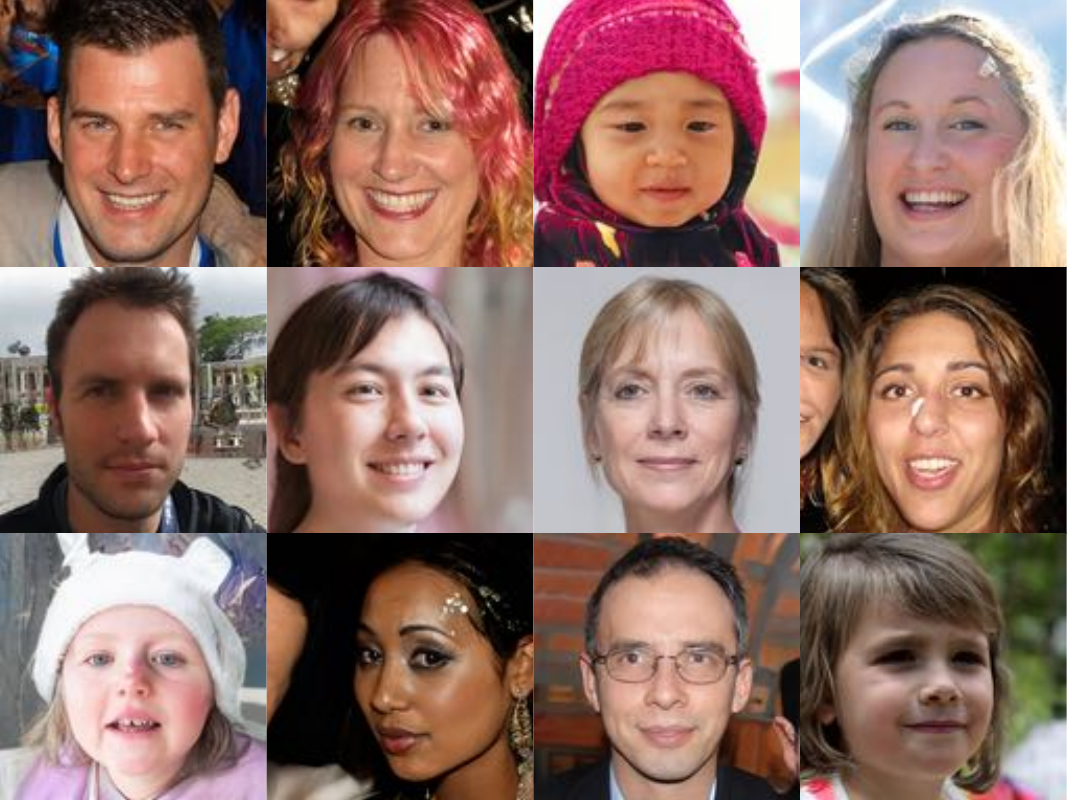}}\\
    \caption{Illustration of generators with different truncation factors. From (a) to (e), $\tau=0.01, 0.2, 0.4, 0.6, 0.8, 1.0$, where $\tau$ is the truncation parameter. }
    \label{fig:styleG}
\end{figure}

\begin{figure}[tb!]
    \centering
    \subfloat[Client 1]{\includegraphics[width=.14\textwidth]{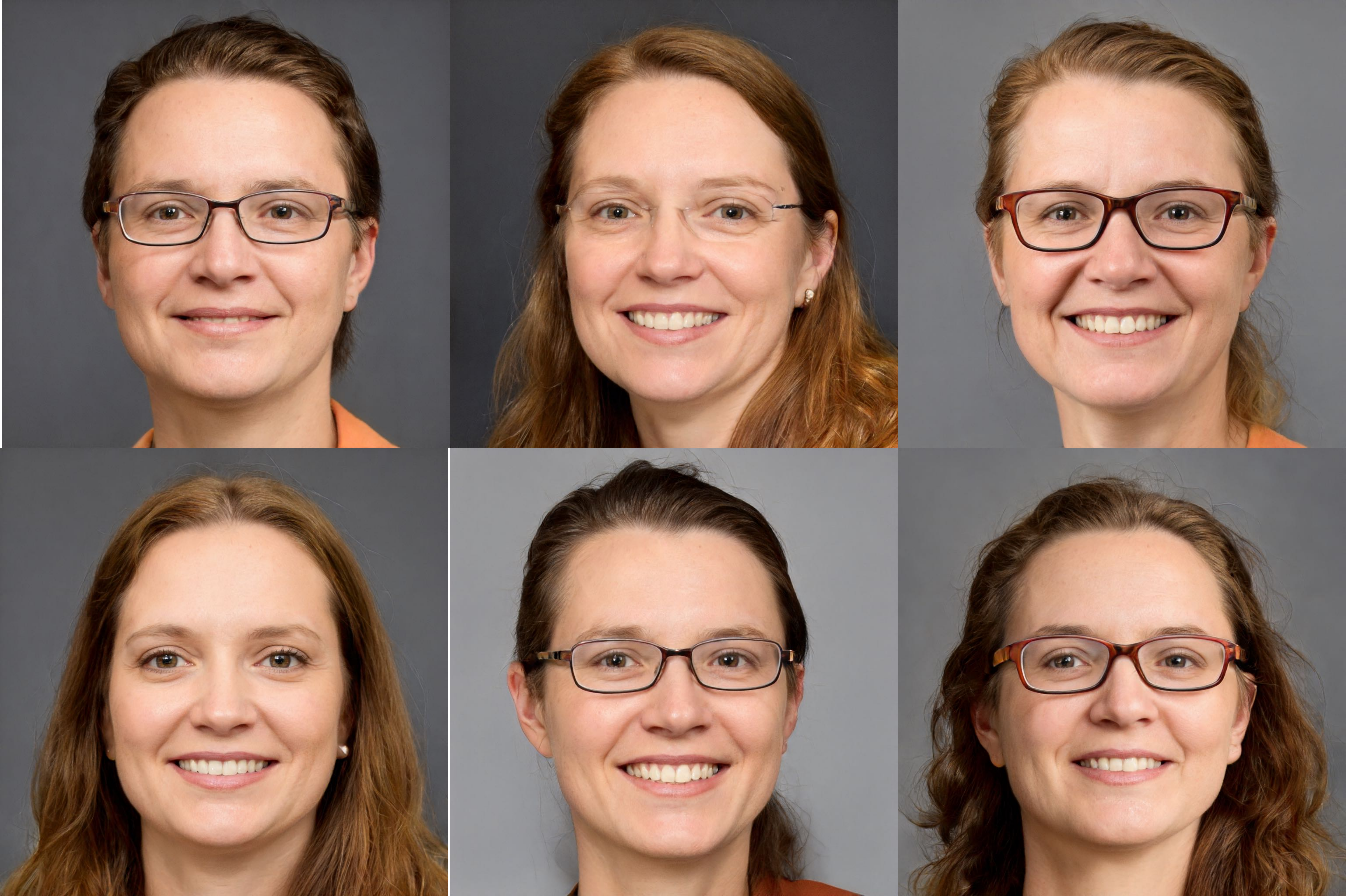}}
    \hspace{.01\textwidth}
    \subfloat[Client 2]{\includegraphics[width=.14\textwidth]{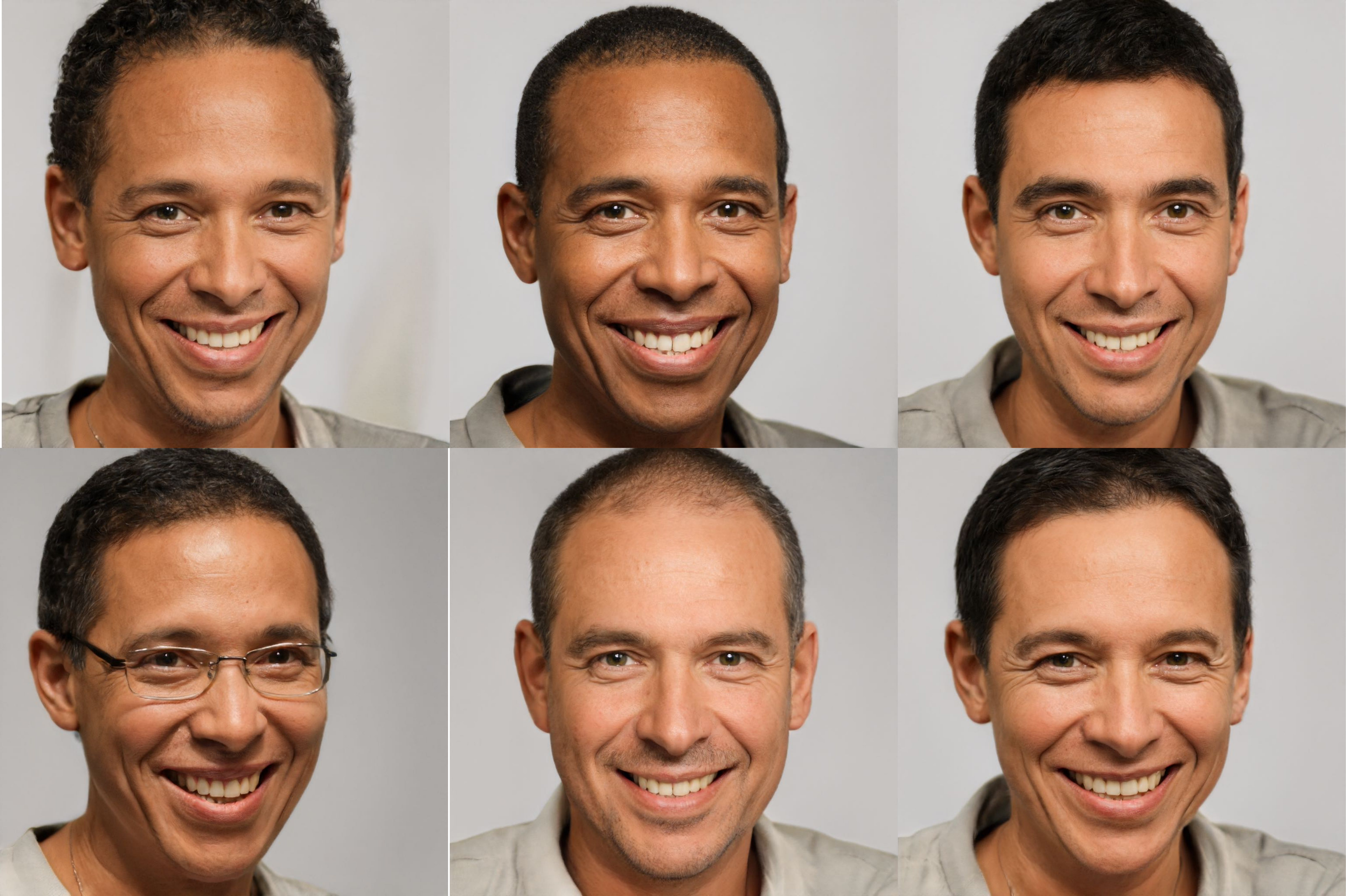}} 
    \hspace{.01\textwidth}
    \subfloat[Client 3]{\includegraphics[width=.14\textwidth]{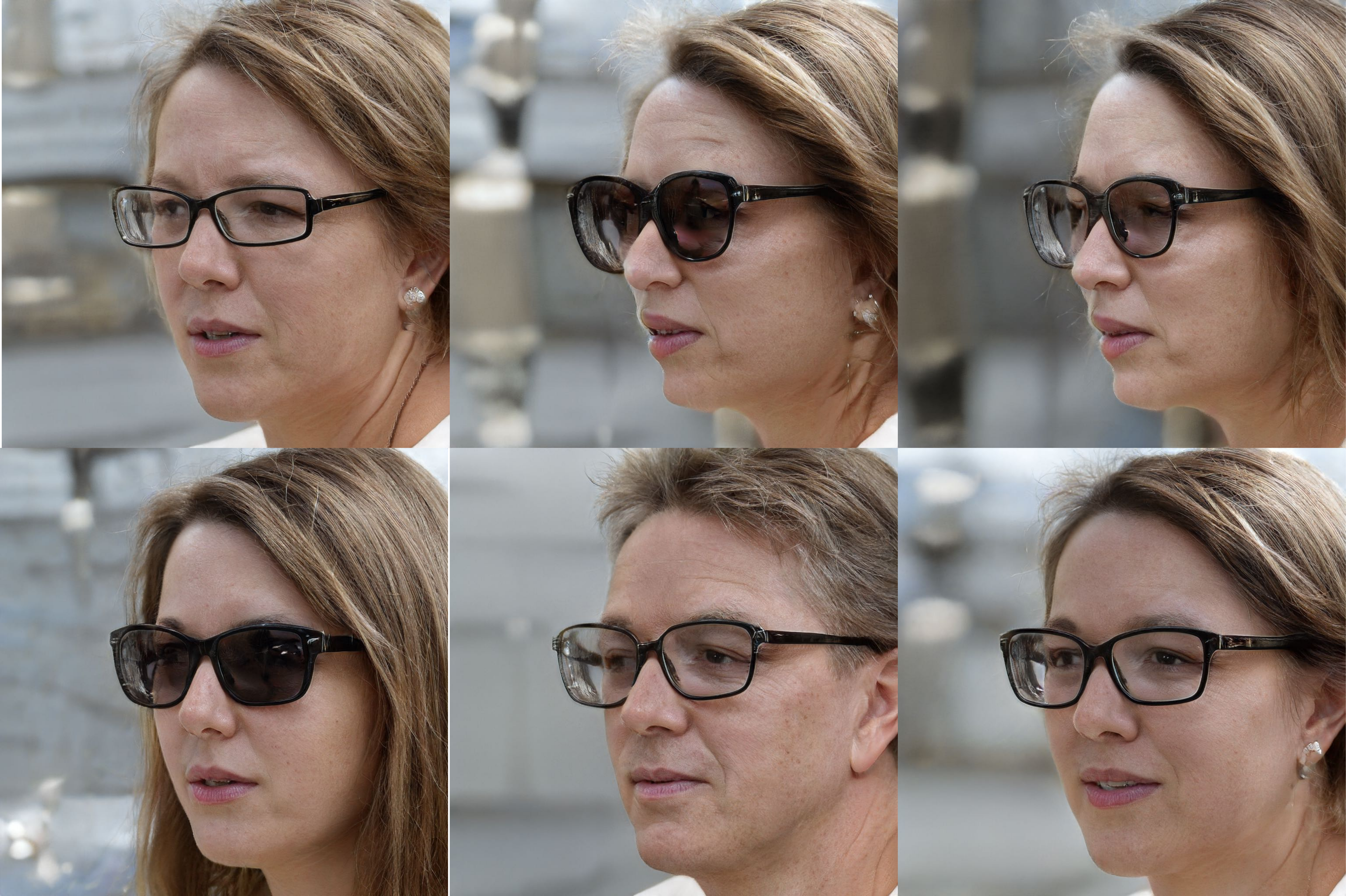}} 
    \hspace{.01\textwidth}
    \subfloat[Client 4]{\includegraphics[width=.14\textwidth]{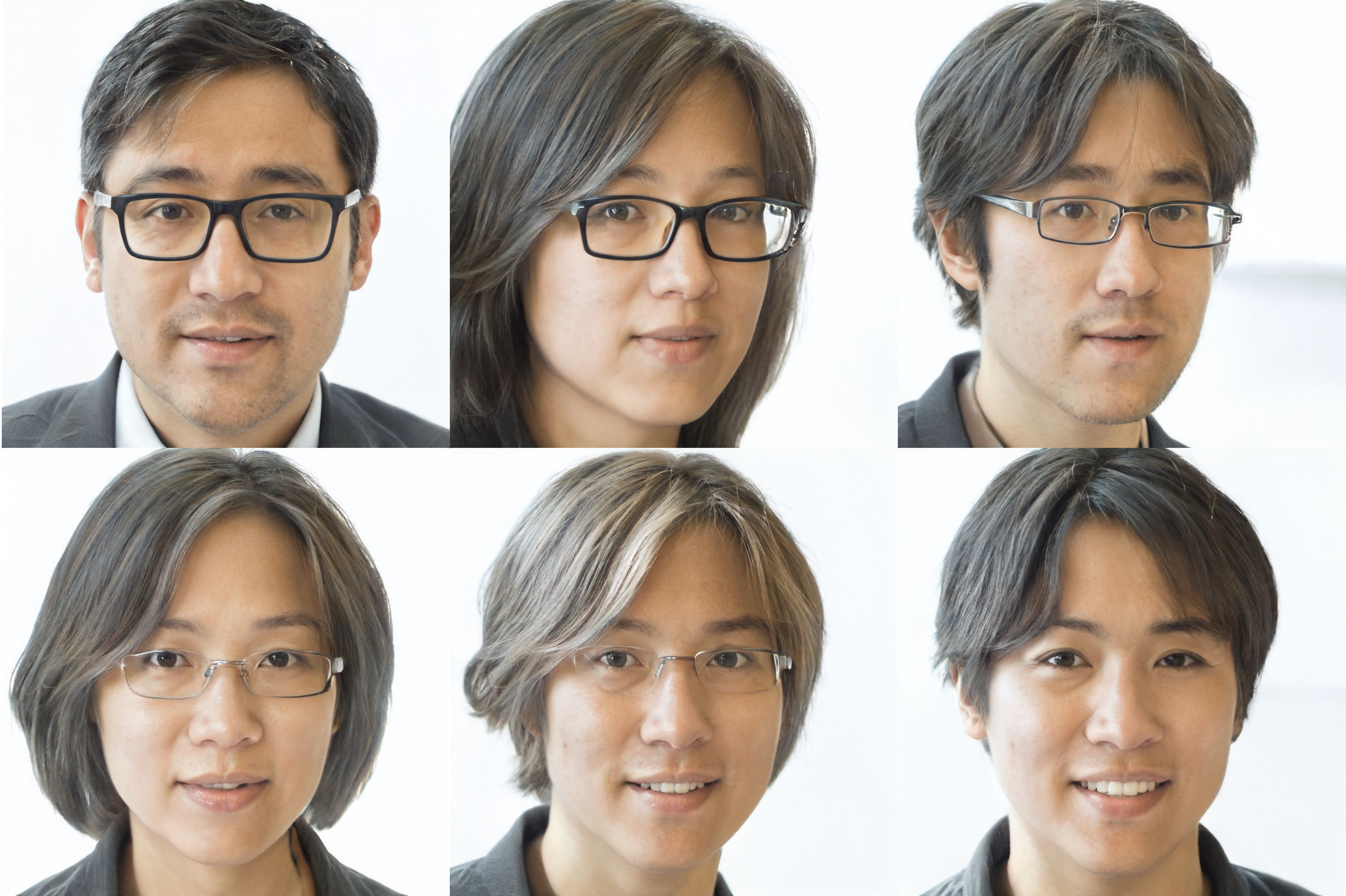}}
    \hspace{.01\textwidth}
    \subfloat[Client 5]{\includegraphics[width=.14\textwidth]{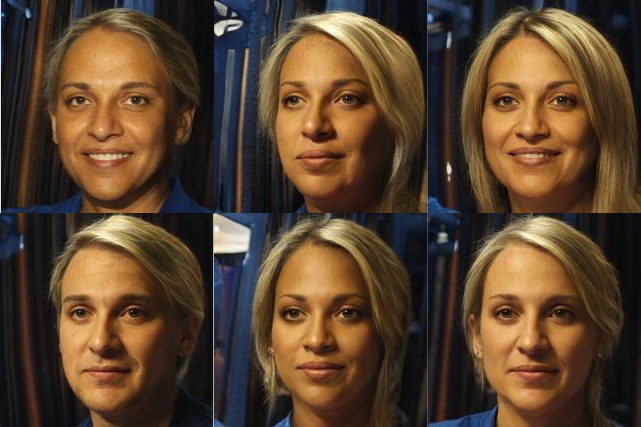}}
    \hspace{.01\textwidth}
    \subfloat[Client 6]{\includegraphics[width=.14\textwidth]{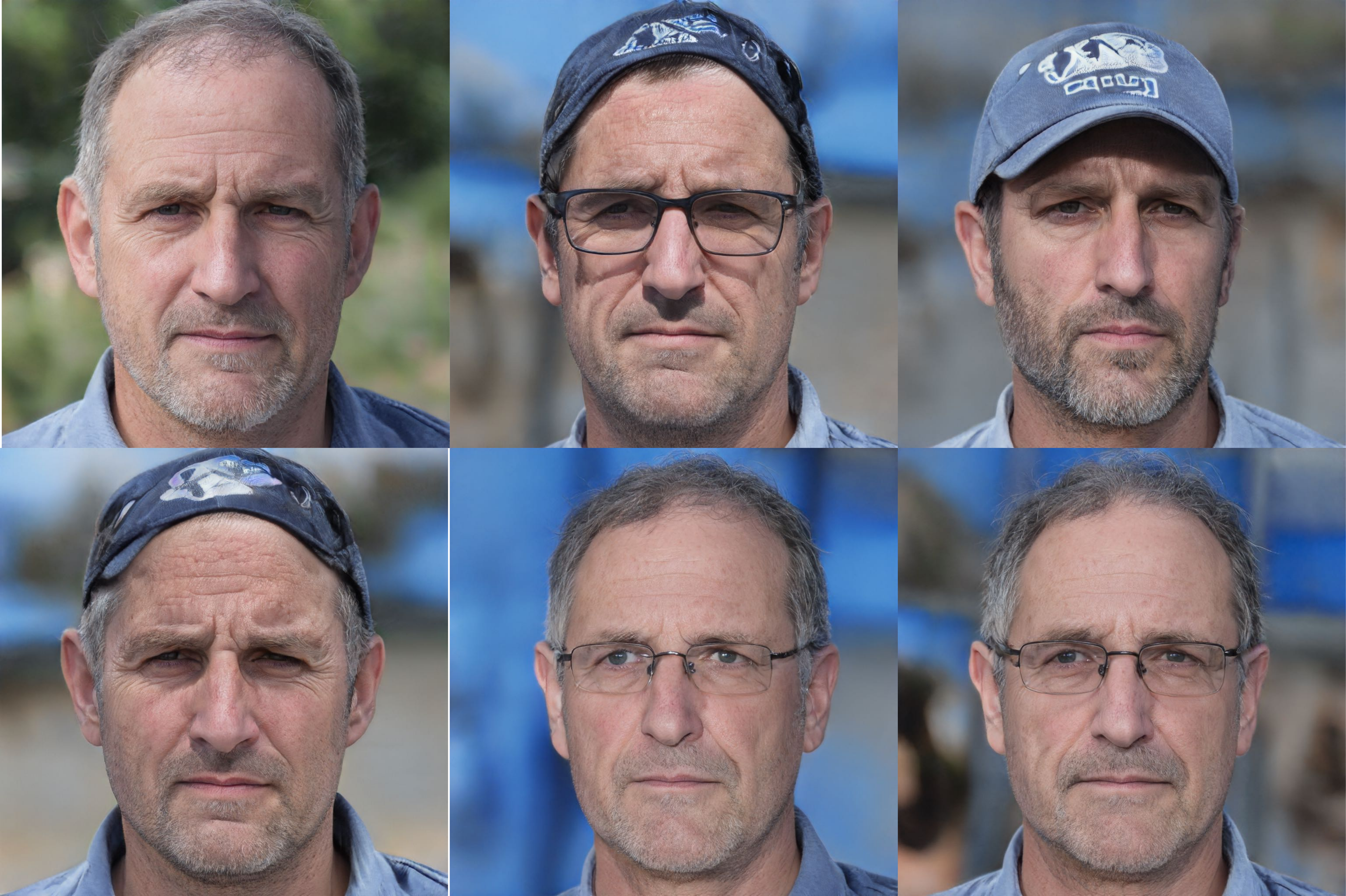}}\\
    \caption{Illustration of simulated clients with heterogeneous distributions via truncation technique.}
    \label{fig:styleC}
\end{figure}

\minisection{Experiment Setting} Following a similar methodology as described in the previous subsection, we synthesized a series of diversity-controlled generators by applying the standard truncation technique~\citep{kynkaanniemi2019improved} to the random noise vector $\mathbf{z}$. We varied the truncation factor $\tau$ over $[0.01,1.0]$. The effect of changing the truncation factor on the generated samples' diversity is illustrated in \Cref{fig:styleG}. For every attempted $\tau$, we generated 50K samples. Additionally, to simulate a distributed setting with heterogeneous data distributions, we simulated 1000 clients, each with images synthesized with truncation factor $\tau = 0.25$. The centers of image distributions for each client varied, resulting in intra-client similarity and inter-client diversity as depicted in \Cref{fig:styleC}. We evaluated the generators in the distributed setting of the simulated clients using both the discussed FD and KD-based evaluation scores.

\minisection{Numerical Results} As shown in \Cref{fig:stylefid} and \Cref{fig:stylekid}, FD-avg and KD-avg led to different rankings of the models. The plot of FD-avg versus FD-all led to a U-shaped curve, revealing inconsistent rankings of the models. On the other hand, the difference between KD-avg and KD-all remained constant for the generators, as shown in Theorem~\ref{Thm: KD}. These findings are also visible in the comparative rankings based on FD scores in \Cref{fig:stylerank-2}. 
The results suggest that although KD scores and FD-all prefer generators with higher truncation factors and diversity, FD-avg preferred generated data with limited diversity matching the bounded diversity level at each client.

\begin{figure}[ht!]
    \centering
    \includegraphics[width=\linewidth]{supp/style_Gafhq.pdf}
    \caption{\textcolor{black}{Illustration of randomly-generated samples from the variance-controlled generators on AFHQ dataset. Images in each row are synthesized with the same truncation parameter.}}
    \label{fig:afhqg}
\end{figure}

\subsection{Results on AFHQ dataset}

\begin{figure}[ht]
    \centering
    \subfloat[Client 1]{\includegraphics[width=.3\linewidth]{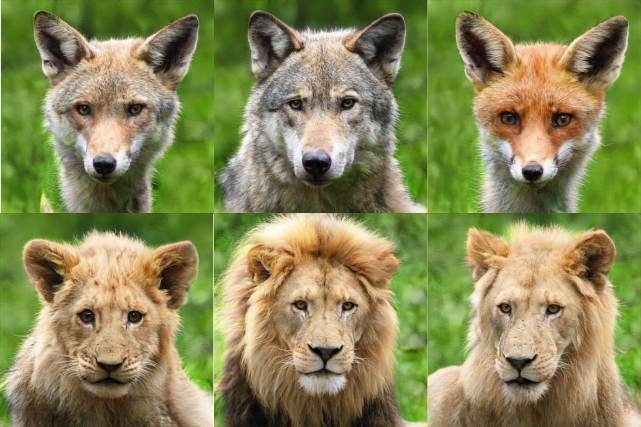}}
    \hfill
    \subfloat[Client 2]{\includegraphics[width=.3\linewidth]{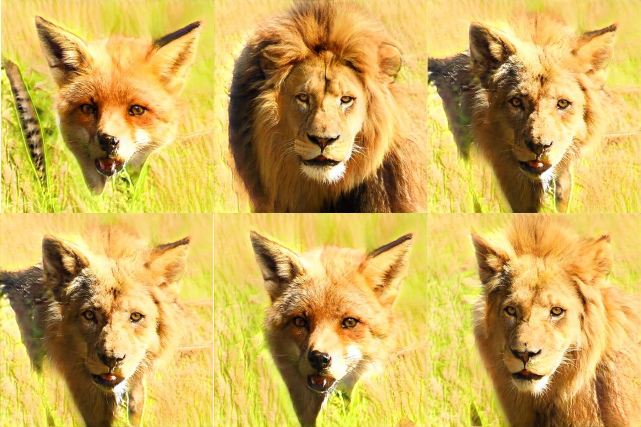}}
    \hfill
    \subfloat[Client 3]{\includegraphics[width=.3\linewidth]{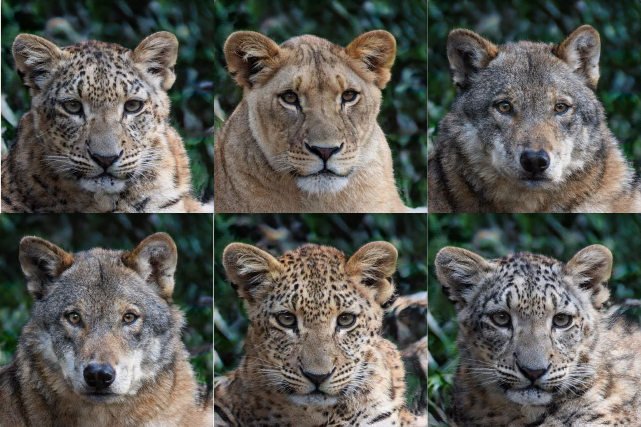}}\\
    
    \subfloat[Client 4]{\includegraphics[width=.3\linewidth]{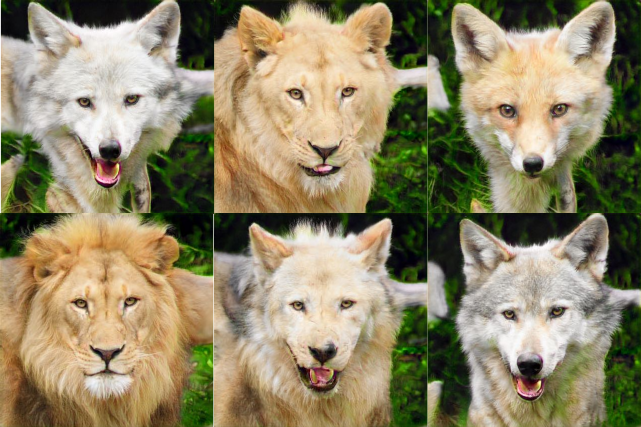}} 
    \hfill
    \subfloat[Client 5]{\includegraphics[width=.3\linewidth]{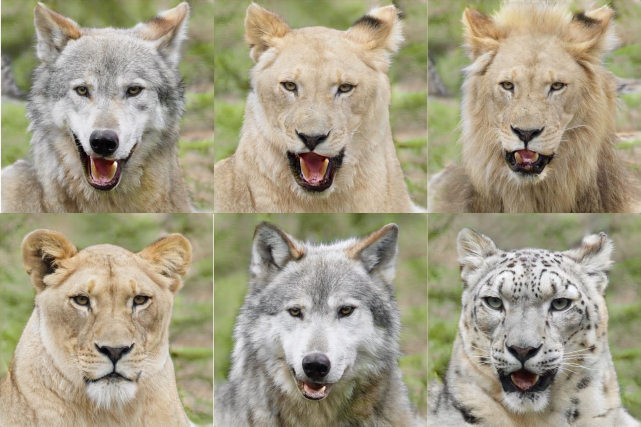}}
    \hfill
    \subfloat[Client 6]{\includegraphics[width=.3\linewidth]{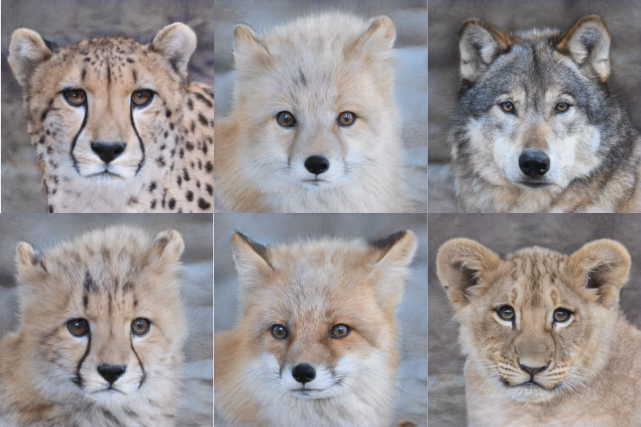}}

    \caption{Illustration of random samples from randomly selected variance-limited clients on AFHQ dataset.}
    \label{fig:afhqclients}
\end{figure}

\begin{figure}[ht!]
    \centering
    \includegraphics[width=\linewidth]{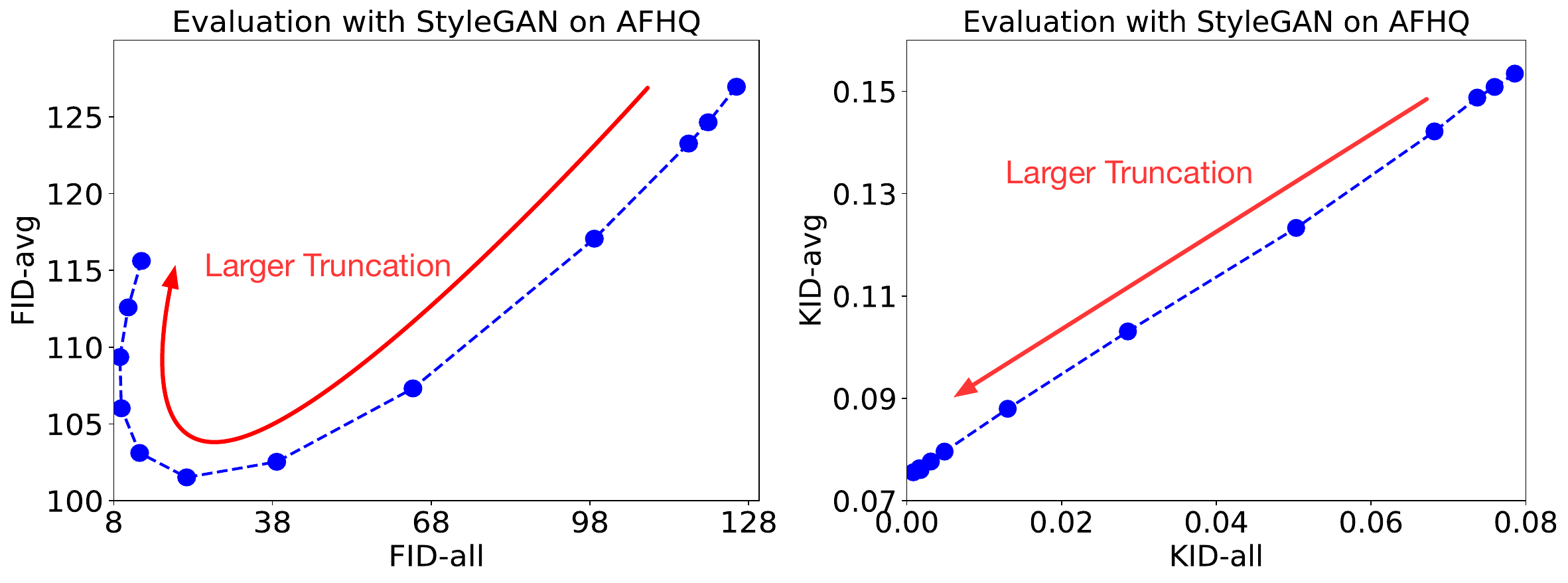}
    \caption{Evaluation on AFHQ dataset with StyleGAN2.}
    \label{fig:afhqfidkid}
\end{figure}

Using another model weight that is pre-trained on the AFHQ-wild dataset~\citep{choi2020starganv2}, we extend the above experiment. We gradually increase the truncation parameter of the generators. We show examples from generators and clients in \Cref{fig:afhqclients} and \Cref{fig:afhqg}, and plot the relationship between FD-scores and KD-scores in \Cref{fig:afhqfidkid}. The results illustrated in the figures still indicate that FD-avg can lead to different rankings from FD-all, while KD-avg can be treated as a more stable evaluation metric in the distributed learning setting.


\subsection{Results on CIFAR and ImageNet}

\minisection{Experiment Settings}
We evaluated our theoretical results on standard image datasets.
In our experiments, we simulated heterogeneous federated learning experiments consisting of non-i.i.d. data at different clients: for CIFAR-10 \citep{krizhevsky2009learning}, we considered 10 clients, each owning samples exclusively from a single class of the image dataset. Therefore, every client's dataset contains images having the same label. Similar to the federated CIFAR10 experiment, federated CIFAR100 and federated ImageNet-32 are constructed by grouping samples from each class.

\begin{figure*}[ht]
    \centering
    \includegraphics[width=0.85\textwidth]{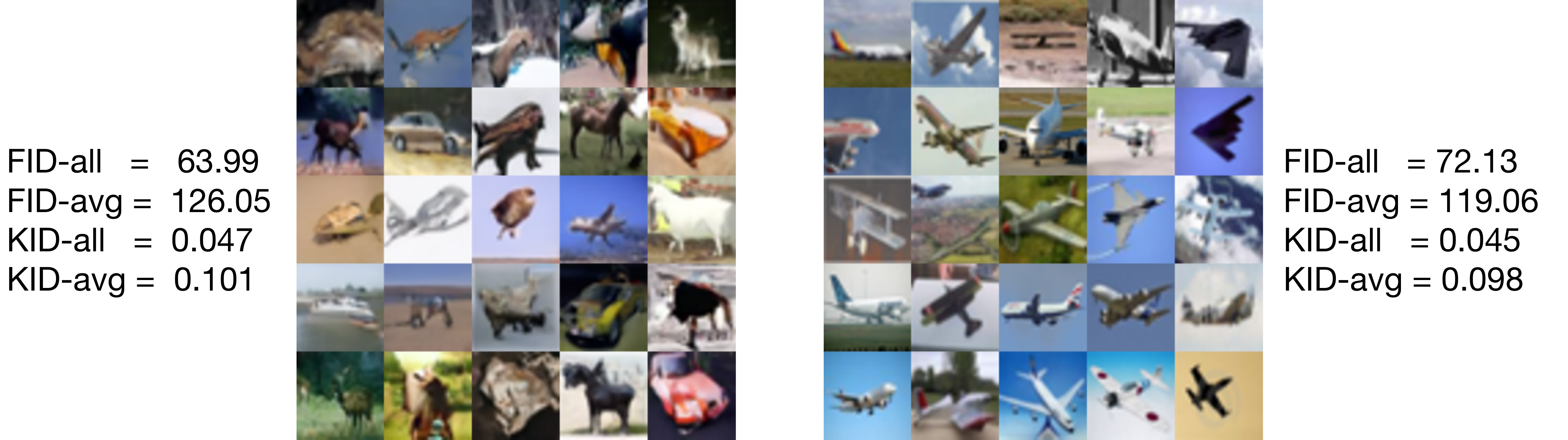}
    \caption{Left: Images generated by a generative model obtain a lower FD-all. Right: Images from real datasets with class 'airplane' obtain a lower FD-avg.  FD-avg and FD-all lead to inconsistent rankings, while KD-avg and KD-all result in the same ranking.}
    \label{fig:vis}
\end{figure*}

\textbf{Neural Net-based Generators.} We have trained WGAN-GP~\citep{salimans2016improved} and DDPM~\citep{ho2020denoising} in a federated learning setting by utilizing FedAvg approach~\citep{mcmahan2017communication}. The experiment protocols for WGAN-GP and DDPM are copied from original works. The communication interval of FedAvg is set as 160 iterations for both WGAN-GP and DDPM. We have tried different communication intervals for both models. The communication frequency will affect model performance but have no influence on the conclusions in the main part of our paper. 

\textbf{Perfect Data-simulating Generators.}
In our CIFAR-10 experiments, we also simulated and evaluated an "ideal generator" capable of perfectly replicating all samples belonging to the 'airplane' class in CIFAR10. In this scenario, the samples "generated" by the ideal generator exhibit impeccable fidelity but lack diversity since no samples from other categories can be produced.

\minisection{FD-based and KD-based Evaluation of Generative Models.}
We evaluated the generative models according to FD-all, FD-avg, KD-all, and KD-avg as defined in \Cref{sec:4}. In several cases, we observed that FD-all~/~FD-avg could assign inconsistent rankings to the generators.
Specifically, we computed FD-all and FD-avg for the ideal 'airplane'-class-based generator and neural net-based DDPM generators under the distributed CIFAR10 setting.
We present some examples generated from the two generators in \Cref{fig:vis} and report their scores according to the four metrics. The results suggest that FD-avg assigns a considerably lower score to the ideal 'airplane'-based generator, whose images preserve perfect details but lack diversity in image categories. Conversely, FD-all assigns a relatively lower value to the DDPM model because its images possess greater diversity. 
On the other hand, we also observed that KD-avg and KD-all give consistent rankings. Both of them led to the evaluation that the ideal airplane generator is slightly better than the DDPM generator. In our implementation of KD-scores, we utilized the standard implementation of KD measurement from data with a polynomial kernel, $k(\bf{x},\bf{y}) = \left(\frac{1}{d}\bf{x}^T\bf{y}+1\right)^3$, where $d$ is the dimension of feature vector. We note that our theoretical finding on the evaluation consistency under KD-all and KD-avg applies to every kernel similarity function. We also test KD-scores with a Gaussian RBF kernel $k_{\sigma}^{\text{rbf}}(\bf{x},\bf{y}) = \exp\left(-\frac{1}{2\sigma^2}\|\bf{x}-\bf{y}\|^2\right)$ as formulated in \citet{binkowski2018demystifying}, where we chose $\sigma = \sqrt{d}$ in the experiments. For images generated by the diffusion model, $\text{KD}^{\text{rbf}}$-all gives $4.277e^{-3}$ while $\text{KD}^{\text{rbf}}$-avg gives $4.295e^{-3}$. For the airplane images in CIFAR10, $\text{KD}^{\text{rbf}}$-all gives $4.283e^{-3}$ while $\text{KD}^{\text{rbf}}$-avg gives $4.301e^{-3}$. The results indicate that for the Gaussian RBF kernel $k_{\sigma}^{\text{rbf}}$, $\text{KD}^{\text{rbf}}$-all and $\text{KD}^{\text{rbf}}$-avg still give consistent results. In this case, the $\text{KD}^{\text{rbf}}$-based evaluation suggests the images sampled from the diffusion model have higher quality than the set of airplane images in the CIFAR10 dataset.

\minisection{Evaluate Sequence of Net-based Generator on Federated CIFAR10}
We trained the WGAN-GP~\citep{gulrajani2017improved} generative models multiple times using different random states, and we set different training lengths for every training procedure. We saved the models at different checkpoints every $10$ epochs, which is common in training generative models to select the best-performing saved model according to an evaluation metric.

\begin{figure}[tb!]
    \centering
    \includegraphics[width=\linewidth]{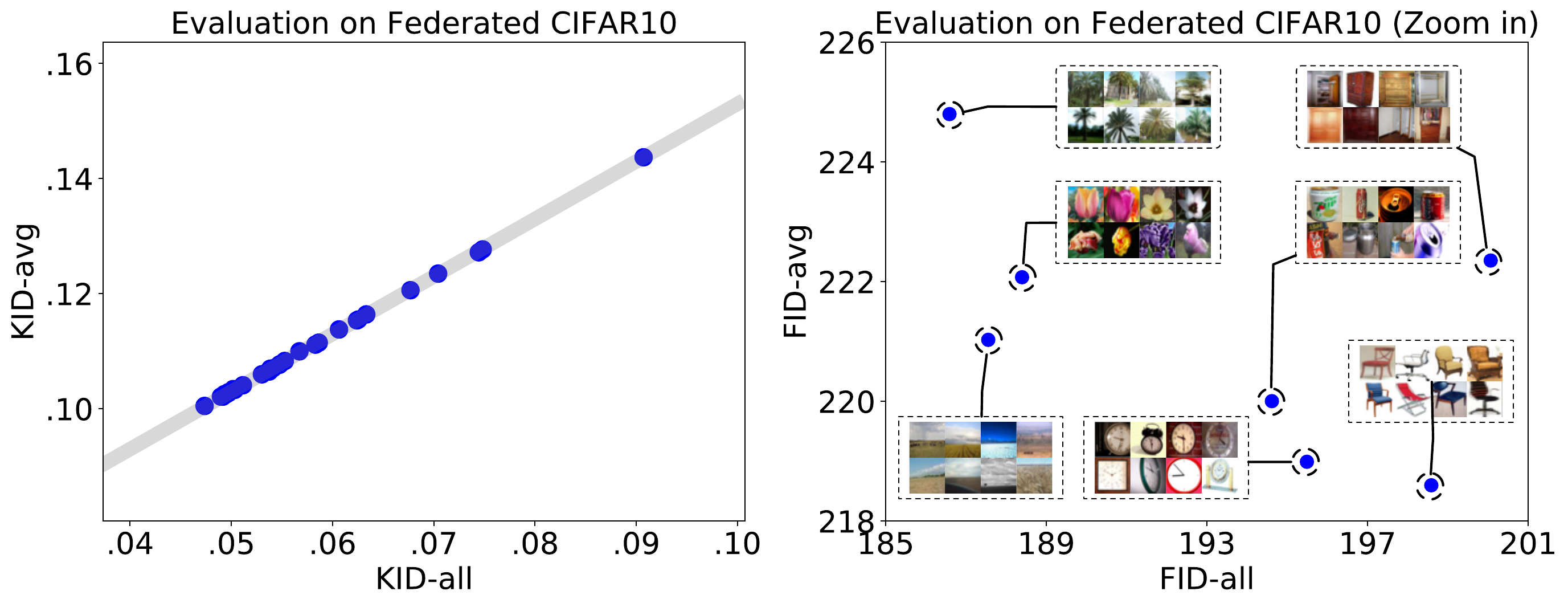}
    \caption{
        Left: KD evaluations of WGAN-GP checkpoints on federated CIFAR10. Right: FD-based evaluations on federated CIFAR-10.
    }
    \label{fig:fedc10}
\end{figure}

Our numerical results suggest that the gap between KD-all and KD-avg remains constant and hence they lead to the same rankings of the generative models. Here, we conducted our evaluations on all the generative models instances as previously described, and the results are visualized in the left sub-figure of \Cref{fig:fedc10}. These findings reveal that all distinct generators consistently exhibit a uniform gap between KD-avg and KD-all. Consequently, our results indicate that the rankings established by KD-avg consistently align with those of KD-all in distributed learning settings.

\minisection{Evaluate CIFAR100 Generator on Federated CIFAR10} To further experiment the ranking of generative models according to the discussed aggregate scores, we extracted samples from each class of CIFAR-100 and treated them as the output of one hundred distinct generators, each corresponding to a single class. By assessing these generators on the federated CIFAR-10 dataset, we obtained one hundred pairs of FD-avg~/~FD-all values, and a subset of these pairs with inconsistent rankings according to FD-all/FD-avg is visualized in the right of \Cref{fig:fedc10}. 
The complete set of evaluation results is available on \Cref{tab:fullc100}.
These results further highlight that the rankings provided by FD-all and FD-avg can exhibit inconsistencies in the context of distributed learning. Such inconsistencies could pose a challenge when selecting from a series of checkpoints or model architectures during the training of generative models in distributed learning scenarios, where a distributed computation of FD-all is more challenging than obtaining FD-avg due to privacy considerations.

\begin{table*}[ht]
    \centering
    \resizebox{.85\linewidth}{!}
    {
        \begin{tabular}{c|cc|cc||c|cc|cc}
            \toprule
            Class &FD-all & FD-avg & KD-all & KD-avg&Class &FD-all & FD-avg & KD-all & KD-avg\\ \midrule
            0&267.4&285.9&0.201&0.253&25&142.4&173.2&0.067&0.116\\
1&173.2&205.5&0.109&0.165&26&139.4&175.5&0.063&0.114\\
2&151.8&185.4&0.072&0.123&27&114.5&153.7&0.054&0.102\\
3&124.3&162.2&0.047&0.100&28&182.0&205.0&0.112&0.160\\
4&117.0&156.1&0.049&0.100&29&143.3&185.2&0.083&0.133\\
5&157.6&185.2&0.088&0.138&30&147.4&179.8&0.084&0.134\\
6&142.1&179.7&0.062&0.115&31&160.4&193.6&0.108&0.158\\
7&144.9&179.2&0.066&0.113&32&115.4&156.4&0.032&0.085\\
8&155.0&187.7&0.085&0.135&33&146.9&181.2&0.077&0.126\\
9&180.7&203.6&0.106&0.156&34&127.5&163.4&0.068&0.117\\
10&183.2&208.8&0.099&0.151&35&150.4&184.7&0.080&0.131\\
11&151.9&184.9&0.073&0.127&36&142.5&176.5&0.070&0.126\\
12&126.1&163.1&0.056&0.108&37&126.7&162.9&0.066&0.118\\
13&127.0&159.4&0.058&0.112&38&110.5&151.3&0.050&0.100\\
14&145.8&182.9&0.071&0.126&39&233.5&257.2&0.145&0.196\\
15&124.2&165.2&0.054&0.107&40&158.9&187.6&0.078&0.128\\
16&190.3&213.6&0.112&0.161&41&151.1&181.0&0.072&0.126\\
17&161.5&192.8&0.117&0.165&42&137.2&174.7&0.071&0.122\\
18&142.7&179.8&0.069&0.118&43&152.0&186.0&0.090&0.139\\
19&112.4&154.1&0.047&0.101&44&126.7&165.8&0.050&0.103\\
20&194.2&215.9&0.123&0.175&45&135.4&173.0&0.059&0.111\\
21&164.8&194.9&0.099&0.148&46&155.1&187.1&0.080&0.133\\
22&205.1&227.0&0.126&0.180&47&174.2&203.4&0.118&0.169\\
23&191.1&219.7&0.130&0.186&48&145.0&175.3&0.077&0.129\\
24&174.9&202.8&0.105&0.155&49&164.8&194.6&0.110&0.164\\
50&113.8&154.0&0.045&0.095&75&151.7&184.3&0.096&0.145\\
51&150.4&184.6&0.073&0.125&76&151.5&183.5&0.068&0.118\\
52&195.3&222.1&0.167&0.218&77&139.0&174.9&0.066&0.117\\
53&279.7&299.1&0.217&0.270&78&197.3&228.3&0.130&0.182\\
54&170.2&201.3&0.098&0.149&79&138.9&174.8&0.062&0.114\\
55&104.7&146.3&0.034&0.086&80&111.1&149.9&0.043&0.095\\
56&144.2&178.5&0.075&0.126&81&131.7&165.3&0.068&0.119\\
57&192.5&219.9&0.115&0.163&82&197.6&227.2&0.123&0.178\\
58&131.5&161.0&0.067&0.121&83&202.6&230.6&0.122&0.173\\
59&149.7&183.7&0.093&0.144&84&144.5&177.3&0.065&0.111\\
60&188.0&216.0&0.144&0.197&85&123.0&160.7&0.079&0.125\\
61&249.2&270.6&0.179&0.229&86&168.0&193.0&0.083&0.135\\
62&202.1&230.6&0.133&0.184&87&170.5&196.0&0.098&0.152\\
63&140.6&175.3&0.071&0.120&88&133.4&170.6&0.068&0.120\\
64&118.7&156.1&0.049&0.100&89&122.3&158.1&0.059&0.112\\
65&102.2&142.2&0.024&0.077&90&110.7&148.3&0.041&0.093\\
66&121.7&159.1&0.054&0.105&91&124.0&160.7&0.048&0.098\\
67&132.5&167.8&0.063&0.115&92&175.4&206.2&0.096&0.149\\
68&139.7&173.1&0.073&0.123&93&129.7&166.3&0.054&0.109\\
69&143.2&176.0&0.068&0.121&94&213.4&235.2&0.162&0.212\\
70&178.4&209.5&0.095&0.148&95&154.7&185.2&0.084&0.133\\
71&169.4&199.0&0.120&0.167&96&147.8&181.7&0.090&0.138\\
72&114.1&155.6&0.043&0.094&97&137.1&171.4&0.068&0.119\\
73&137.9&170.8&0.071&0.124&98&157.1&188.6&0.082&0.134\\
74&124.7&162.1&0.061&0.108&99&204.9&233.9&0.143&0.192\\
            \bottomrule 
        \end{tabular}
    }
    \caption{Full evaluation of CIFAR100 on Federated CIFAR10.}

    \label{tab:fullc100}
\end{table*}
            


\begin{figure}[tb!]
    \centering
    \subfloat[\label{fig:C100XIN}]{
        \includegraphics[width=0.47\linewidth]{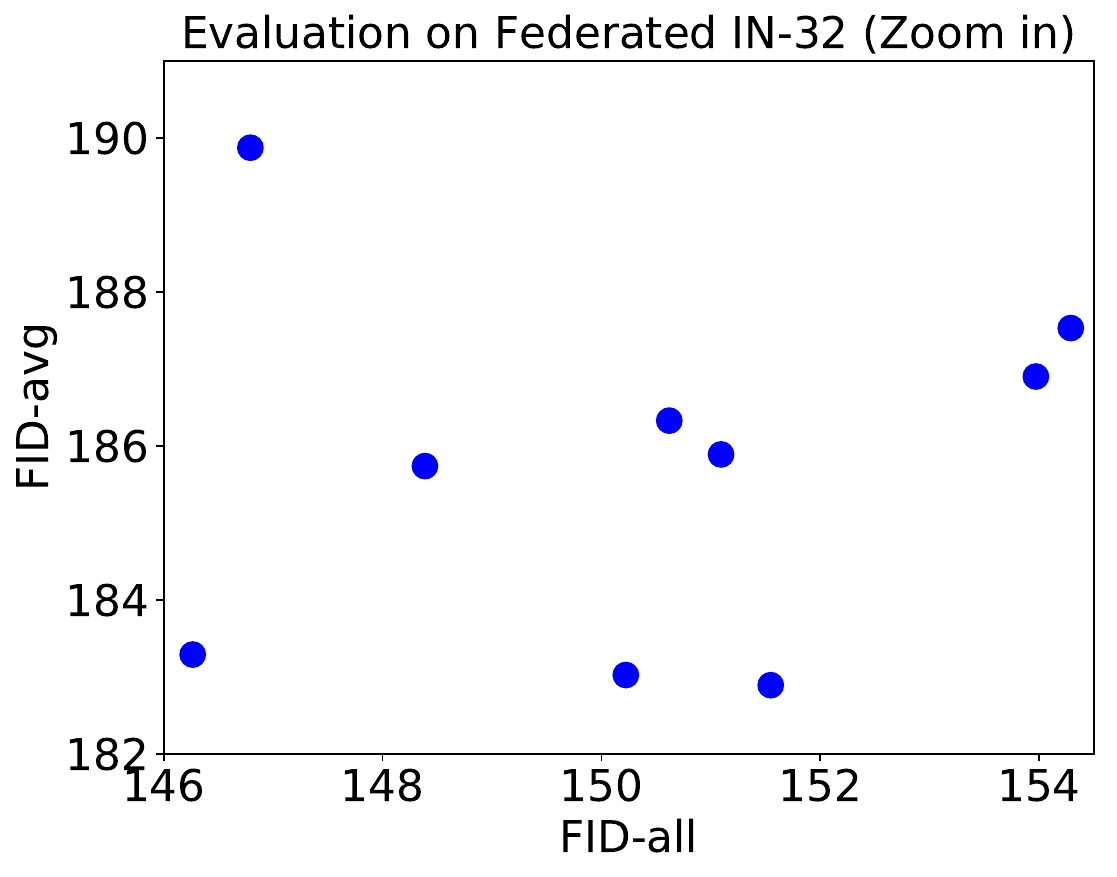}
    }%
    \hfill
    \subfloat[\label{fig:inrank}]{
        \includegraphics[width=0.47\linewidth]{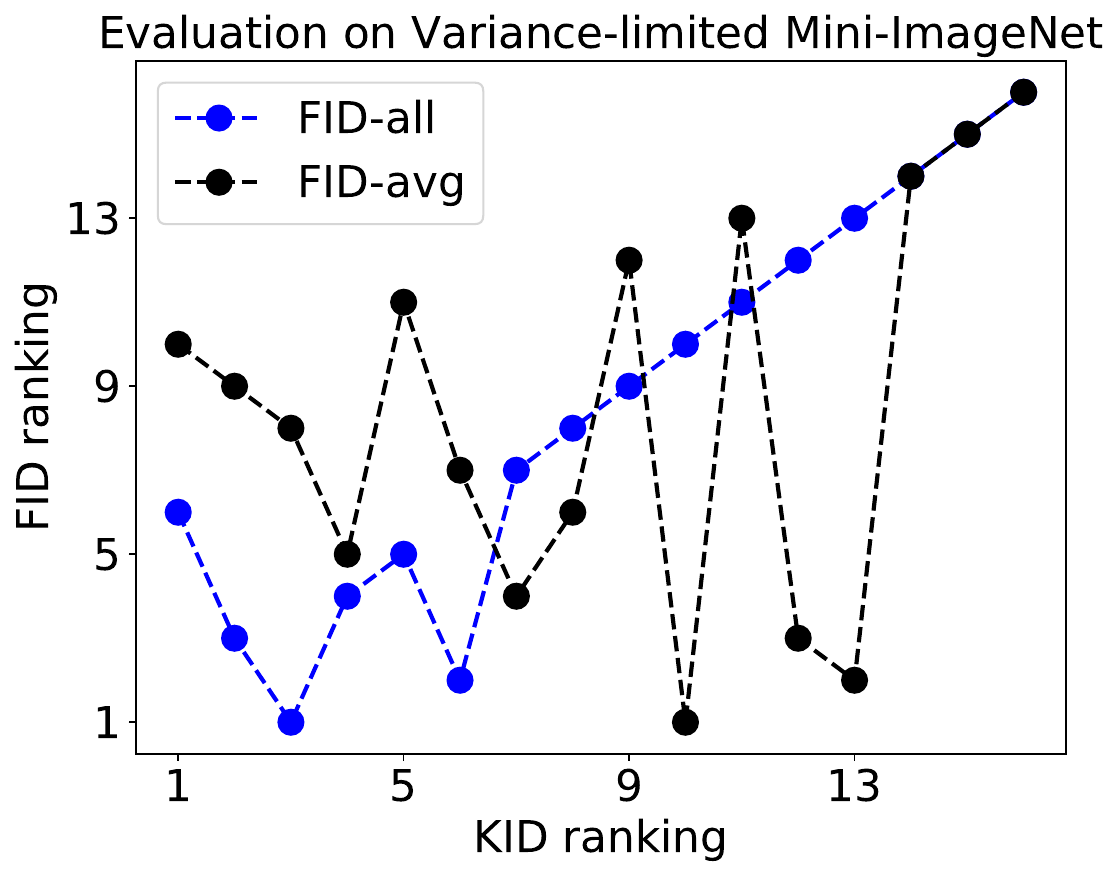}
    }
    \caption{
        \textbf{Left (a):} Evaluation of CIFAR100 generators on Federated ImageNet-32. 
        \textbf{Right (b):} Comparison of FD-based and KD-based rankings for variance-limited federated Mini-ImageNet models (lower rank is better).
    }
    \label{fig:combined}
\end{figure}

\minisection{Evaluate CIFAR100 on Federated ImageNet-32} We expand the evaluation of CIFAR100 to the Federated ImageNet-32 dataset. Similarly, we extracted samples from each class of CIFAR-100 and treated them as the output of one hundred
distinct generators, each corresponding to a single class. We also keep the first one hundred classes of ImageNet-32 and simulate one hundred clients. Each client holds all images ($\sim$1300) from a single class. We evaluate all the generators on Federated ImageNet-32 and the result is shown in \Cref{fig:C100XIN}. The ranks provided by FD-avg and FD-all are inconsistent in a much more complex distributed learning setting.

\begin{figure}[tb!]
    \centering
    \includegraphics[width=\linewidth]{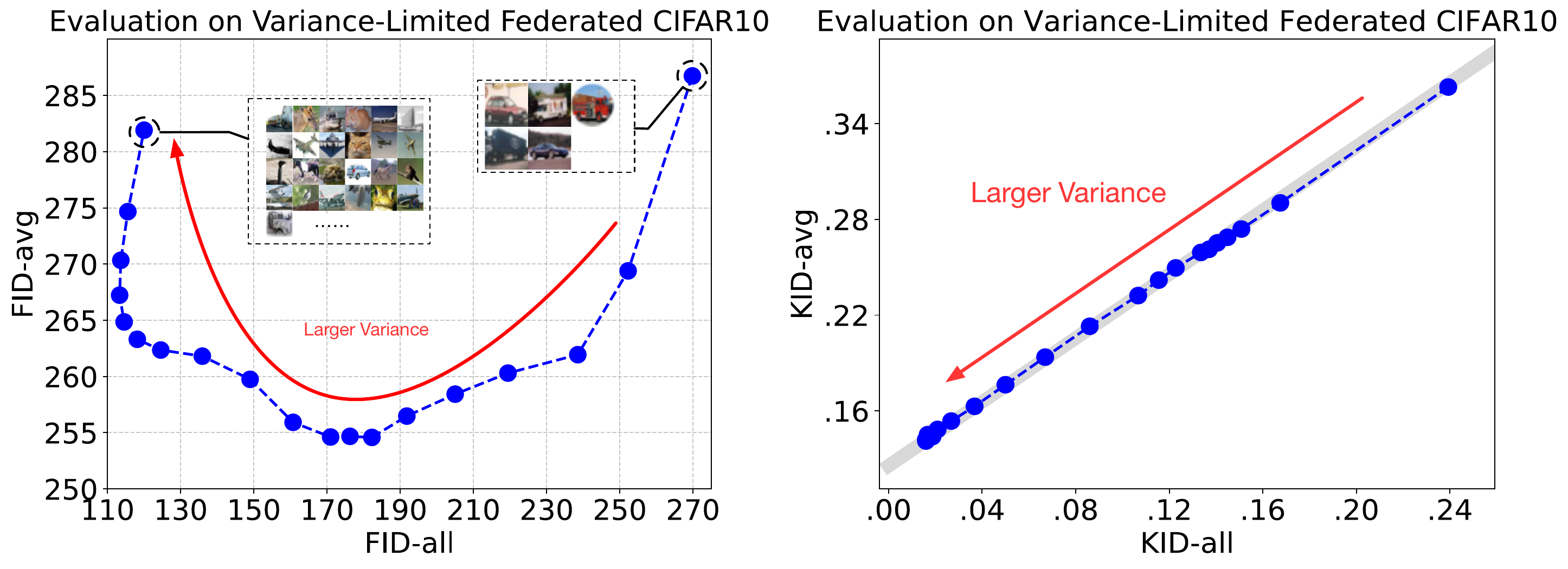}
    \caption{
        Evaluation on Variance-Limited Federated CIFAR10.
    }
    \label{fig:vlfedc10}
\end{figure}

\begin{figure}[ht!]
    \centering
    \includegraphics[width=\linewidth]{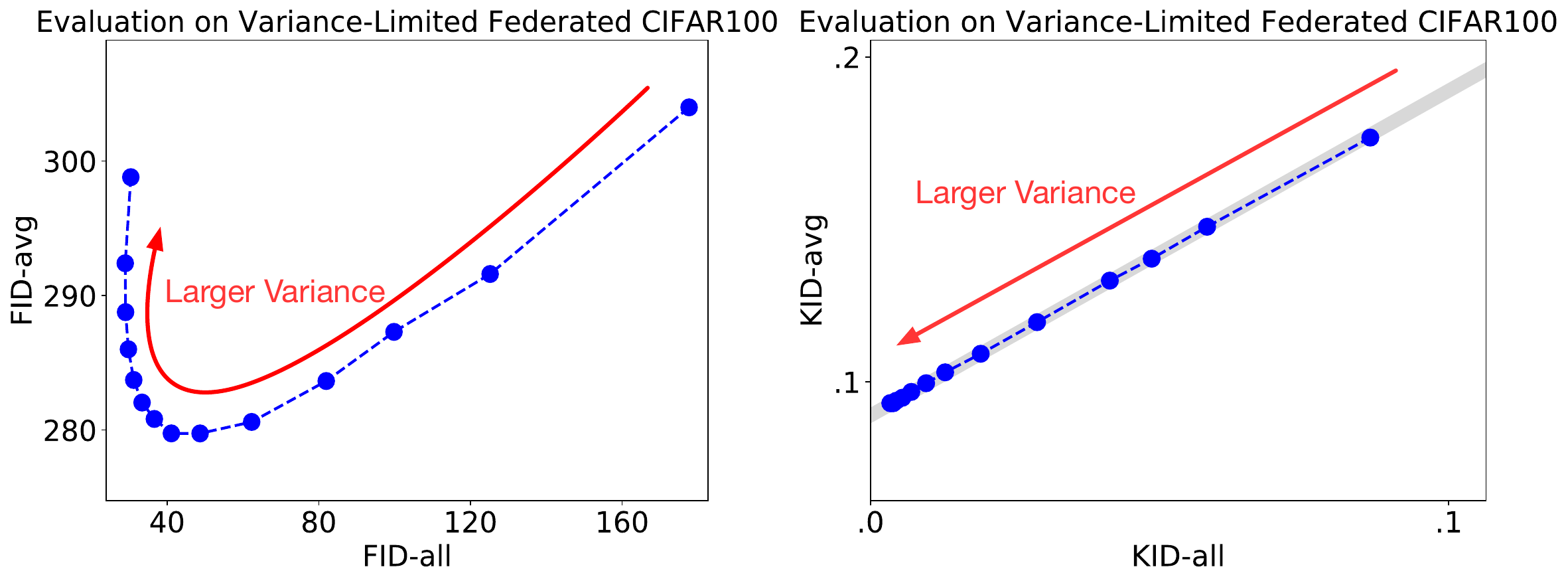}
    \caption{Evaluation on Variance-Limited Federated CIFAR100.}
    \label{fig:cifar100}
\end{figure}

\begin{figure}[tb!]
    \centering
    \includegraphics[width=\linewidth]{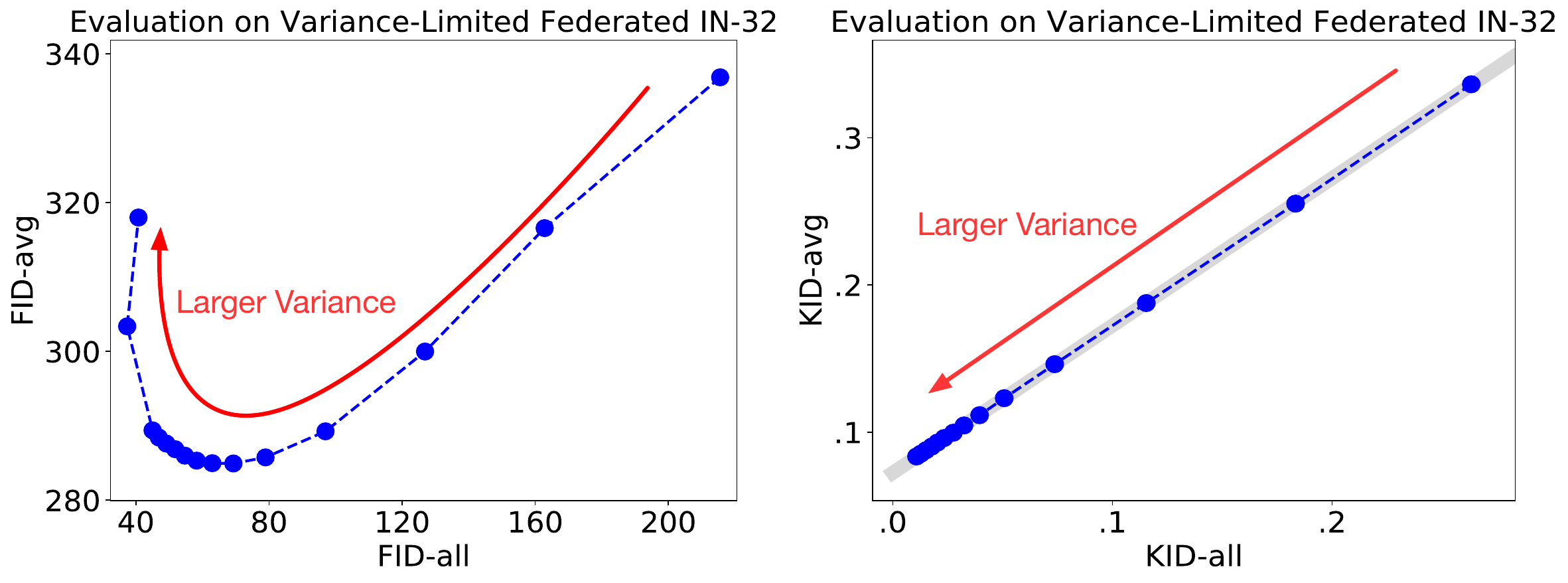}
    \caption{
        FD and KD-based Evaluations of variance-controlled generators on variance-limited federated ImageNet-32.
    }
    \label{fig:in}
\end{figure}

\subsection{Evaluation on Variance-Limited Federated Datasets}

\minisection{Experiment Setting} In the federated learning literature, it is relatively common that each client possesses only a small portion of the collective dataset, and the data diversity within each client's holdings is significantly constrained. To illustrate, consider the case of smartphone users who exclusively own pictures of themselves, all of which share remarkable similarity. Nevertheless, in a network comprising millions of users, the overall dataset's distribution still exhibits significant variance. In such scenarios, our theoretical framework suggests that the disparity between FD-all and FD-avg can become more pronounced.
To experiment the effect of such distribution heterogeneity, we simulated and evaluated generative models under \emph{variance-limited federated datasets}. To obtain a variance-limited federated dataset, for each class in the image dataset, we kept only a single image and its K-nearest neighbors. To find the $K$ nearest neighbors, we used the $L_2$-distance in the Inception-V3 2048-dimensional semantic space.
It is worth noting that our experimentation has shown that varying K within the range of 5 to 100 does not alter the core conclusions. 
This approach effectively mimics scenarios where each client's data has limited variance. We simulated the variance-limited federated learning setting for CIFAR-10, CIFAR-100 and a 32$\times$32 version of ImageNet (IN-32). For CIFAR-10 and CIFAR-100, we utilized all the classes in the dataset and for IN-32 we utilized the first 100 classes. We chose $K=20$ in the experiments. Intuitively, a larger $K$ leads to a more significant intra-client variance.

\textbf{Variance-controlled Generators.} To simulate a generator, we initiate the process by randomly selecting a sample from the dataset. We then gather its M-nearest neighbors from the original dataset without the federated learning partition. We consider this subset of samples as generated samples from a generator denoted by $G_M$. By increasing the value of $M$, we generated a sequence of generators with progressively higher variance values. We tried the $M$ range from 100 to 50000. We evaluated all the generative models, denoted as $G_M$ with the chosen $M$ values, using the Variance-Limited Federated datasets. 

\begin{figure}[tb!]
    \centering
    \subfloat[]{ \includegraphics[width=0.49\linewidth]{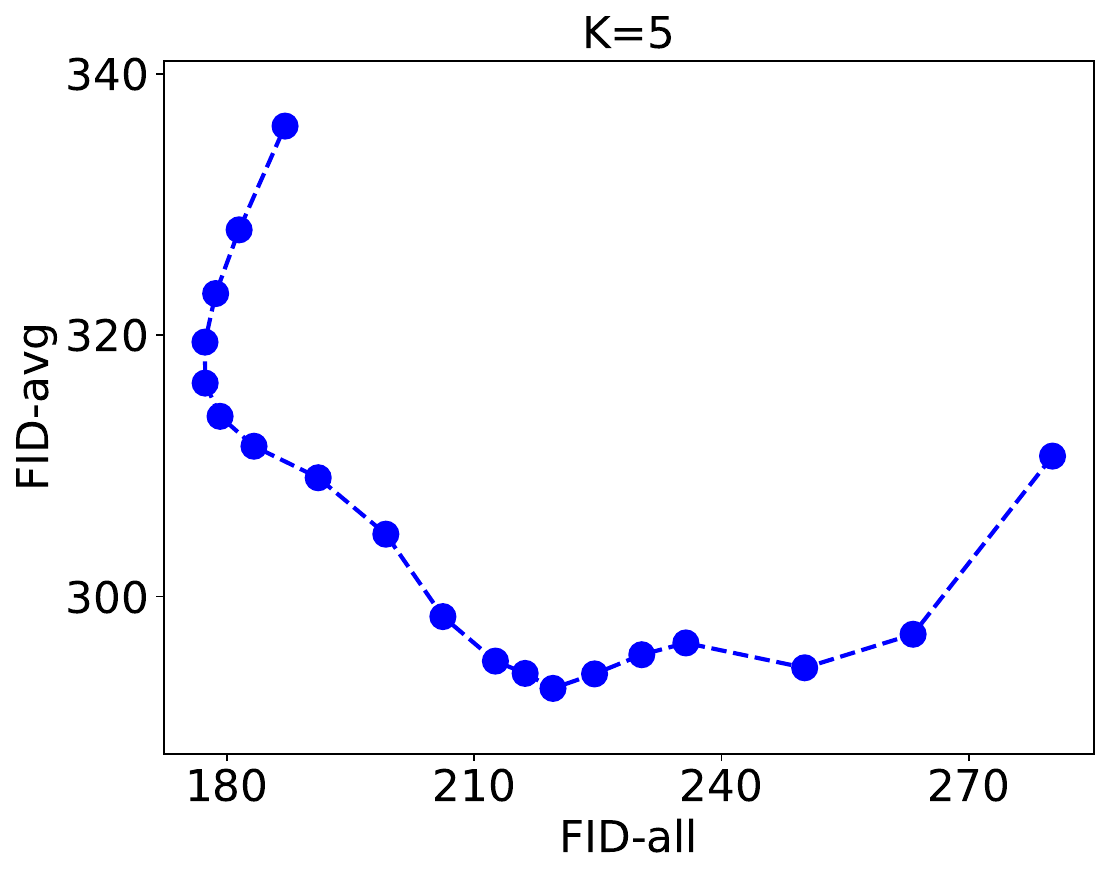} }
    \subfloat[]{ \includegraphics[width=0.49\linewidth]{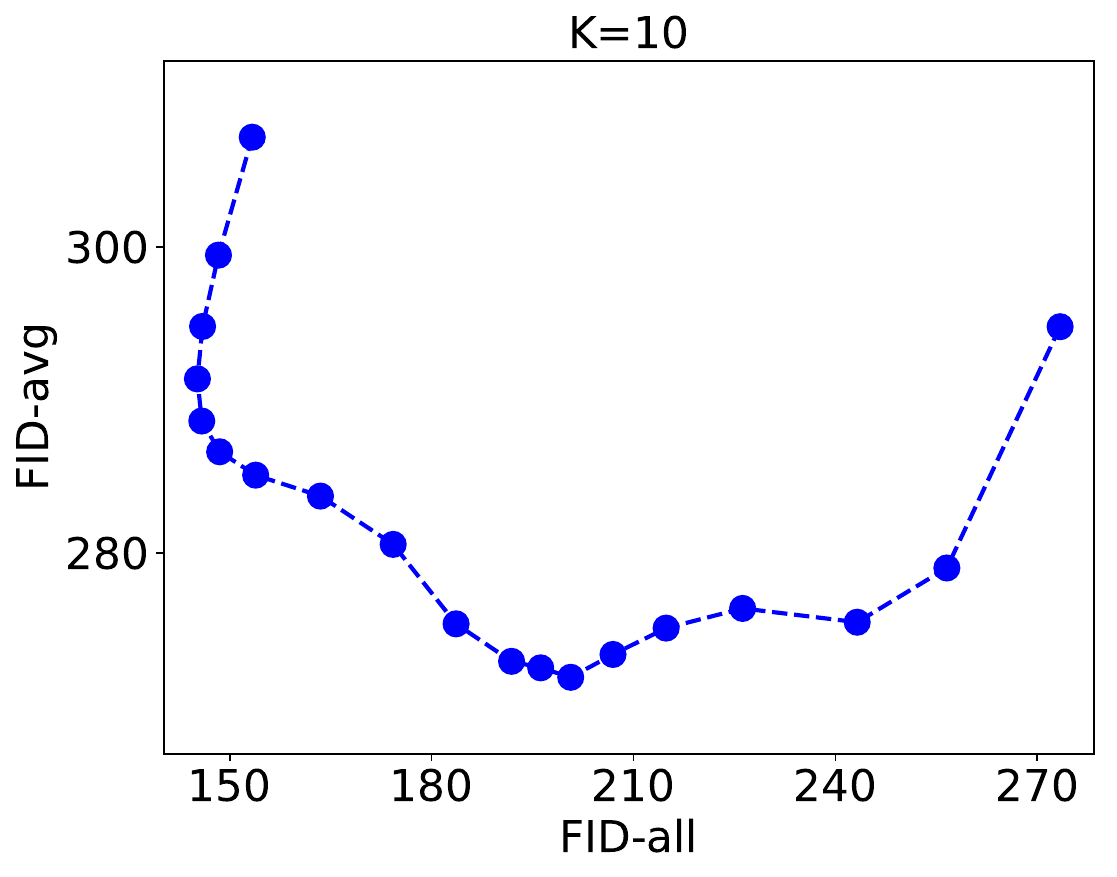} }\\
    \subfloat[]{ \includegraphics[width=0.49\linewidth]{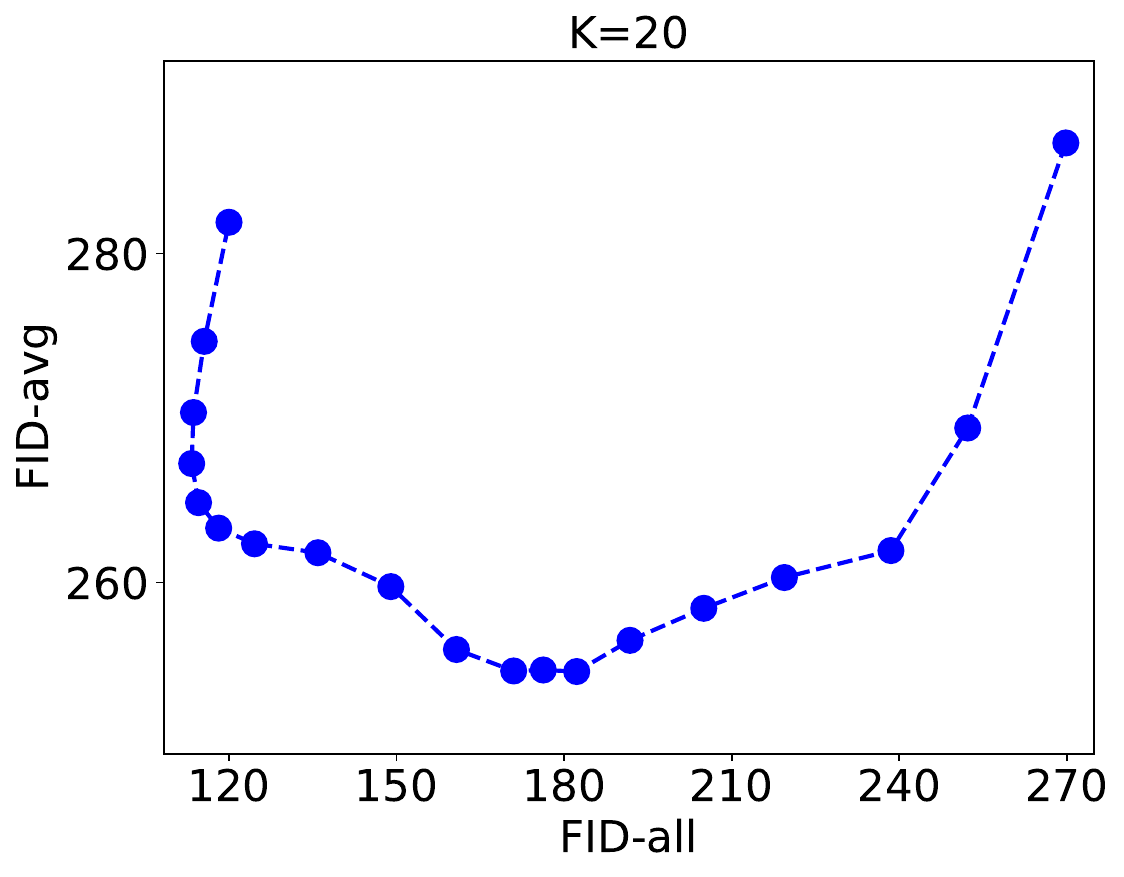} }
    \subfloat[]{ \includegraphics[width=0.49\linewidth]{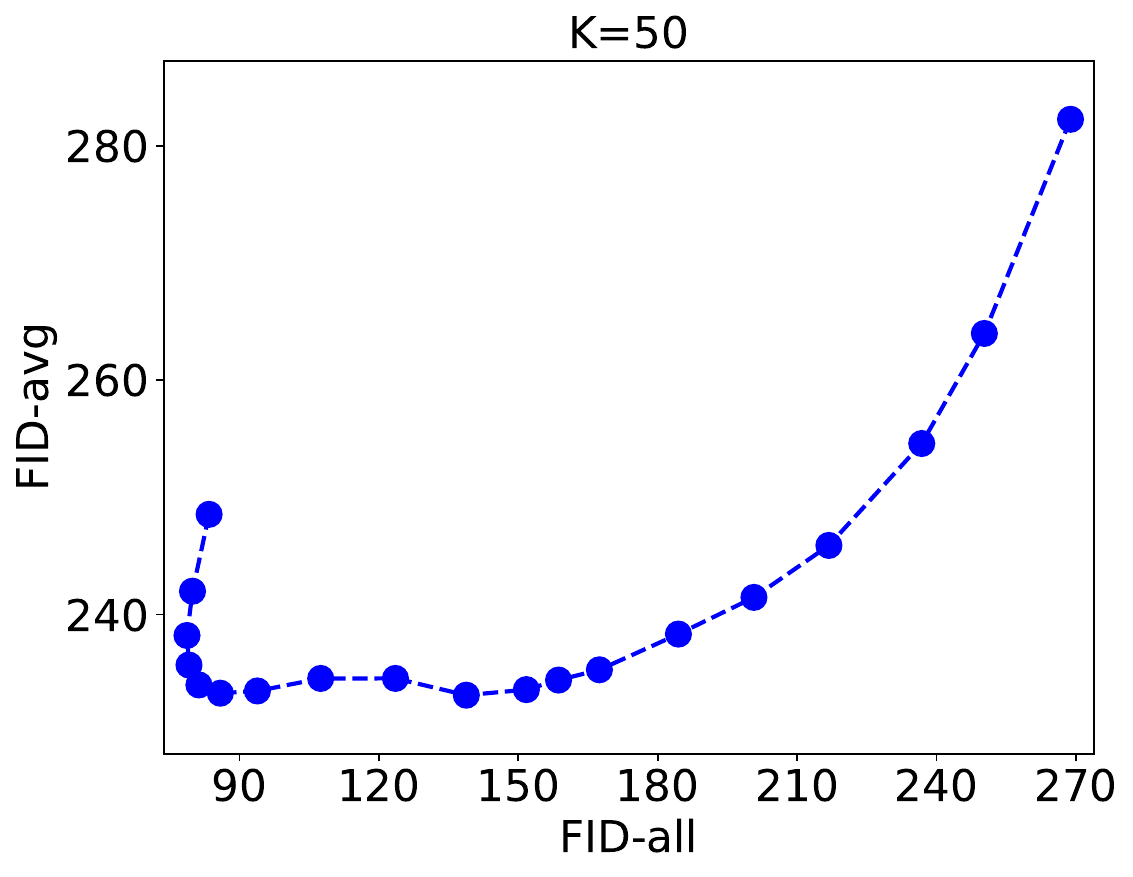} }
    \caption{
        Ablation study on hyper-parameter K.
    }
    \label{fig:k}
\end{figure}

\minisection{Results on Variance-Limited Federated CIFAR10} The evaluation results on CIFAR10 are shown in \Cref{fig:vlfedc10}. Our findings reveal a distinct pattern in the behavior of FD-avg and FD-all as generator variance varies while the distance between KD-avg and KD-all remains the same. Our numerical results highlight the impact of the choice of FD-all and FD-avg on model rankings in federated learning settings with limited intra-client variance, which can be broadly categorized into three phases.

\minisection{Results on Variance-Limited Federated CIFAR100} Similar to the experiments on CIFAR10, we have also applied the variance-limited federated dataset setting to CIFAR100. We keep $K=20$ images in each class. For variance-controlled generators, we select a sample from the original CIFAR100 dataset and gather the M-nearest neighbors. The range of $M$ remains the same as that in the previous subsection. We show the results in \Cref{fig:cifar100}. The results still support our main claims: FD-avg and FD-all give inconsistent results while KD-avg and KD-all give the same rankings.

\minisection{Results on Variance-Limited Federated IN32} Results on ImageNet-32 are illustrated in \Cref{fig:in}. We further plot the relationship between ranking given by FD-scores and KD-scores in \Cref{fig:inrank}.

\minisection{The Effect of Intra-Client Variance} In the main body of this paper, we choose $K=20$ when we conduct the variance-limited federated CIFAR10 dataset. Hyper-parameter $K$ controls the intra-client variance: the larger the $K$, the larger the variance. The value of $K$ does not affect the key conclusion. We support this claim by conducting an ablation study on hyper-parameter $K$.
The value of $K$ is selected from \{5,10,20,50\} in our experiment. The results are illustrated in \Cref{fig:k}. Each of these figures gives a U-shape curve, which indicates that the rankings given by FD-all and FD-avg are highly inconsistent, especially when the intra-client variance and inter-client variance are mismatched.

\clearpage

\section{Proofs}
\subsection{Proof of Theorem 1}
To show this theorem, we note that if $\phi(X)$ is the kernel feature map for kernel $k$ used to define the KD distance, i.e. $k(x,y)=\langle \phi(x),\phi(y)\rangle$ is the inner product of the feature maps applied to $x,y$, then it can be seen that the kernel-$k$-based MMD distance can be written as 
\begin{align*}
    \mathrm{MMD}\bigl(P_X,P_G\bigr) :=& \mathbb{E}_{X,X'\sim P_X}\bigl[k(X,X')\bigr] + \mathbb{E}_{Y,Y'\sim P_G}\bigl[k(Y,Y')\bigr] -2\,\mathbb{E}_{X\sim P_X,\, Y\sim P_G}\bigl[k(X,Y)\bigr] \\
    =& \Bigl\Vert \mathbb{E}\bigl[\phi(X)\bigr] - \mathbb{E}\bigl[\phi(Y)\bigr] \Bigr\Vert^2.
\end{align*}
Therefore, following the definition of KD-avg, we can write
\begin{align*}
     \mathrm{KD}_{\mathrm{avg}}\Bigl( P_{X_1},\ldots,P_{X_k}\, ; \, P_G \Bigr) \, \: :=&\: \sum_{i=1}^k \lambda_i  \mathrm{KD}\bigl( P_{X_i}, P_G\bigr) \\
     =&\: \sum_{i=1}^k \lambda_i  \mathrm{MMD}_{\phi}\bigl( P_{X_i}, P_G\bigr) \\
     \stackrel{(a)}{=}&\: \sum_{i=1}^k \lambda_i  \Bigl\Vert \mathbb{E}\bigl[\phi(X_i)\bigr] - \mathbb{E}\bigl[\phi(G(Z))\bigr] \Bigr\Vert^2 \\
     \stackrel{(b)}{=}&\: \bigl\Vert\mathbb{E}\bigl[\phi(\widehat{X})\bigr] - \mathbb{E}\bigl[\phi(G(Z))\bigr]  \bigl\Vert^2 + \sum_{i=1}^k \Bigl[\lambda_i  \bigl\Vert \mathbb{E}\bigl[\phi(X_i)\bigr] - \mathbb{E}\bigl[\phi(\widehat{X})\bigr] \bigr\Vert^2 \Bigr]  \\
     \stackrel{(c)}{=}&\:  \mathrm{MMD}_\phi\bigl(\widehat{P}_{X},P_G\bigr) + \sum_{i=1}^k \Bigl[\lambda_i  \mathrm{MMD}_\phi\bigl(\widehat{P}_{X},P_{X_i}\bigr) \Bigr]  \\
     \stackrel{(d)}{=}&\:  \mathrm{KD}\bigl(\widehat{P}_{X},P_G\bigr) + \sum_{i=1}^k \Bigl[\lambda_i  \mathrm{KD}\bigl(\widehat{P}_{X},P_{X_i}\bigr) \Bigr]  \\
     \stackrel{(e)}{=}&\:   \mathrm{KD}_{\mathrm{all}}\Bigl( P_{X_1},\ldots,P_{X_k}\, ; \, P_G \Bigr) + \sum_{i=1}^k \lambda_i  \mathrm{KD}\bigl(\widehat{P}_{X},P_{X_i}\bigr) .
\end{align*}
In the above, $(a)$ and $(c)$ follow from the feature-map-based formulation of the MMD distance. $(b)$ is the consequence of the fact that $\Vert\cdot\Vert$ is the norm in a reproducing kernel Hilbert space and for $\widehat{X}$ distributed as $\widehat{P}_X = \sum_{i=1}^k \lambda_i P_{X_i}$ we know that $\mathbb{E}\bigl[\phi(\widehat{X})\bigr]$ is the weighted barycenter of the individual mean vectors $\mathbb{E}\bigl[\phi({X}_1)\bigr],\ldots,\mathbb{E}\bigl[\phi({X}_k)\bigr]$. $(d)$ is based on the definition of KD. Finally, $(e)$ follows from the definition of KD-all, which completes the proof.

\subsection{Proof of Theorem 2}
\begin{enumerate}[leftmargin=*]
    \item Note that according to the definition,
    \begin{equation*}
        \mathrm{FD}_{\mathrm{all}}\Bigl( P_{X_1},\ldots,P_{X_k}\, ; \, P_G \Bigr) \, = \, \mathrm{FD}\bigl( \sum_{i=1}^k \lambda_i P_{X_i}, P_G \bigr). 
    \end{equation*}
    Since the FD score depends only on the mean and covariance parameters in the embedding-based semantic space, we can replace $ \sum_{i=1}^k \lambda_i P_{X_i}$ with any other distribution that shares the same mean and covariance parameters, and the FD value will not change. Observe that given mean parameters $\boldsymbol{\mu}_1,\ldots ,\boldsymbol{\mu}_k$, the embedding-based mean of $\sum_{i=1}^k \lambda_i P_{X_i}$ will be $\widehat{\boldsymbol{\mu}}=\sum_{i=1}^k \lambda_i \boldsymbol{\mu}_i$. Therefore, the embedding-based covariance matrix of $\sum_{i=1}^k \lambda_i P_{X_i}$ follows from
    \begin{align*}
        \sum_{i=1}^k \lambda_i \mathbb{E}_{P_i}\bigl[\bigl( X_i - \widehat{\boldsymbol{\mu}} \bigr)\bigl( X_i - \widehat{\boldsymbol{\mu}} \bigr)^\top \bigr]\: & =\: \sum_{i=1}^k \lambda_i \Bigl[C_i + \bigl( \boldsymbol{\mu}_i - \widehat{\boldsymbol{\mu}} \bigr)\bigl( \boldsymbol{\mu}_i - \widehat{\boldsymbol{\mu}} \bigr)^\top \Bigr] \\
        & =\: \sum_{i=1}^k \lambda_i \Bigl[C_i + \boldsymbol{\mu}_i\boldsymbol{\mu}_i^\top  \Bigr]- \widehat{\boldsymbol{\mu}} \widehat{\boldsymbol{\mu}} ^\top \\
        & =\: \widehat{C}.
    \end{align*}
    Therefore, since we assume $\widehat{X}$ has the embedding-based mean and covariance $\widehat{\boldsymbol{\mu}}$ and $\widehat{C}$, the proof of this part is complete.
    \item According to the definition, FD-avg can be written as
    \begin{equation*}
        \mathrm{FD}_{\mathrm{avg}}\Bigl( P_{X_1},\ldots,P_{X_k}\, ; \, P_G \Bigr) \, \: :=\: \sum_{i=1}^k \lambda_i  \mathrm{FD}\bigl( P_{X_i}, P_G\bigr).
    \end{equation*}
    Therefore, we have
    \begin{align*}
        &\mathrm{FD}_{\mathrm{avg}}\Bigl( P_{X_1},\ldots,P_{X_k}\, ; \, P_G \Bigr) \, \\
        \stackrel{(a)}{=}\: &\sum_{i=1}^k \lambda_i  W^2_2\Bigl( \mathcal{N}(\boldsymbol{\mu}_i,C_i), \mathcal{N}(\boldsymbol{\mu}_G,C_G)\Bigr) \\
        \stackrel{(b)}{=}\: &\sum_{i=1}^k \lambda_i  \Bigl[ \Vert \boldsymbol{\mu}_i - \boldsymbol{\mu}_G\Vert^2_2 + \mathrm{Tr}\bigl(C_i +C_G - 2(C_iC_G)^{1/2}\bigr)\Bigr] \\
        =\: &\sum_{i=1}^k \Bigl[\lambda_i   \Vert \boldsymbol{\mu}_i - \boldsymbol{\mu}_G\Vert^2_2\Bigr] + \sum_{i=1}^k \Bigl[\lambda_i\mathrm{Tr}\bigl(C_i +C_G - 2(C_iC_G)^{1/2}\bigr)\Bigr] \\ 
        \stackrel{(c)}{=}\: &\Vert \widehat{\boldsymbol{\mu}} - \boldsymbol{\mu}_G\Vert^2_2 + \sum_{i=1}^k \Bigl[\lambda_i\Vert \widehat{\boldsymbol{\mu}} - \boldsymbol{\mu}_i\Vert^2_2\Bigr]  \\
        &\qquad +\mathrm{Tr}\bigl(C_G +\widehat{C} - 2(C_G\widehat{C})^{1/2}\bigr) + \sum_{i=1}^k \Bigl[\lambda_i\mathrm{Tr}\bigl(C_i +\widehat{C} - 2(C_i\widehat{C})^{1/2}\bigr) \Bigr] \\
        =\: &\Vert \widehat{\boldsymbol{\mu}} - \boldsymbol{\mu}_G\Vert^2_2 +\mathrm{Tr}\bigl(C_G +\widehat{C} - 2(C_G\widehat{C})^{1/2}\bigr)  \\
        &\qquad + \sum_{i=1}^k \Bigl[\lambda_i\Vert \widehat{\boldsymbol{\mu}} - \boldsymbol{\mu}_i\Vert^2_2 + \lambda_i\mathrm{Tr}\bigl(C_i +\widehat{C} - 2(C_i\widehat{C})^{1/2}\bigr) \Bigr] \\
         =\: &\Vert \widehat{\boldsymbol{\mu}} - \boldsymbol{\mu}_G\Vert^2_2 +\mathrm{Tr}\bigl(C_G +\widehat{C} - 2(C_G\widehat{C})^{1/2}\bigr)  \\
        &\qquad + \sum_{i=1}^k \lambda_i\Bigl[\Vert \widehat{\boldsymbol{\mu}} - \boldsymbol{\mu}_i\Vert^2_2 + \mathrm{Tr}\bigl(C_i +\widehat{C} - 2(C_i\widehat{C})^{1/2}\bigr) \Bigr] \\
         \stackrel{(d)}{=}\: &\mathrm{FD}(P_{\widehat{X}},P_G) + \sum_{i=1}^k  \lambda_i\mathrm{FD}(P_{\widehat{X}},P_{X_i}).
    \end{align*}
    In the above, $(a)$ follows from the Wasserstein-based definition of FD distance. $(b)$ comes from the well-known closed-form expression of the 2-Wasserstein distance between Gaussian distributions \citep{villani2009optimal}. $(c)$ is the result of applying the weighted barycenter of vector $\boldsymbol{\mu}_1,\ldots,\boldsymbol{\mu}_k$ that can be seen to be $\widehat{\boldsymbol{\mu}}$ and the weighted barycenter of positive semi-definite covariance matrices $C_1,\ldots , C_k$ that has been shown to be the unique matrix $\widehat{C}$ that solves the equation $\widetilde{C} = \sum_{i=1}^k \lambda_i \bigl(\widetilde{C}^{1/2}C_i \widetilde{C}^{1/2}\bigr)^{1/2}$ \citep{ruschendorf2002n,puccetti2020computation}. $(d)$ is the direct consequence of the Wasserstein-based definition of the FD distance and the closed-form expression of the 2-Wasserstein distance between Gaussians. Therefore, the proof is complete.
\end{enumerate}

\subsection{Proof of Proposition \ref{prop: FD-equal}}

Consider the FD-all-minimizing parameters in Theorem~\ref{Thm: FD} resulting in 
\begin{align*}
        \mathrm{FD}_{\mathrm{all}}\Bigl( P_{X_1},\ldots,P_{X_k}\, ; \, P_{\widehat{G}} \Bigr) \, &= \, \mathrm{FD}\bigl( \mathcal{N}(\widehat{\boldsymbol{\mu}},\widehat{C}), \mathcal{N}({\boldsymbol{\mu}_{\widehat{G}}},{C}_{\widehat{G}}) \bigr) \\
        \,  &= \, \bigl\Vert \widehat{\boldsymbol{\mu}} - \boldsymbol{\mu}_{\widehat{G}} \bigr\Vert^2_2 + \mathrm{Tr}\Bigl(\widehat{C}+C_{\widehat{G}} - 2\bigl(\widehat{C} C_{\widehat{G}}\bigr)^{1/2}\Bigr).
    \end{align*}

    Note that since we assume  the number of clients $k$ is less than the dimension of the embedding, there exists a unit-norm vector $\boldsymbol{\beta}$ ($\Vert \boldsymbol{\beta}\Vert_2 =1$)  in the embedding space that is orthogonal to all mean vectors $\boldsymbol{\mu}_1,\ldots ,\boldsymbol{\mu}_k $ and hence to their mean $\widehat{\boldsymbol{\mu}} = \frac{1}{k}\sum_{i=1}^k \boldsymbol{\mu}_i$. Given $u=\mathrm{Tr}\bigl(\sum_{i=1}^k \lambda_i \bigl(\boldsymbol{\mu}_i\boldsymbol{\mu}_i^\top - \widehat{\boldsymbol{\mu}}\widehat{\boldsymbol{\mu}}^\top\bigr) \bigr)$, we then consider the generator $G'$ with the following mean and covariance parameters:
    \begin{align*}
        \boldsymbol{\mu}_{G'} \, =\, \widehat{\boldsymbol{\mu}} + \sqrt{u}\boldsymbol{\beta},\qquad  C_{G'} = C_{\widehat{G}} -  \sum_{i=1}^k \lambda_i \bigl(\boldsymbol{\mu}_i\boldsymbol{\mu}_i^\top - \widehat{\boldsymbol{\mu}}\widehat{\boldsymbol{\mu}}^\top \bigl)\, =\, \sum_{i=1}\lambda_i C_i.
    \end{align*}

    We claim that the generators $\widehat{G}$ and $G'$ lead to the same client-based FD scores as for every $i$
    \begin{align*}
        \mathrm{FD}\bigl(P_{X_i}, P_{G'} \bigr) \, &= \, \mathrm{FD}\bigl( \mathcal{N}({\boldsymbol{\mu}}_i,{C}_i), \mathcal{N}({\boldsymbol{\mu}_{{G}'}},{C}_{{G}'}) \bigr) \\
        \,  &= \, \bigl\Vert {\boldsymbol{\mu}}_i - \boldsymbol{\mu}_{{G}'} \bigr\Vert^2_2 + \mathrm{Tr}\Bigl({C}_i+C_{{G}'} - 2\bigl({C}_i C_{{G}'}\bigr)^{1/2}\Bigr) \\
        &= \bigl\Vert {\boldsymbol{\mu}}_i - \boldsymbol{\mu}_{\widehat{G}} \bigr\Vert^2_2 + u + \mathrm{Tr}\Bigl({C}_i+C_{{G}'} - 2\bigl({C}_i C_{\widehat{G}}\bigr)^{1/2}\Bigr) -u 
        \\
        &= \bigl\Vert {\boldsymbol{\mu}}_i - \boldsymbol{\mu}_{\widehat{G}} \bigr\Vert^2_2 + \mathrm{Tr}\Bigl({C}_i+C_{{G}'} - 2\bigl({C}_i C_{\widehat{G}}\bigr)^{1/2}\Bigr) \\
        &= \mathrm{FD}\bigl(P_{X_i}, P_{\widehat{G}} \bigr).
    \end{align*}
    On the other hand, for the FD-all of $G'$ we have

    \begin{align*}
        \mathrm{FD}_{\mathrm{all}}\Bigl( P_{X_1},\ldots,P_{X_k}\, ; \, P_{{G}'} \Bigr) \, &= \, \mathrm{FD}\bigl( \mathcal{N}(\widehat{\boldsymbol{\mu}},\widehat{C}), \mathcal{N}({\boldsymbol{\mu}_{{G}'}},{C}_{{G}'}) \bigr) \\
        \,  &= \, \bigl\Vert \widehat{\boldsymbol{\mu}} - \boldsymbol{\mu}_{{G}'} \bigr\Vert^2_2 + \mathrm{Tr}\Bigl(\widehat{C}+C_{{G}'} - 2\bigl(\widehat{C} C_{{G}'}\bigr)^{1/2}\Bigr) \\ \,  &= \, \bigl\Vert \widehat{\boldsymbol{\mu}} - \boldsymbol{\mu}_{\widehat{G}} \bigr\Vert^2_2 + u \mathrm{Tr}\Bigl(\widehat{C}+C_{\widehat{G}} - 2\bigl(\widehat{C} C_{\widehat{G}}\bigr)^{1/2}\Bigr) + u \\ 
        \,  &= \, \bigl\Vert \widehat{\boldsymbol{\mu}} - \boldsymbol{\mu}_{\widehat{G}} \bigr\Vert^2_2 + \mathrm{Tr}\Bigl(\widehat{C}+C_{\widehat{G}} - 2\bigl(\widehat{C} C_{\widehat{G}}\bigr)^{1/2}\Bigr)\\
        &\;\; + 2\mathrm{Tr}\bigl(\sum_{i=1}^k \lambda_i \bigl(\boldsymbol{\mu}_i\boldsymbol{\mu}_i^\top - \widehat{\boldsymbol{\mu}}\widehat{\boldsymbol{\mu}}^\top\bigr) \bigr) \\
        &= \, \mathrm{FD}_{\mathrm{all}}\Bigl( P_{X_1},\ldots,P_{X_k}\, ; \, P_{\widehat{G}} \Bigr)\\
        &\;\; + 2\mathrm{Tr}\bigl(\sum_{i=1}^k \lambda_i \bigl(\boldsymbol{\mu}_i\boldsymbol{\mu}_i^\top - \widehat{\boldsymbol{\mu}}\widehat{\boldsymbol{\mu}}^\top\bigr) \bigr)
    \end{align*}
Therefore, Proposition~\ref{prop: FD-equal}'s proof is complete.

\end{document}